\documentclass{article}

\usepackage{arxiv}

\usepackage{fontspec}
\usepackage{hyperref}
\usepackage{url}
\usepackage{booktabs}
\usepackage{amsmath}      
\usepackage{amsfonts}
\usepackage{amssymb}
\usepackage{nicefrac}
\usepackage{microtype}
\usepackage{cleveref}
\usepackage{graphicx}
\usepackage[round]{natbib}
\usepackage{doi}

\usepackage{multirow}
\usepackage{array}
\usepackage{tabularx}
\usepackage{longtable}
\usepackage{xcolor}
\usepackage{listings}

\newfontfamily\devanagarifont[Path=fonts/,Script=Devanagari]{NotoSansDevanagari-Regular.ttf}
\newfontfamily\bengalifont   [Path=fonts/,Script=Bengali]   {NotoSansBengali-Regular.ttf}
\newfontfamily\tamilfont     [Path=fonts/,Script=Tamil]     {NotoSansTamil-Regular.ttf}
\newfontfamily\telugufont    [Path=fonts/,Script=Telugu]    {NotoSansTelugu-Regular.ttf}
\newfontfamily\kannadafont   [Path=fonts/,Script=Kannada]   {NotoSansKannada-Regular.ttf}
\newfontfamily\malayalamfont [Path=fonts/,Script=Malayalam] {NotoSansMalayalam-Regular.ttf}
\newfontfamily\gujaratifont  [Path=fonts/,Script=Gujarati]  {NotoSansGujarati-Regular.ttf}
\newfontfamily\gurmukhifont  [Path=fonts/,Script=Gurmukhi]  {NotoSansGurmukhi-Regular.ttf}
\newfontfamily\oriyafont     [Path=fonts/,Script=Oriya]     {NotoSansOriya-Regular.ttf}
\newcommand{\devtext}[1]{{\devanagarifont #1}}
\newcommand{\bentext}[1]{{\bengalifont #1}}

\newcommand{\teltext}[1]{{\telugufont #1}}
\newcommand{\kantext}[1]{{\kannadafont #1}}

\definecolor{codegreen}{rgb}{0,0.6,0}
\definecolor{codegray}{rgb}{0.5,0.5,0.5}
\definecolor{codeblue}{rgb}{0,0,0.7}
\definecolor{backcolour}{rgb}{0.96,0.96,0.94}
\lstdefinestyle{paperstyle}{
  backgroundcolor=\color{backcolour},
  commentstyle=\color{codegreen},
  keywordstyle=\color{codeblue},
  numberstyle=\tiny\color{codegray},
  stringstyle=\color{codegreen},
  basicstyle=\ttfamily\footnotesize,
  breakatwhitespace=false, breaklines=true, captionpos=b,
  keepspaces=true, showspaces=false, showstringspaces=false, showtabs=false,
  tabsize=2
}
\title{Reversible Foundations:\\Training a 120B Sparse MoE through State-Preserving Scaling}

\author{
  Rohan Shravan \\
  The School of AI \\
  Bengaluru, India \\
  \texttt{rshravan@theschoolofai.in}
}

\renewcommand{\shorttitle}{Reversible Foundations}
\renewcommand{\headeright}{Technical Report}
\renewcommand{\undertitle}{Technical Report}

\hypersetup{
  hidelinks,
  pdftitle={Reversible Foundations: Training a 120B Sparse MoE through State-Preserving Scaling},
  pdfauthor={Rohan Shravan},
  pdfkeywords={mixture of experts, model growth, reversibility, single-node training, low-rank adaptation, TurboQuant},
}

\begin{document}
\maketitle

\begin{abstract}
This paper reports on training a hundred-billion-parameter sparse mixture of experts on a single eight-GPU node, end to end, and on what made it possible. LightningLM 0.1V is a recurrence-backbone language model family grown in four stages from a small dense seed, through a 5B and a 9B mixture of experts, to a 120B model with 460 routed experts under top-12 routing. Each larger model is grown from the trained weights of the smaller one rather than trained from scratch, and the active parameter count rises monotonically across the lineage, from 1.78B in the dense seed to 5.93B at 120B, roughly five percent of the 118.67B stored. The full lineage runs on single nodes, the early stages at 4K context and the larger stages from 9B onward at full 8K, reaching a released training loss of 1.78 at the 120B scale.

This is a systems and experience report rather than a benchmark paper. Its purpose is to write down the operational recipe that public reports of model growth usually withhold, and the failures that recipe is shaped to avoid. The work is organized around three disciplines. The first is reversibility. A reversible recurrence stack reconstructs activations in the backward pass instead of storing them, which holds activation memory flat as the model grows in depth and experts and is the reason the larger stages fit on one node at long context. The second is state-preserving growth, which treats each expansion as the preservation of a learned interface: dense to mixture of experts, shallow to deep, few experts to many, each operation given as a reproducible principle and paired with the failure that results from getting it wrong. Several of those failures are silent, producing a plausible checkpoint or a plausible loss while violating an invariant the run depends on, and those are the expensive ones. The third discipline is single-node economics. The 120B is trained through TQP, a strategy of quantized base expert weights and trained low-rank adapters that carries optimizer state on 2.26B adapter parameters rather than the more than 100B parameters resident in the routed experts, cutting expert-path optimizer state by roughly a factor of 45. The quantizer is a repurposing of a published vector quantizer, and that repurposing, rather than any quantization or adaptation primitive, is the novel part.

None of the underlying primitives is new. Growth, reversibility, gated linear and learnable sparse attention, low-rank adaptation, and dynamic data selection are each established. What is new at the architectural level is narrow and stated as such: a three-scale recurrence, carrying token-level linear-attention state, cross-stream mixing, and a cross-chunk memory vector, held unchanged across the grown family. What is new at the systems level is the integration, one grown lineage that runs end to end on a single node, documented at the level of detail a practitioner would need to reproduce it. The measure of success is defined and defended rather than apologized for: sustained loss reduction on held-out data of increasing difficulty, with per-domain held-out loss as the evidence that targeted capabilities, multilingual Indic competence and code among them, were learned by construction rather than by chance. The model family, the tokenizer, and the training code are released, and the report marks throughout which claims rest on released artifacts, which on saved logs, and which on training-session records that were not committed to durable storage.
\end{abstract}


\section{Introduction}
\label{sec:introduction}

The prevailing assumption about training a model in the hundred-billion-parameter class is that it requires a cluster. This paper reports on training one on a single eight-GPU node, end to end, and on what that required. The claim is not that a single node is the right way to train at this scale, or the fastest, but something narrower and more useful: substantially more usable training can be extracted from one node than the field generally assumes. Reaching that ceiling forces a set of decisions about memory, about how a model is grown, and about how its training cost is bounded, and most of those decisions are absent from the public record. Writing them down is the point of the paper.

LightningLM 0.1V is the model family this report is built around. It is a recurrence-backbone language model grown in four stages from a small dense seed, through a 5B and a 9B mixture of experts, to a 120B sparse mixture of experts with 460 routed experts under top-12 routing. The family shares one architecture. The larger models are grown from the trained weights of the smaller ones rather than trained from scratch, and the active parameter count rises across the lineage from 1.78B in the dense seed to 5.93B at 120B, about five percent of the 118.67B parameters stored at the largest scale. The whole lineage runs on single eight-GPU nodes at real batch sizes, with the early stages trained at 4K context and the larger stages from 9B onward at full 8K. At the 120B scale that regime is only reachable because of the systems decisions this paper is mostly about.

Two components of the family are documented separately, in companion work, and are leaned on here rather than re-derived. The vocabulary is the 131{,}072-token BrahmicTokenizer-131K, a byte-level BPE tokenizer that closes the Indic compression gap at the 131K-vocabulary class while preserving English and code compression \citep{Brahmic}. The input embedding is not a learned lookup table but a Kronecker construction that builds each token's vector from frozen byte and position bases through a single trained projection \citep{Kronecker}. The accounting consequence is larger than it first appears and matters most at the small end of the family. A standard table at this vocabulary and width would be roughly 537M trainable parameters; the Kronecker projection that replaces it is about 33.55M. Against the 118B model that difference is rounding, but against the 1.78B dense seed it is most of a quarter of the model, and the active-parameter counts reported here exclude it throughout. The memory consequence is separate and, for single-node training, decisive. The full reconstruction table would occupy about 2.15\,GB per GPU and roughly 17\,GB across a node if it were ever materialized, and because the embedding is computable from byte and position structure rather than stored, it is reconstructed on the fly and never pays that residency, an option available to this architecture precisely because the embedding is a function rather than a lookup. Without it the smaller models would not have fit a single node at the batch size that gives good throughput, and the speed reported at every scale depends on it. Both the tokenizer and the embedding are prior contributions of this group; this paper treats them as settled foundations and concentrates on what it takes to grow and train the model that sits on top of them.

\subsection{The open secret of growth}
\label{sec:intro-growth}

Growing a large model from a smaller trained one is not novel, and this paper does not present it as such. Public models are upcycled from dense checkpoints into mixtures of experts, scaled in depth by restacking trained layers, and expanded in expert count mid-training. What is missing from the public record is the operational recipe. Frontier reports acknowledge growth in passing, sometimes only implicitly, as when a technical report discloses that results were extrapolated from smaller runs while withholding the method that relates them. The practitioner who actually wants to grow a model is left to rediscover, usually the expensive way, which parts of a trained model can be transformed and which break if touched.

This is the gap the paper aims at. Growth is best understood as the preservation of learned interfaces. A trained model is a set of parts that have learned to talk to each other, and successful growth carries those interfaces forward intact rather than forcing the larger model to relearn what the smaller one already knew. Section~\ref{sec:growth} states this as a set of concrete, reproducible principles, one per growth operation, each paired in Section~\ref{sec:failure-modes} with the failure that results from getting it wrong. The failures are not incidental to the report. Several of them produce a plausible checkpoint or a plausible loss while silently violating an invariant the run depended on, and those are the expensive ones, worth documenting so another team does not pay for them again.

\subsection{Reversibility as the enabler}
\label{sec:intro-reversibility}

If there is a single decision that made the rest possible, it is reversibility. A conventional transformer stores the activations of every layer during the forward pass, so activation memory grows with depth and sequence length and, at long context, dominates the memory budget. A reversible stack reconstructs those activations during the backward pass instead of storing them, trading extra compute for memory that does not grow with depth; this work adopts the reversible-transformer construction of Gal and colleagues \citep{Reversibility} as the architectural backbone, and carries it to a regime that work poses as future scope. This is what keeps the larger stages fitting on one node. As the model grows in depth and experts and the context stays long, the term that would otherwise explode, stored activations, is held flat, and what grows instead is parameter and optimizer state, which can be sharded and quantized. The quantization that bounds the largest stage is itself a repurposing: a published data-oblivious vector quantizer, built for inference-time compression, is used here as the trainable quantized base of the expert stack, which Section~\ref{sec:turboquant} develops as one of the paper's distinct contributions.

The kind of claim made for reversibility is deliberately bounded, because it governs the evidence. This is a systems and experience report rather than a controlled study. No matched reversible-versus-non-reversible comparison of final loss at scale is presented, because the non-reversible configuration is the one that does not fit the regime in question. The evidence for reversibility is feasibility and throughput. It turns configurations that do not fit into configurations that fit with headroom, and with the kernel and runtime work of Section~\ref{sec:throughput} the resulting stack trains fast enough to finish. Reversibility is treated as the first contribution of this paper not because it is novel in isolation but because it is the load-bearing decision the rest of the recipe stands on.

\subsection{How success is measured}
\label{sec:intro-success}

This paper does not report benchmark scores, and a reader expecting a leaderboard will not find one. The measure of success is defined and defended rather than apologized for. It is sustained loss reduction on held-out data of increasing difficulty. As the model grows and the curriculum advances from easier to harder material, the held-out loss on each domain should continue to fall, and the per-domain held-out loss at each stage is the evidence reported. The choice of a loss-based measure is deliberate and follows from what the model was built to do. The model was built to acquire specific capabilities, multilingual Indic competence and code among them, by construction rather than by chance, and the per-domain held-out loss is the direct measurement of whether that construction worked. Section~\ref{sec:data} explains the data methodology that makes those capabilities a per-batch guarantee, and Section~\ref{sec:results} reports the loss evidence that they were learned. Where generation samples are shown, they are qualitative illustration rather than scored results, and are marked as such.

\subsection{Contributions}
\label{sec:intro-contributions}

The contributions of this paper are not new primitives. They are an integration, a recipe, and a catalogue of what the recipe is shaped to avoid, developed in the order the paper presents them and each tied to a defined class of evidence. They are stated with their boundaries attached, because the boundary is as much the contribution as the claim.

\textbf{1. Reversibility as the scaling backbone.} A reversible recurrence stack, adopted from \citet{Reversibility}, with its production integrator configuration and the runtime and allocator work needed to make it stable and fast, holds activation memory flat as the model grows and is the reason the larger stages fit on one node. The reversible-transformer construction itself is prior work; the contribution here is its production hardening, sharding-aware backward path, and concrete deployment at the 120B-class MoE scale that \citet{Reversibility} themselves frame as future scope. The claim is a feasibility-and-throughput claim rather than a loss claim. It does not assert that reversibility lowers final loss, because the comparison that would establish that is the non-reversible run that does not fit the regime. What it asserts, supported with memory-feasibility and production-throughput evidence, is that reversibility makes the 5B, 9B, and 120B regime reachable on a single node, with the released 2B trained non-reversibly because at that scale the memory did not bind. Sections~\ref{sec:reversibility} and~\ref{sec:throughput}.

\textbf{2. An operational recipe for state-preserving growth.} Concrete, reproducible principles are given for each growth operation performed, dense to mixture of experts, shallow to deep, and few experts to many, framed throughout as the preservation of a learned interface. Each principle is stated in a liftable form, with the mechanism in enough detail to reproduce, the alternatives tested, the principle another team could apply, and an explicit grade of how strong the evidence is. The growth idea is prior work, surveyed in Section~\ref{sec:background}. The operational detail at this level, across all three axes in one lineage, is the contribution. Section~\ref{sec:growth}.

\textbf{3. A catalogue of growth failure modes.} The failures the principles are designed to avoid are documented at the same level of detail as the principles themselves, because a recipe that names only its final choices is half a recipe. Each failure is given as a violated invariant, with what was tried, how it broke, the root cause, the correction, and its evidence grade. Several of these failures are silent. They pass every cheap check, producing a plausible checkpoint or a plausible loss while an invariant the run depended on has already been broken, and those are the ones worth the most to document. Section~\ref{sec:failure-modes}.

\textbf{4. Single-node economics at 120B through TQP.} TQP is the training strategy that makes a 120B sparse mixture of experts trainable on one node, by holding the base expert weights in a quantized representation and training low-rank adapters over them, so optimizer state is carried on 2.26B adapter parameters rather than the more than 100B parameters resident in the routed experts. This cuts expert-path optimizer state by roughly a factor of 45 and is the arithmetic that lets the run fit. The novelty boundary is stated precisely. Quantized low-rank adaptation and the periodic merge-and-reset flush are both prior work. A measurement of the 9B weight spectrum argues against the usual low-rank justification for the method, and that measurement is reported as a falsifier that reshaped the account of why the method works. The periodic flush, carried up from sub-1B experiments where it succeeded, diverged at 120B and was abandoned; the released lineage is flushless, and the divergence is reported as a finding rather than smoothed over. Sections~\ref{sec:turboquant} and~\ref{sec:training-120b}.

\textbf{5. Repurposing a data-oblivious vector quantizer as a trainable base.} The quantizer underneath TQP is TurboQuant, a published data-oblivious online vector quantizer whose demonstrated targets were inference-time vectors such as KV-cache and nearest-neighbor compression. Here its MSE path is used instead as the trainable quantized base of a 460-expert mixture-of-experts stack, with the model trained through that representation by low-rank adapters. The quantizer is not a contribution of this work; the contribution is the repurposing and the demonstration that an analytically-quantized base, which needs no calibration and so survives a moving base, is trainable at this scale. As far as can be determined, an inference-side vector quantizer used as a trainable MoE base has not previously been reported. Section~\ref{sec:turboquant}.

\textbf{6. Guaranteed-composition data methodology.} A capability-selecting data scorer, left to its own objective, systematically starves the capabilities its reference proxy under-values, Indic text and code among them. The three-tier system fixes this by guaranteeing a fixed share of every batch to data the scorer would discard, regardless of what it prefers. The per-domain held-out loss is the measured payoff, and each result is traced back to the per-batch guarantee that produced it. The selector being wrapped is prior work; the guaranteed-composition architecture around its blind spot, and the mechanism-to-result link, are the contribution. Section~\ref{sec:data}.

None of the underlying primitives is presented as new. Growth, reversibility, gated linear and learnable sparse attention, low-rank adaptation, the periodic flush, the vector quantizer, and dynamic data selection are each prior work, as Section~\ref{sec:background} lays out, as are the tokenizer and the embedding the family is built on, which are this group's own companion contributions \citep{Brahmic,Kronecker}. What this paper contributes is their integration into one grown lineage that runs end to end on a single node, documented at the level of detail a practitioner would need to do it again, together with the failures that detail is shaped to avoid. The model family is the evidence that the integration works. The recipe and the failures are the part meant to be carried forward.

\section{Background and Related Work}
\label{sec:background}

This work sits at the intersection of four lines of research: growing models by reusing trained weights, linear and sparse attention as a backbone, low-rank and periodically-merged adaptation, and dynamic data selection. The contributions are placed against each, with care throughout about the boundary between what is prior work and what is new, because the rest of the paper depends on that boundary being honest. Two further foundations the model rests on, its tokenizer and its embedding, are this group's own prior work rather than part of this contribution, and are marked as such where they enter.

\subsection{Growth as a scaling strategy, and the open secret}
\label{sec:bg-growth}

The idea that a larger model can be initialized from a smaller trained one, rather than from scratch, is not new and is increasingly common. The sparsely-gated mixture-of-experts architecture itself, foundational to the width-axis growth this paper performs, is from \citet{Shazeer}. The originating reference for growing into that architecture is sparse upcycling, which initializes a sparsely activated mixture of experts from a dense checkpoint and recovers most of the dense model's sunk training cost, shown across T5 and Vision Transformer scales \citep{SparseUpcycling}. The pattern has since become a standard production move. Qwen1.5-MoE upcycles a 1.8B dense model by partitioning its feed-forward network into fine-grained experts, reaching quality comparable to a 7B dense model at roughly a quarter of the training cost \citep{QwenMoE}. The same fine-grained pattern, with a shared general expert alongside partitioned specialists, recurs in work that extends a 7B dense model into a fine-grained MoE during continued pretraining \citep{Innovator}. On the depth axis, SOLAR extends a 7B model to 10.7B by duplicating and restacking layers \citep{SOLAR}, and LLaMA Pro grows depth by inserting and selectively training new layers under a stabilizing recipe \citep{LLaMAPro}. NVIDIA's Nemotron work converts a trained dense model into an MoE with virtual-group initialization and weight scaling \citep{Nemotron}, and GroveMoE expands capacity mid-training by upcycling a 30B dense model \citep{GroveMoE}.

A distinct axis is distillation, where a larger model trains a smaller one, as in Gemma 2 \citep{Gemma2}. It is noted here only to set it aside. Distillation moves knowledge downward, from large to small, whereas the concern of this work is the upward direction, building a larger model from a smaller trained one. They are different operations, and the distillation literature is not claimed as a precedent for growth.

What is striking about this body of work is less that growth happens and more what is withheld. Frontier reports acknowledge growth obliquely. The GPT-4 technical report, for instance, discloses that performance was extrapolated from smaller training runs, which presupposes a methodology for relating small and large models, but withholds the method itself \citep{GPT4}. The pattern across the field is that growth is an open secret. Many labs do it, few publish the operational recipe, and the practitioner who wants to grow a model from a small seed is left to rediscover the seam-preservation principles that Section~\ref{sec:growth} sets out. The idea of growth is prior, and none of it is claimed here. The contribution on this axis is the operational recipe at the level of detail needed to reproduce it, together with the documentation of the failures that recipe avoids. The growth in this work spans all three axes in one lineage, dense to MoE, shallow to deep, and few experts to many, on a single node, with the transforms reported rather than gestured at.

\subsection{Architectural shape: depth, width, and the backbone}
\label{sec:bg-architecture}

Two architectural choices in this work draw on prior arguments and deserve placement before the method sections use them.

The first is the relationship between depth and scale. The conventional path stacks layers as parameter count grows. Recent analysis argues this is often the wrong move: beyond a critical depth that scales only sublinearly with width, additional layers raise loss rather than lowering it, so capacity is better spent on width or sparsity than on further depth \citep{DepthDelusion}. This work adopts the standard parameter accounting for transformers, that the active dense path of a transformer is approximately $N \approx 12Ld^2$ in the number of layers $L$ and width $d$, but applies it to a quantity that critical-depth analysis does not: the active path of a sparse model rather than its total parameter count. Section~\ref{sec:system} sets out the resulting rule, which fixes depth from the size of the computation a single token actually traverses and provides the remaining capacity through experts. The heuristic is borrowed; anchoring it to the active path of a grown sparse model is the part used here as original.

The second is the backbone itself, which departs from full quadratic attention. The token-level recurrence is a gated linear-attention mechanism in the DeltaNet family, maintaining a matrix-valued state updated by a delta rule with content-dependent gates \citep{DeltaNet,GatedDeltaNet}, which gives linear-in-length cost rather than quadratic. The complementary layers use a learnable sparse attention that attends to a variable per-token budget of keys \citep{SparseAttn}. The cross-stream mixing is Manifold-Constrained Hyper-Connections \citep{mHC}, a constrained form of the Hyper-Connections residual scheme \citep{HyperConnections} that uses Sinkhorn-Knopp normalization \citep{SinkhornKnopp} to keep the mix doubly-stochastic. The reversible-transformer construction the backbone is wrapped in, which holds activation memory flat as depth and sequence length grow, is from \citet{Reversibility}; that work introduces the reversible midpoint stack and the integrator configuration this paper deploys at scale. None of these mechanisms is original to this work. They are used, in a fixed repeating arrangement, as the constant backbone the growth operates over. The input embedding is likewise not a learned lookup table but a Kronecker construction over frozen byte and position bases, documented in companion work and treated here as a settled foundation rather than a contribution of this paper \citep{Kronecker}. What is original at the architectural level is the three-scale recurrence composition of Section~\ref{sec:system}, the combination of token-level linear-attention state, cross-stream mixing, and a cross-chunk memory vector, carried unchanged across the grown family. The individual mechanisms are from the literature. Their composition, and its preservation through growth, are the architectural commitment this paper reports.

\subsection{Low-rank structure and periodically-merged adaptation}
\label{sec:bg-lowrank}

The economics of training the 120B rest on low-rank adaptation, and the boundary between what is established and what is ours matters most here, because it is the easiest place to over-claim.

That neural network updates have low intrinsic dimension is well established. Aghajanyan and colleagues showed that fine-tuning operates in a low-dimensional subspace and that larger models have lower intrinsic dimension \citep{Aghajanyan}. LoRA built a method on the related observation that the update to a weight matrix can be well approximated by a low-rank factorization \citep{LoRA}. A line of theory and measurement explains why low rank emerges during training at all. Stochastic optimization with weight decay provably biases toward low-rank solutions \citep{Galanti}, the spectral dynamics of weights over training track this rank reduction \citep{Yunis}, and weight decay has been shown to induce low-rank structure specifically in attention during large-model training \citep{Kobayashi}. Methods that vary the adapter rank per layer according to an estimate of intrinsic dimension also exist \citep{GeLoRA}. This literature is relied on in Section~\ref{sec:turboquant}, where it frames the rank measurement as a forward-looking instrument, and is not contributed to here.

The flush primitive, periodically merging a low-rank adapter into the base weights and reinitializing it so a sequence of low-rank updates accumulates into a higher-rank change, is prior work. ReLoRA introduced merge-and-reinitialize for pretraining from scratch \citep{ReLoRA}. PeriodicLoRA developed the same accumulate-merge-reset idea for fine-tuning \citep{PeriodicLoRA}. Quantized dynamic low-rank adaptation has likewise been studied \citep{QDyLoRA}. This work makes no claim to be first to flush a low-rank adapter, first to do so during pretraining, or first to combine quantization with low-rank adaptation. Each of those is established. The flush is relevant here for the opposite reason to a usual building block: it was the assumed mechanism going into the 120B and it did not transfer, diverging at scale and being abandoned in favor of flushless training, which Sections~\ref{sec:turboquant} and~\ref{sec:failure-modes} report as a finding.

What this work claims on the adaptation axis is twofold, and neither part is the flush. The first is the repurposing of TurboQuant, a published data-oblivious vector quantizer built for inference-time compression \citep{TurboQuant}, as the trainable quantized base of a drop-upcycled 120B mixture of experts, with the model trained through that representation by low-rank adapters on a single node at 8K context. As far as can be determined, an inference-side vector quantizer used as a trainable expert base has not previously been reported. The second is the empirical finding that the flush, effective below 1B, diverges at this scale. A study of ReLoRA on small language models reports that merge-and-restart helps larger models but not capacity-limited small ones, because a small model is already rank-deficient and the rank-expanding update has nowhere to expand \citep{ReLoRASLM}; that finding and the divergence reported here both indicate that flush value depends on rank headroom, which varies with scale and training maturity, and Section~\ref{sec:turboquant} develops the connection. The expert expansion itself uses drop-upcycling, reinitializing a fraction of each cloned expert to break symmetry, following Nakamura and colleagues, whose observation that the benefits of drop-upcycling emerge only after extended training is independently consistent with the clone-collapse dynamics documented in Section~\ref{sec:failure-modes} \citep{DropUpcycling}.

\subsection{Dynamic data selection and the tokenizer foundation}
\label{sec:bg-data}

The data pipeline of Section~\ref{sec:data} uses a dynamic selector, OPUS, which scores candidate data by projecting optimizer-shaped updates onto a target direction derived from a held-out proxy \citep{OPUS}. The selector is published work, used unchanged. What this work adds is the architecture around its blind spot: the guaranteed tier, and the per-batch composition guarantee it enforces. The blind spot is that a proxy-aligned scorer systematically under-values data far from its reference direction, and the guaranteed tier gives such data, Indic text and code in this case, a fixed share of every batch regardless of what the selector prefers. That mechanism, and the per-batch guarantee it enforces, are the contribution; the selector they wrap is published.

The Indic capability the data system protects is only realizable because the vocabulary can represent Indic text economically in the first place. The model uses BrahmicTokenizer-131K, a 131{,}072-token byte-level BPE tokenizer that closes the Brahmic compression gap at the 131K-vocabulary class while preserving English, European-language, and code compression \citep{Brahmic}. It is this group's own prior work, treated here as the tokenizer foundation the data methodology assumes rather than as a result of this paper. Its relevance to Section~\ref{sec:data} is direct: several of the corpus failures found and fixed, dead native-numeral tokens among them, are properties of the vocabulary and corpus considered as one design, and the tokenizer paper is the reference for that design.

\subsection{Position of this work}
\label{sec:bg-position}

Against all four lines, the shape of the contribution is the same. The components are largely from the literature. Growth, depth-versus-width architectural guidance, linear and sparse attention, low-rank and periodically-merged adaptation, and dynamic selection are each established, and the tokenizer and embedding the model rests on are this group's own prior work rather than claims of this paper. What this paper adds is their integration into a single grown lineage that runs end to end on one node, the operational detail needed to reproduce each transform, the failures that the detail avoids, and the systems work, reversibility and on-the-fly embedding reconstruction, that makes the whole thing fit. The paper does not present a new primitive. It presents a working composition of known primitives at a scale and on hardware where that composition was not previously shown to be possible, documented at the level a practitioner would need to do it again.

\section{The LightningLM System}
\label{sec:system}

This section describes the architecture that the rest of the paper refers back to. The important thing to understand about LightningLM is that the 2B, 5B, 9B, and 120B models are not four architectures. They are one architecture grown four times. A single dense backbone defines the model's identity, and every larger model inherits that backbone unchanged, altering only the feed-forward path and the depth. The backbone is described first, because it is the constant, and then each larger stage's change is stated precisely.

The backbone's identity is its recurrence. Most of this section is about three recurrence mechanisms that operate at three different scales, because that three-scale structure is the architectural commitment carried bit-for-bit from the 2B model up to the 120B. Everything above the recurrence ledger is scale.

\subsection{The constant backbone}
\label{sec:system-backbone}

The backbone is a recurrence stack over a residual stream of width 4096, with a 131{,}072-token vocabulary supplied by the BrahmicTokenizer-131K of the companion work \citep{Brahmic}. Layers come in two types arranged in a fixed repeating pattern. A D-layer runs gated linear attention; a G-layer runs learnable sparse attention. The eight-layer 2B backbone is the pattern \textit{DDDGDDDG}, two repeats of the \textit{DDDG} motif, and the deeper models repeat the same motif. Both layer types terminate in a feed-forward network, and the feed-forward network is the one part of the backbone that changes with scale.

\begin{table}[h]
\centering
\caption{Backbone properties shared across the LightningLM 0.1V family.}
\label{tab:backbone-shared}
\begin{tabular}{ll}
\toprule
Property & Value (shared across the family) \\
\midrule
Residual width & 4096 \\
Vocabulary & 131{,}072 \citep{Brahmic} \\
Embedding & Kronecker product, $256 \times 32 = 8192$ factored dim, projected to 4096 \citep{Kronecker} \\
Layer motif & \texttt{DDDGDDDG} (two \texttt{DDDG} repeats) \\
DeltaNet heads & 32, head dimension 128 \\
GSA heads & 16, head dimension 256 \\
Cross-stream mixing & mHC, 4 parallel streams \\
Cross-chunk memory & Memory Stream, one 4096-vector per sequence \\
Auxiliary objective & multi-token prediction at $t{+}2$, weight 0.3 \\
Integration & reversible midpoint stack (5B/9B/120B; 2B release non-reversible) \\
\bottomrule
\end{tabular}
\end{table}

These choices are constant from 2B to 120B. To say the family is grown rather than rebuilt is to say that this table is what stays fixed. One row carries a caveat. The reversible midpoint integration is used by the 5B, 9B, and 120B models. The released 2B used a non-reversible dense backbone, because at that scale the activation memory did not bind, and it did not bind partly because of the embedding. A standard table at this vocabulary and width would have occupied gigabytes of resident GPU memory; the Kronecker construction, reconstructed rather than stored, adds effectively none, which left the dense 2B fitting comfortably at the intended batch and sequence settings without paying for reversibility. The reasoning is developed in Sections~\ref{sec:system-other} and~\ref{sec:throughput-kronecker}. A reversible 2B variant existed and informed the growth path, but the released 2B checkpoint did not use it. Section~\ref{sec:reversibility} covers why.

\subsection{Three recurrences at three scales}
\label{sec:system-recurrences}

The architectural identity of LightningLM is that it carries state at three different granularities at once. They are independent mechanisms, and naming them separately is the clearest way to understand the model.

\paragraph{Token-level: gated DeltaNet.} Each D-layer maintains a per-head matrix-valued state that evolves token by token. Writing the state as $S_t$, with $q_t, k_t, v_t$ the head-wise projections of the residual stream, a per-token decay gate $\alpha_t$, and a per-token write gate $\beta_t$, the recurrence is
\[
S_t = \alpha_t \left( S_{t-1} - \beta_t \, S_{t-1} k_t k_t^\top \right) + \beta_t \, v_t k_t^\top, \qquad y_t = S_t q_t.
\]
This is the delta-rule linear transformer of Schlag and colleagues \citep{DeltaNet}, in the gated, hardware-parallel formulation of Yang and colleagues \citep{GatedDeltaNet}, realized in this implementation through a chunked linear-attention kernel. The cost is linear in sequence length, an $O(T)$ scan rather than the $O(T^2)$ of full attention, which is what lets the backbone run long sequences cheaply. The state is carried within a sequence, within a layer.

\paragraph{Cross-stream: mHC.} The cross-stream mechanism is Manifold-Constrained Hyper-Connections, introduced by Xie and colleagues at DeepSeek \citep{mHC}, itself a constrained form of the Hyper-Connections residual scheme \citep{HyperConnections}. It is used here as published, not invented, and adopted for the reason the originating work demonstrates: in deep residual stacks the repeated stream mixing can compound into severe signal amplification across layers, and constraining the mix to the doubly-stochastic manifold is what keeps that signal stable as depth grows. That motivation is directly relevant to a model meant to be grown deeper over its lineage, and the DeepSeek work is credited for it. The mechanism is described here because it is one of the three recurrences the family preserves through growth. Every layer is wrapped so the single residual stream of width 4096 is expanded into four parallel sibling streams, mixed, and collapsed back. The mixing is the part worth stating in detail. Each token produces, from its own content, a four-by-four score matrix, which is passed through Sinkhorn-Knopp normalization \citep{SinkhornKnopp} to a doubly-stochastic mixing matrix, one whose rows and columns each sum to one. Sinkhorn-Knopp reaches that doubly-stochastic form by alternately normalizing rows and columns, a procedure from 1967; a small fixed number of iterations is enough at this size, and the cost is negligible against the layer's other work because the matrix is four-by-four per token rather than a function of sequence length. The doubly-stochastic constraint is the point of the mechanism. It makes the mix a conservative redistribution of information across the four streams, neither inventing mass nor destroying it, so the streams can specialize and still exchange content at every layer without one stream washing the others out. After the layer's computation, the four streams are collapsed back to the single residual width. This is the layer-scale recurrence, a four-by-four content-dependent mix carried across the streams within a layer.

\paragraph{Cross-chunk: the Memory Stream.} The third recurrence is what lets the model carry information across the chunks of a long document. After the current chunk is embedded but before stream expansion, the model injects a gated, layer-normalized version of a single summary vector from the previous chunk:
\[
x_{\text{inject}} = x + \lambda_r \cdot g_t \cdot \mathrm{LN}(m_{\text{prev}}),
\]
where $m_{\text{prev}}$ is the previous chunk's summary, $\lambda_r$ is a single learned scalar initialized small, around 0.078 via a softplus on a raw value of $-2.5$, and $g_t$ is a per-token content gate in $[0,1]$. After the layer stack collapses the streams and applies the final normalization, the model writes the next summary as the last token's collapsed hidden state, detached from the gradient:
\[
m_{\text{out}} = \mathrm{stop\_grad}\!\left( h_{\text{main}}[:, T-1, :] \right).
\]

Three properties make this design deliberate rather than incidental. It blocks nothing on the forward pass, because the summary is detached and there is no autograd path between chunks, so the current chunk stays fully parallel in its length. It carries $O(1)$ state between chunks, one 4096-vector per sequence. And it is content-aware, because the per-token gate and the layer-norm together prevent the degenerate solutions of always injecting the full memory or never injecting it.

The trained values at the end of 2B training, read directly from the released checkpoint, show the mechanism settling into a non-degenerate regime rather than collapsing. The injection scalar grew from its initialization near 0.079 to about 0.391, a fivefold increase, so the model learned to rely on cross-chunk memory more than it started out doing. The content gate has no bias offset, so it sits at one half before content shifts it, and the gate projection is small and zero-mean. The memory layer-norm scale settled to a near-constant 0.295. Multiplying these through, the average additive injection magnitude is on the order of $0.391 \times 0.5 \times 0.295 \approx 0.058$, so roughly six percent of the current chunk's embedding is added from the previous chunk's summary on average. The injection is additive rather than an overwrite, and the magnitude is neither saturating nor dead.

The combined picture is the recurrence ledger.

\begin{table}[h]
\centering
\caption{The three-scale recurrence ledger preserved unchanged across every grown stage of the LightningLM family.}
\label{tab:recurrence-ledger}
\begin{tabular}{llll}
\toprule
Scale & Mechanism & State carried & Visible across \\
\midrule
Token & Gated DeltaNet matrix state & per-head matrix, per layer & within one sequence \\
Layer & mHC Sinkhorn doubly-stochastic mix & four-by-four per token & across 4 streams, within a layer \\
Chunk & Memory Stream summary & one 4096-vector per sequence & across chunks of a document \\
\bottomrule
\end{tabular}
\end{table}

This three-tier structure is preserved across every larger model. The mechanism and the dimensional contract are carried forward unchanged, even though the code paths and configurations differ from stage to stage. The architectural commitment of the family is precisely this ledger.

\subsection{The other backbone components}
\label{sec:system-other}

Three more components complete the backbone and are also carried forward unchanged.

The G-layers run a learnable top-$k$ sparse attention with a per-token, variance-adaptive budget. A small indexer produces match scores between queries and keys, and the number of keys each query attends to is set per token by how variable that token's match scores are relative to a running average:
\[
k(t) = \mathrm{clip}\!\left( k_{\text{base}} \cdot \frac{\mathrm{var}(t)}{\mathrm{var}_{\text{EMA}}},\; k_{\text{min}},\; k_{\text{max}} \right),\quad k_{\text{base}}=128,\; k_{\text{min}}=32,\; k_{\text{max}}=256.
\]
Here $\mathrm{var}(t)$ is the per-token variance of the indexer match-logits and $\mathrm{var}_{\text{EMA}}$ is a running exponential moving average of that variance, updated each step with momentum 0.99. The rule is content-adaptive rather than length-adaptive. A token whose indexer variance is high relative to the running average attends to more keys, a token with low variance attends to fewer, and the EMA in the denominator keeps the long-run average budget near $k_{\text{base}}$. At the 4096 training context this means an easy token attends to about 0.8 percent of the keys and the hardest tokens are capped at about 6 percent. The upper cap is a hardware decision rather than an architectural one. $k_{\text{max}}$ was reduced from its natural value of 1024 down to 256 to bound atomic-add contention in the backward kernel, trading a little expressivity for a faster, less contended gradient computation. The running-variance EMA is the one piece of layer-local state that persists between minibatches, and it is snapshotted at the start of each reversible block so the backward recomputation stays consistent, the detail that connects to the reversibility discipline of Section~\ref{sec:reversibility}.

The multi-token-prediction (MTP) head, following \citet{MTP}, predicts the token at $t{+}2$ from a fusion of the backbone hidden state at $t$ and the embedding of the ground-truth token at $t{+}1$. The choice of $t{+}2$ rather than the more usual $t{+}1$ auxiliary target introduced by that work is deliberate. A $t{+}1$ auxiliary objective largely echoes the main next-token loss and adds little gradient the main path does not already supply, whereas predicting two ahead forces the backbone hidden state at $t$ to carry information about a token it does not yet directly condition on, which is a genuinely different and more demanding signal. The head uses sparse attention rather than DeltaNet, because it runs once per step and the higher-quality gradient is worth the quadratic cost at that single use. Its output feeds the same language-model head as the main path, and it enters the loss at weight 0.3. The block is present in the 2B, 5B, and 9B stages and removed at 120B, a simplification Section~\ref{sec:system-stages} returns to.

The embedding is the Kronecker construction that Section~\ref{sec:throughput-kronecker} returns to from the systems side, and the accounting is worth setting down here because it matters far more at the small end of the family than a first glance suggests \citep{Kronecker}. A standard embedding table at this vocabulary and width, $131{,}072 \times 4096$, would be roughly 537M trainable parameters. The Kronecker construction replaces that table. Each token is built from frozen byte and position identity bases, of size $256 \times 256$ and $32 \times 32$, totaling 66{,}560 non-trainable values, which produce a $256 \times 32$, or 8192-dimensional, factored embedding per token. A single trainable projection, \textit{pf\_to\_model}, brings that 8192-dimensional factored embedding down to the 4096 residual width at a cost of $8192 \times 4096 \approx 33.55$M trainable parameters, and that projection is the only trained parameter in the embedding path. The bases themselves never train. The trainable-parameter consequence is a reduction from roughly 537M to about 33.55M on the input side. Against the 120B that difference rounds away. Against the 2B dense seed it is close to a quarter of the entire model, so the active-parameter counts this paper reports are counts the Kronecker construction made small enough to fit and train at the batch sizes that give good throughput. The construction is introduced here as architecture. The separate and equally consequential saving, that the full reconstruction table never has to be resident in GPU memory because the embedding is computable rather than stored, is a systems result and is treated in Section~\ref{sec:throughput-kronecker}; the full method is in the companion Kronecker work \citep{Kronecker}.

The integration is the reversible midpoint stack of Section~\ref{sec:reversibility}. Architecturally it advances the residual stream by a midpoint update of the form
\[
x_{l+1} = x_l + \delta_l\!\left( x_l + \tfrac{1}{2}\delta_l(x_l) \right),
\]
constructed so the backward pass can reverse it to recover intermediate states rather than storing them. The systems consequences, the configuration, and the constraints it imposes, including the requirement that dropout be zero, are in Section~\ref{sec:reversibility}.

\subsection{Parameter ledger at 2B}
\label{sec:system-2b-params}

The 2B model is fully dense and fully active. Its parameters break down as follows.

\begin{table}[h]
\centering
\caption{Parameter breakdown of the released 2B dense seed, read directly from the shipped checkpoint.}
\label{tab:2b-params}
\begin{tabular}{lr}
\toprule
Component & Parameters \\
\midrule
Kronecker frozen basis values ($256{\times}256 + 32{\times}32 = 66{,}560$) & non-trainable \\
\texttt{pf\_to\_model} projection ($8192 \to 4096$) & 33.6M \\
6 DeltaNet attention layers & 505M \\
2 GSA attention layers & 207M \\
8 dense feed-forward layers & 201M \\
mHC routing across layers & 7M \\
MTP block & 291M \\
Memory Stream parameters & 20K \\
Language-model head & 537M \\
\midrule
Total stored & 1.78B \\
\bottomrule
\end{tabular}
\end{table}

These counts are read directly from the shipped checkpoint and reconcile to its total of $1{,}781{,}570{,}624$ parameters with no residual. Two rows deserve a note so the table cannot be misread.

Both concern rows that are larger than a naive count would predict. The GSA layers count their gating projections, not only query, key, value, and output, which is why they come to about 207M rather than the roughly 134M a four-projection count alone would give. The multi-token-prediction block is not a thin output head. It carries its own small reversible stack of attention and feed-forward, which is why it is about 291M rather than the order of 100M a single projection head would imply. The Kronecker row is the inverse case, a row that is far smaller than the table it replaced, since the 33.6M projection stands in for the roughly 537M a standard embedding table would have cost, as Section~\ref{sec:system-other} sets out. Because the 2B is dense, every one of these parameters is active on every token. The active and stored counts are equal here, and they diverge only once routing is introduced at the next stage.

The training objective combines the main next-token cross-entropy at $t{+}1$, the multi-token-prediction cross-entropy at $t{+}2$ weighted at 0.3, and, in the MoE stages, a router z-loss term:
\[
\mathcal{L} = \mathcal{L}_{\text{ntp}} + 0.3\,\mathcal{L}_{\text{mtp}} + \mathrm{MOE\_W\_Z}\cdot \mathcal{L}_z.
\]
In the dense 2B there is no routing and the z-loss term is absent.

All three MoE stages, the 5B, 9B, and 120B, use the loss-free expert-balancing scheme of \citet{LossFreeMoE} rather than a classical differentiable load-balancing loss. Routing scores are produced by a sigmoid, expert selection is biased by a per-expert routing logit bias that is adjusted to equalize load, and balance is achieved by moving that bias rather than by adding a balancing term to the gradient. This is uniform across the MoE family, and it is not a property the 120B introduced. What differs between the stages is the geometry of the bias-update controller rather than whether balancing is loss-free. The 5B and 9B use a simple sign-based update that nudges each expert's bias against the sign of its load error, while the 120B uses a three-tier controller that adds a quadratic boost when an expert's load falls outside a target band. The z-loss weight is small or zero depending on stage. The optimizer keeps separate learning-rate groups, with the routers and the mHC mixing scalars held to a small fraction of the base rate so they move slowly, while the Memory Stream parameters train at the full base rate.

The released metric streams for the MoE stages carry \textit{null\_rate} and \textit{l\_null} columns. These are dormant outputs of an earlier null-routing code path. They are not balance signals; production routing uses real experts only, with null routing off in every stage, as Section~\ref{sec:system-stages} notes.

\subsection{What the larger stages change}
\label{sec:system-stages}

Against that constant backbone, each growth stage changes a small, specific set of things. Everything not listed inherits the same mechanism and dimensional contract, including the entire three-scale recurrence ledger, though configurations and code paths differ across stages.

\begin{table}[h]
\centering
\caption{Per-stage growth operations against the constant backbone.}
\label{tab:stage-changes}
\footnotesize
\setlength{\tabcolsep}{4pt}
\begin{tabularx}{\textwidth}{l r X}
\toprule
Stage & Layers & What changes from the stage below \\
\midrule
2B dense & 8 & the seed; dense feed-forward, single active path per layer \\
5B MoE & 8 & feed-forward becomes MoE: one shared expert at width 2048 plus 20 routed experts at width 1024, top-2 routing, loss-free sign-based balancing \\
9B MoE & 20 & same MoE feed-forward and same loss-free balancing as 5B; backbone deepened from 8 to 20 layers by depth-mapped re-entry \\
120B MoE & 20 & routed experts expanded from 20 to 460, top-12 routing; bias-controller geometry changed from sign-based to a three-tier band controller; expert-parallel size 8; trained with the TQP strategy of 8-bit quantized expert weights and rank-16 adapters (Sections~\ref{sec:turboquant} and~\ref{sec:training-120b}) \\
\bottomrule
\end{tabularx}
\end{table}

The same family, counted by verified parameters from each shipped checkpoint, makes the growth concrete. The active-per-token count grows monotonically across all four stages even though the stored count jumps by more than an order of magnitude at the last, because the 120B expansion is in expert count rather than in the fraction of experts active on any token.

\begin{table}[h]
\centering
\caption{LightningLM family by verified parameter counts from each shipped checkpoint.}
\label{tab:family-counts}
\footnotesize
\setlength{\tabcolsep}{4pt}
\begin{tabular}{lrlrrrr}
\toprule
Stage & Layers & D:G ratio & Experts per layer & top-$k$ & Active per token & Total stored \\
\midrule
2B dense & 8 & 6:2 & dense FFN & n/a & 1.78B & 1.78B \\
5B MoE & 8 & 6:2 & 20 routed + 1 shared & 2 & 2.24B & 4.96B \\
9B MoE & 20 & 15:5 & 20 routed + 1 shared & 2 & 3.92B & 9.36B \\
120B MoE & 20 & 15:5 & 460 routed + 1 shared & 12 & 5.93B & 118.67B \\
\bottomrule
\end{tabular}
\end{table}

The three-axis story reads off this table cleanly. From 2B to 5B the change is dense feed-forward becoming a mixture of experts, with depth held at eight layers, and active parameters rise modestly. From 5B to 9B the change is depth, eight layers to twenty, with the three-to-one ratio of linear-attention to sparse-attention layers preserved. From 9B to 120B the change is expert count alone, twenty routed experts to 460 at the same depth and ratio, which multiplies stored parameters by more than twelve while keeping the active fraction small. At 120B only about five percent of the stored parameters are active on a given token. The multi-token-prediction block, present in the smaller stages, is removed at 120B as a deliberate architectural simplification at scale.

\subsubsection*{Why these depths}

The two depths in the family, eight layers at 2B and twenty from 9B onward, were not picked by eye. They follow a rule about where depth should come from. Recent analysis argues that transformers are routinely built deeper than is useful, that beyond a critical depth which grows only sublinearly with width each added layer raises loss rather than lowering it, and that capacity is better spent on width or sparsity than on stacking further layers \citep{DepthDelusion}. This work pairs that critical-depth observation with the standard parameter accounting for transformers, that the active dense path is approximately $N \approx 12Ld^2$ in the number of layers $L$ and width $d$, and applies the accounting to the quantity that matters for a sparse model, the active path a single token traverses, rather than to the total parameter count.

At width $d = 4096$, the per-layer active cost is about $12 d^2 \approx 201.3$M parameters. A dense model of roughly 1.5 to 1.6B active parameters therefore corresponds to about $1.6\text{B} / 201.3\text{M} \approx 8$ layers, which fixed the depth of the 2B seed at eight. The same reasoning applied to the 120B uses its active path rather than its 118.67B stored total. The active dense computation per token was designed to land near 4.0 to 4.5B parameters, which corresponds to about $4.5\text{B} / 201.3\text{M} \approx 22$ layers, and the family settles at twenty. Depth, in other words, was set by the effective active size a token actually sees, and the remaining capacity of the 120B was delivered through experts and routing rather than through additional sequential layers. The heuristic is borrowed. Anchoring it to the active path of a grown sparse model, so that a twelve-times jump in stored parameters changes no depth at all, is the use of it that is ours.

A point of fact about the routing, because earlier logging scaffolding can mislead. The released models route over real experts only, with no null or placeholder expert slots: 20 real experts at top-2 for the 5B and 9B, and 460 real experts at top-12 for the 120B. Some metric streams still carry null-routing diagnostics, which survived from an earlier experimental branch and are dormant in the released path. They are not evidence that null routing was used.

The growth operations that produce each stage from the one below, the partition upcycle into the first MoE, the depth-mapped re-entry, and the expert expansion, are the subject of Section~\ref{sec:growth}, and the systems work that makes each stage trainable on one node is Sections~\ref{sec:reversibility} and~\ref{sec:throughput}. The point of this section is the architecture those operations preserve. From the recurrence ledger up, the family is one model.

\section{Reversibility as the Backbone}
\label{sec:reversibility}

The central systems decision in LightningLM 0.1V was to make the scaling path reversible. The family reached the larger MoE stages not through any single parameter-growth step but because activation memory was held under control while parameters, experts, sequence length, and training batch all grew at once. Reversibility is the mechanism that kept the later growth stages within the memory budget of a single node.

The claim is stated narrowly because it governs what the evidence can support. This is a systems result, not a final-loss ablation. Matched reversible and non-reversible 120B models were not trained for a perplexity comparison, because the non-reversible configuration is the one that does not fit the single-node regime under study. The evidence concerns feasibility, stability, and throughput: whether a configuration fits, whether the runtime remains stable, and whether the 5B, 9B, and 120B models train at the intended sequence and batch settings. The argument is about what reversibility made reachable, not about a loss delta attributable to it.

Stated compactly, reversibility removes the standard layerwise activation-storage bottleneck, which allowed progressive single-node growth through 5B MoE, 9B MoE, and 120B sparse MoE. The 2B dense seed was trained non-reversibly because at that scale activation memory did not bind. Reversibility does not reduce total cost. It changes the limiting term from stored activations, which shard poorly, to parameters, experts, routing, and recomputation, which are managed through sharding, sparsity, the TQP strategy, and kernel work.

\subsection{The memory problem reversibility solves}
\label{sec:rev-memory}

In a conventional transformer stack, each layer stores its forward activations so the backward pass can compute gradients. At long sequence length this stored-activation term scales roughly as the product of batch size, sequence length, hidden size, and number of layers, and it becomes the dominant consumer of GPU memory, frequently exceeding parameters and optimizer state combined.

For this family the term was especially severe, because the scaling target extended beyond parameter count alone. Several requirements pressed on activation memory simultaneously.

\begin{table}[h]
\centering
\caption{Sources of activation-memory pressure in the LightningLM training pipeline.}
\label{tab:activation-pressure}
\begin{tabular}{ll}
\toprule
Requirement & Effect on activation memory \\
\midrule
4K and 8K training sequence lengths & Stored activations grow with sequence length \\
Depth growth from 8 to 20 layers & Standard activation storage grows with depth \\
MoE routing and expert dispatch & Adds routing state, expert activations, and balance statistics \\
Multi-token-prediction auxiliary path & Adds a second prediction path and its activations \\
Single-node execution & Caps aggregate memory at one eight-GPU node \\
\bottomrule
\end{tabular}
\end{table}

A reversible stack changes this accounting. Rather than storing every intermediate layer state, the forward pass retains a small boundary state and the backward pass reconstructs the intermediate states it requires through recomputation. The trade is between compute and memory. The backward pass spends additional compute to recover the activation memory that would otherwise scale with depth.

\begin{table}[h]
\centering
\caption{Standard versus reversible stack: where the cost moves.}
\label{tab:standard-vs-reversible}
\begin{tabularx}{\textwidth}{X X}
\toprule
Standard stack & Reversible stack \\
\midrule
Stores layer activations during forward & Reconstructs activations during backward \\
Lower recomputation cost & Higher recomputation cost \\
Stored-activation memory grows with depth & Dominant stored term no longer grows with depth in the usual way \\
Single backward path & A second recompute backward path to maintain \\
\bottomrule
\end{tabularx}
\end{table}

From 5B onward the recovered memory was worth more than the recomputation it cost, and this held at every stage above the seed. The recomputation cost is not reported as a single figure, because it depends on depth, sequence length, the fraction of the step spent on MoE dispatch rather than dense compute, and the kernel path. The throughput consequence is reported in Section~\ref{sec:throughput}, where those factors are fixed.

\paragraph{Principle.} Reversibility is warranted when the stored-activation term, rather than parameter count, is the constraint that prevents the target sequence length and batch from fitting. It exchanges a resource that does not shard, activation memory, for one that can be provisioned, compute.

\subsection{The reversible midpoint stack, and the configuration that matters}
\label{sec:rev-midpoint}

The implementation across the 5B, 9B, and 120B models is a reversible midpoint integrator, which advances the hidden state in a form the backward pass can run in reverse to recover intermediate states; the construction follows \citet{Reversibility}, who introduce the reversible midpoint stack and pose hundred-billion-parameter-class deployment as future work. The released 2B is the exception, trained non-reversibly for the reason established in Table~\ref{tab:r1-2b-feasibility}; a reversible 2B configuration exists as the feasibility experiment reported there. The multi-token-prediction head carries its own reversible sub-stack, so the auxiliary path follows the same memory discipline as the main stack.

The production integrator configuration is \textit{step\_size} $= 0.25$, \textit{a} $= 0.5$, \textit{noise\_eps} $= 0.0$, and \textit{bootstrap} $=$ \textit{euler}, identical across every reversible model in the family. One point of reproduction warrants emphasis, because it can silently consume a replicator's time. These are not the integrator's library defaults. The default construction is a near-leapfrog integrator at \textit{step\_size} $= 0.05$, \textit{a} $= 0.95$, \textit{bootstrap} $=$ \textit{no\_kick}. The production code overrides these to the more dissipative, less leapfrog-like $0.25\,/\,0.5\,/$~\textit{euler} setting. A replicator who clones the integrator at the same commit and relies on its defaults will not reproduce the configuration that trained these models; the values must be set explicitly.

\begin{table}[h]
\centering
\caption{Reversible midpoint integrator settings: production values versus library defaults.}
\label{tab:integrator-config}
\footnotesize
\setlength{\tabcolsep}{4pt}
\begin{tabularx}{\textwidth}{l l l X}
\toprule
Setting & Production value & Library default & Role \\
\midrule
Integrator & reversible midpoint & reversible midpoint & Recompute hidden states in backward \\
Step size & 0.25 & 0.05 & Larger, more dissipative step than the near-leapfrog default \\
Stabilizing coefficient & 0.5 & 0.95 & Midpoint blend \\
Noise epsilon & 0.0 & 0.0 & No injected noise; coincides with the default \\
Bootstrap & euler & no\_kick & Half-step Euler bootstrap for the recurrence \\
Dropout & 0.0 & n/a & Required for correctness, see below \\
\bottomrule
\end{tabularx}
\end{table}

Dropout is a correctness requirement rather than a tuning choice. A reversible backward pass reconstructs the forward activations by recomputation, and that reconstruction is valid only if the forward pass was deterministic. Nonzero dropout introduces random masks in the forward pass that the backward reconstruction cannot reproduce, so the recomputed activations would diverge from those the forward pass used. Dropout is therefore set to zero in every reversible model in the family, with an explanatory note to that effect in each configuration.

\paragraph{Principle.} The integrator configuration must be recorded as explicit production overrides rather than inherited from library defaults, and the forward pass must be made deterministic. A reversible stack is correct only when its forward pass can be reconstructed exactly, and removing stochasticity is a more reliable guarantee than reproducing it.

\subsection{Reversibility had to become sharding-aware}
\label{sec:rev-sharding}

A reversible implementation is not sufficient on its own, because the recompute path interacts with parameter sharding. The backward pass re-enters blocks to reconstruct their activations, and when those blocks' parameters are partitioned across GPUs under ZeRO-style sharding \citep{ZeRO}, the recompute path cannot treat them as ordinary resident tensors. The parameters must be gathered for the recomputation in the same way they were gathered for the original forward pass, or the reconstruction proceeds against incomplete weights. This applies specifically at the ZeRO stage where parameters themselves are sharded rather than only optimizer state and gradients, since that is the stage at which a block's weights are not fully resident when the backward pass recomputes against them.

The stack therefore carries a sharding-aware backward path alongside the standard reversible one, so recomputation under partitioned parameters reconstructs against correctly gathered weights rather than against the locally resident shard. Reversibility and parameter sharding are not independent features that compose without cost. The recompute path is a second site at which the model accesses its parameters, and every parameter-management scheme the forward pass depends on must be honored there as well.

\paragraph{Principle.} The recompute path must be treated as a full second forward pass for the purpose of parameter management. Whatever gathering, sharding, or offloading the forward pass requires, the backward recomputation requires identically, and a reversible stack that omits this reconstructs against incorrect weights.

\subsection{The memory evidence, and why 2B is the exception}
\label{sec:rev-2b}

The 2B model can be run in both configurations, which makes it the clearest available evidence of what reversibility buys.

\begin{table}[h]
\centering
\caption{2B memory feasibility, reversible versus non-reversible (sequence length 4096).}
\label{tab:r1-2b-feasibility}
\begin{tabular}{p{3.8cm}p{3.0cm}rp{4.5cm}}
\toprule
Configuration & Hardware & Batch size & Result \\
\midrule
2B non-reversible & 8x A100-40GB & above 8 & does not fit \\
2B reversible & 8x A100-40GB & 32 & fits, 25 to 29\,GB free per GPU \\
2B non-reversible & 8x A100-80GB & 32 & fits, \textasciitilde 73\,GB reserved \\
\bottomrule
\end{tabular}
\end{table}

The first two rows isolate the effect. On identical 40\,GB hardware, reversibility raises the feasible batch from 8 to 32 at full sequence length. The third row explains why the 2B production run could omit reversibility, which is the single case in the family where it was deliberately disabled.

On 80\,GB A100s the non-reversible 2B model fit batch 32 at 4096 with approximately 73\,GB reserved, and ran faster than the reversible path, which pays a recomputation cost the non-reversible path does not. At 2B on 80\,GB hardware the memory pressure that justifies reversibility was absent, so the configuration favored throughput. The general position is that reversibility was applied where memory binds and omitted where it does not. From 5B onward, with experts, depth, and 8K context pressing on memory simultaneously, it binds at every stage, and every stage from 5B up is reversible.

\paragraph{Principle.} Reversibility functions as a scaling path rather than a default. It is appropriate wherever the stored-activation term is the binding constraint and counterproductive where memory is not the limiting resource and recomputation is pure overhead.

\subsection{Runtime stability was its own battle}
\label{sec:rev-runtime}

A reversible stack interacts with the surrounding runtime in ways that required substantial debugging. Two episodes are recorded here because they are characteristic of the failures a recompute path introduces.

The first was a memory leak unrelated to the model. Early training showed GPU memory rising several gigabytes per step until the run terminated within a few steps. The usual candidates, offloading, persistence thresholds, bucket sizes, manual gradient clearing, and sharding-stage changes, did not affect the growth. The cause was in the framework. A single change of runtime version introduced roughly a 15\,GB-per-step allocation increase and an out-of-memory termination by step four, while the attention and linear-attention kernel versions were not implicated. The stable runtime was pinned. The general observation is that when activation memory is recomputed rather than stored, an allocator or lifecycle regression in the framework affects a reversible stack more severely than a standard one, so a step-over-step leak is more likely to originate in the runtime than in the model.

The second was specific to the recompute path at the largest scale. On the 120B runtime a crash traced to the reversible midpoint operations during forward recompute, where the recomputation path triggered a dtype assertion in the linear-attention kernel. The resolution was to switch the reversible checkpointing into reentrant mode, which routes recomputation through the older autograd reentrant path rather than the newer non-reentrant one; that path did not reach the kernel branch raising the assertion. The production 120B launch carried this setting. The recompute path is a second code path with its own failure modes and must be hardened separately from the forward pass it mirrors.

\paragraph{Principle.} A per-step leak in a reversible stack should be attributed to the runtime before the model, and the recompute path should be hardened as a distinct surface. Behaviors that remain benign in a standard stack, including framework allocator regressions and kernel dtype edge cases, manifest as leaks and crashes specifically on the path that reconstructs activations.

\subsection{What this enabled}
\label{sec:rev-enabled}

The result is the production regime on which the remainder of the paper depends. Reversibility allowed each scale to fit its intended settings on a single node. The 5B and 9B reversible MoE models trained at the intended batch and sequence settings, and the 120B model trained at 8K sequence length and batch 64 on one eight-GPU node, carrying weights that reach hundreds of gigabytes once the experts are counted. None of this regime is reachable if activation memory grows with depth in the standard manner. The reversible stack held the activation term flat while the remaining terms scaled. The companion systems optimization, reconstructing the Kronecker embedding on the fly rather than storing it, is covered in Section~\ref{sec:throughput-kronecker}; together they account for both the fit and the throughput of the larger models. Reversibility is the foundation on which the growth stages rest, and the reason a 120B sparse MoE was trainable on a single node.

\section{Growth Principles}
\label{sec:growth}

The model was not trained once at 120B. It was grown in stages from a small dense seed, each stage inheriting the trained state of the one before it. This section states the principles that made the inheritance work, at a level of detail intended to be reproducible rather than illustrative. Each transformation is described so that it can be reapplied from the text alone.

One idea runs through all of them. Growth is the preservation of learned interfaces. A trained model is a set of parts that have learned to exchange information in specific ways, and the parts that matter most during growth are the seams where one learned function hands off to the next. When those seams are preserved, the larger model begins from a state that already functions. When one is broken, the larger model must relearn what the smaller one already encoded, which removes the reason to grow it rather than train it from scratch. Each principle below corresponds to one seam, and each is paired with the failure that results from breaking it, documented in Section~\ref{sec:failure-modes}.

Every principle is given in the same form: the transformation, the mechanism in enough detail to reproduce, the alternatives tested, the principle that the case establishes, and an explicit statement of how strong the evidence is.

\subsection{Dense to MoE: preserve the function before the router specializes}
\label{sec:growth-densemoe}

\paragraph{Transformation.} The first expansion converts the trained dense 2B into the eight-layer 5B MoE. The dense feed-forward network is retained rather than discarded. It is copied into an always-active shared expert, and the routed experts are initialized from overlapping random partitions of the dense feed-forward intermediate neurons.

\paragraph{Mechanism.} The shared-expert copy is literal. The dense gate, up, and down projections are assigned directly into the shared-expert weights, so every token retains access to the exact trained dense computation regardless of the untrained router's behavior. The routed experts are then constructed by sampling neuron subsets from the dense intermediate dimension. With 20 routed experts each sampling half of a 2048-wide dense intermediate, the probability that a given dense neuron is missed by all experts is $(1 - 1024/2048)^{20}$, approximately one in a million, so the expected coverage of the dense function across the expert set is effectively complete. The experts begin as overlapping inherited subfunctions rather than as fresh random initializations. The router is the only component initialized from scratch, because the dense model carried no routing distribution to inherit. The expert weights therefore carry the trained function while the router learns assignment over it, and the shared expert maintains a stable computational floor during the period when the router is still untrained.

\paragraph{Alternatives tested.}

\begin{table}[h]
\centering
\caption{Dense-to-MoE initialization strategies considered for the 2B to 5B step.}
\label{tab:dense-moe-alts}
\footnotesize
\begin{tabularx}{\textwidth}{l X X X}
\toprule
Strategy & Mechanism & Intended benefit & Outcome \\
\midrule
Partition & Overlapping random neuron subsets copied into each expert & Maximum function preservation & Selected; the surviving production lineage \\
Drop-upcycling & Copy most neurons, reinitialize a random fraction & Break symmetry earlier & Implemented and evaluated; not the production path at this step \\
Spectral / SVD & Decompose and rotate the dense structure for explicit diversity & Diverse experts at initialization & Implemented; near-random cold loss, see Section~\ref{sec:fail-spectral} \\
\bottomrule
\end{tabularx}
\end{table}

\paragraph{Principle.} The initialization that preserves the dense function is preferable even when it leaves the experts initially correlated. Expert diversity acquired at the cost of cold-start loss is a poor trade, because specialization is a training objective the router resolves over subsequent steps, whereas function preservation is an initialization property that no later stage can recover once it is lost.

\paragraph{Evidence.} The transformation and the three strategies are reproducible from surviving code and the surviving partition checkpoints. The numeric three-way cold-start comparison exists only in the training-session transcript, since the saved comparison output was never committed to durable storage. The reproducible claim is that partition was selected and is the production path; the specific spectral and partition loss figures are reported as transcript-grade rather than as a recoverable benchmark.

\subsection{Depth growth: end on the layer that learned to be terminal}
\label{sec:growth-depth}

\paragraph{Transformation.} The second expansion takes the eight-layer 5B stack to the twenty layers of the 9B by re-entering the trained layers in a chosen order. The choice of order is the substance of the step. The target depth of twenty follows the active-path depth rule of Section~\ref{sec:system-stages}; this section concerns how the trained layers are arranged to reach it.

\paragraph{Mechanism.} A trained decoder terminates on a specific layer whose output distribution the language-model head learned to read. That terminal layer learned to be terminal, while the interior layers learned to feed the next layer rather than the head. When a deeper stack is built by repeating trained layers, the layer it ends on determines what the head receives. The production mapping is \textit{1-8, 1-4, 1-8}: the full eight-layer stack, the first four layers again, then the full eight-layer stack once more. The grown stack ends on the original terminal layer L7, so the head continues to receive the hidden-state distribution it was trained to consume. The principal alternative, \textit{1-8, 1-8, 1-4}, ends on an interior layer L3, which the head never learned to read.

\paragraph{Candidate mappings and short-run result.}

\begin{table}[h]
\centering
\caption{Candidate depth-mapping schemes for the 8 to 20 layer growth, with short-run loss.}
\label{tab:depth-mappings}
\footnotesize
\setlength{\tabcolsep}{4pt}
\begin{tabular}{lllrrrl}
\toprule
Mapping & Sequence & Terminal layer & Step 1 NTP & Step 100 NTP & Last logged & Selected \\
\midrule
reenter-early & 1-8, 1-4, 1-8 & L7 (trained terminal) & 4.544 & 3.629 & 3.738 & yes \\
full-cycle-partial & 1-8, 1-8, 1-4 & L3 (interior) & 4.519 & 3.634 & 3.657 & no \\
mid-block & mixed reentry & intermediate & 4.893 & 3.646 & 3.702 & no \\
\bottomrule
\end{tabular}
\end{table}

\paragraph{Principle.} The endpoint is the first thing to fix. A grown decoder should terminate on a layer whose output the head has already learned to consume, which for a stack assembled from repeated source layers means selecting the mapping whose final layer is the original terminal layer, absent a deliberate head-recalibration phase to repair the interface.

The case is worth stating sharply because the short-run loss does not enforce it. The interior-ending mapping had a marginally better last-logged loss, by about eight hundredths of a nat, and the terminal-ending mapping was selected regardless. When a small short-run loss delta conflicts with a known learned interface, the interface takes priority. The loss delta is local, and further training recovers it. The terminal-to-head contract is a structural interface, and training does not repair a model that ends on a layer whose output distribution the head was never trained to read, absent a deliberate recalibration phase built to do so. Concretely, the 5B language-model head was trained against the output distribution of layer 8 (zero-indexed layer 7). The reenter-early mapping preserves that contract by terminating the grown 9B stack on the same layer 8. Full-cycle-partial would have asked the language-model head to read layer 4 outputs it had never seen during 5B training. The \textit{lm\_head} had seen layer 8, not layer 4, and that is the reason reenter-early was selected over the lower-loss alternative.

\paragraph{Evidence.} The short-run sweep numbers are from saved training logs and are reproducible. One bookkeeping separation matters and is worth recording precisely. The depth-mapping sweep ran on a second machine against the step-88{,}981 5B checkpoint so the main 5B run on the primary machine could continue uninterrupted toward its public endpoint at step 101{,}000. Running the sweep on the earlier 5B candidate while the production 5B kept training was the wall-clock-saving decision. Once the mapping was selected from that parallel sweep, the same transform was applied to the final 5B checkpoint at step 101{,}000 to produce the production 9B initialization. The sweep established the principle on step 88{,}981 and the production grow applied that selected mapping to step 101{,}000. The two checkpoints are kept distinct rather than implying the sweep ran on the production source.

\subsection{Positional recalibration: DroPE, applied before annealing}
\label{sec:growth-drope}

\paragraph{Transformation.} Partway through the 9B stage, the model is moved off its main rotary position encoding and recalibrated to operate without it, following the DroPE procedure of Gelberg and colleagues \citep{DroPE}. The recalibration is performed at the original training context length and is scheduled before the low-learning-rate annealing phase rather than after it.

\paragraph{Motivation.} The reason to remove the positional encoding is long-context reach. Rotary position encodings provide a strong inductive bias that accelerates pretraining convergence, but a model that comes to rely on explicit rotary information degrades sharply once the inference sequence exceeds the length it was trained on, because the rotary signal is out of distribution at unseen positions. The established rotary-scaling methods that adapt the frequencies for longer sequences still require an expensive long-context finetuning phase to use the extra positions meaningfully. DroPE addresses this by treating the positional encoding as a transient training aid: the encoding is dropped after it has done its work during pretraining, and a short recalibration phase at the original context length adapts the model to function without it, after which the model generalizes to sequences beyond its training length without long-context finetuning \citep{DroPE}. For a pipeline whose intended trajectory is toward very long contexts, where rotary encodings are precisely the component that fails, removing the explicit positional dependence is a deliberate investment in that trajectory rather than a local tuning choice.

\paragraph{Mechanism.} Annealing crystallizes a model. By the end of a low-learning-rate phase the weights have settled into a configuration that is deliberately resistant to further movement. Dropping the positional encoding after the anneal would disrupt a model that was just made rigid, which forces a choice between recovering at the low annealed learning rate, which is slow and may not converge, and raising the rate again, which discards the annealing already paid for. Performing the DroPE recalibration first, while the model is still plastic, removes that choice. The model absorbs the disruption at a useful learning rate, and the subsequent anneal locks in the recalibrated, position-encoding-free state. Following the DroPE recipe for the higher-learning-rate recalibration regime, the recalibration applies QKNorm \citep{QKNorm} after the encoding is dropped; DroPE adopts QKNorm to stabilize the larger gradients that arise once the positional bias is removed.

\paragraph{Principle.} A DroPE-style removal of the positional encoding belongs before the anneal, while the model retains plasticity, with the anneal applied to the recalibrated state. The recalibration is performed at the original context length, consistent with the DroPE procedure, and the long-context generalization it targets is realized at inference rather than paid for through long-context finetuning.

\begin{figure}[t]
\centering
\includegraphics[width=0.92\textwidth]{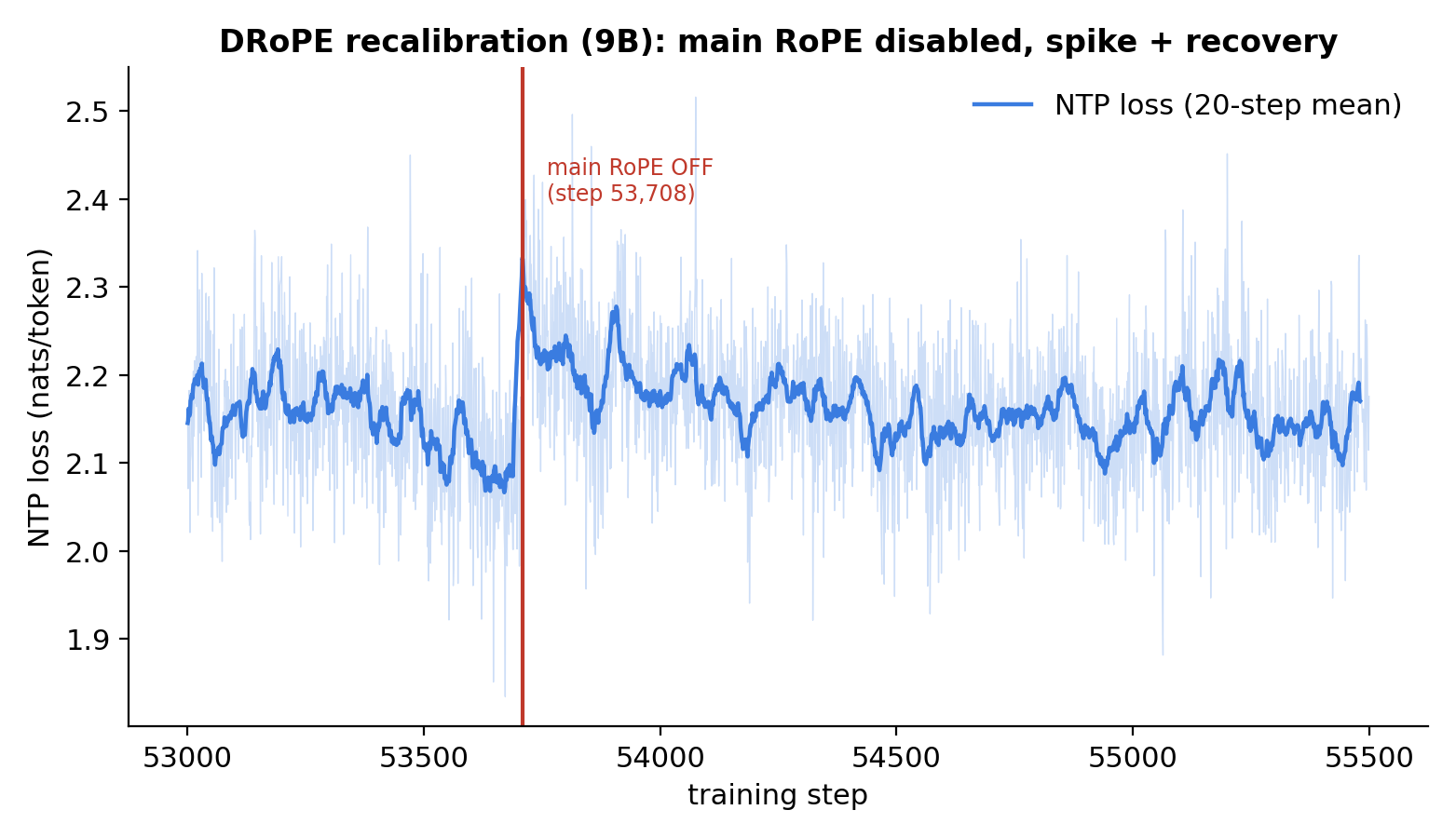}
\caption{DRoPE recalibration in the 9B run: the main rotary encoding is disabled at step 53{,}708, producing a brief loss spike that recovers within a short recalibration window at the original context length.}
\label{fig:drope}
\end{figure}

\paragraph{Evidence and scope.} The transition is in a saved per-step loss log and is shown in Figure~\ref{fig:drope}. The observed cost of dropping the main positional encoding was a loss increase of about 0.12 nats that recovered quickly, consistent with the DroPE finding that a recalibrated model returns to its pre-removal perplexity within a short recalibration window, and that the removal is most effective when performed late in training rather than early \citep{DroPE}. The original work reports that dropping the encoding at the very end of pretraining matches the perplexity of the unmodified model while dropping it earlier degrades it, which is direct external support for the ordering argument above. Two qualifications attach. First, DroPE was introduced and validated on standard RoPE transformers up to 7B parameters, whereas the backbone here is a DeltaNet and sparse-attention recurrence hybrid, so this is a replication of the DroPE procedure in a different architectural setting rather than a reproduction of the original experiments, and the clean loss recovery is the evidence that the procedure transfers to this architecture. Second, the long-context reach is a property the recalibration is designed to enable, reported here as theoretical for this model rather than measured. In the original work DroPE retains needle-in-a-haystack performance at two, four, and eight times the training context where rotary-scaling methods collapse, scores on multi-hop extraction into context buckets of eight to sixteen times the training length, and is evaluated on knowledge-extraction problems exceeding eighty times the pretraining context \citep{DroPE}. Applied to an 8K-trained model, the cleanly demonstrated eight-times factor corresponds to roughly 64K tokens and the largest extraction factor to roughly 640K, which indicates the trajectory the position-encoding-free state is intended to support. The production models here were trained and evaluated at 4K and 8K, the longer reaches were not tested in this work, and the contribution claimed for this step is the successful in-architecture replication and the resulting position-encoding-free state.

\subsection{Keyspace-first checkpoint construction: let the target model drive the transfer}
\label{sec:growth-keyspace}

\paragraph{Transformation.} The checkpoint that the grown model loads is constructed in the target model's own parameter namespace rather than the source model's.

\paragraph{Failure mode it prevents.} The natural way to write a growth converter walks the source weights, transforms them, and writes them under the names they held in the source. If the target model names the parameters that carry its computation differently, a non-strict load silently skips the mismatched names and leaves those parameters at their random initialization while reporting a successful load. This is more dangerous than a crash, because the model appears loaded, has the correct parameter count and checkpoint size, and trains, while the parameters that carry the function remain partly random. The cost is documented in Section~\ref{sec:fail-misload}.

\paragraph{Correct method.} The direction of the transfer is reversed.

\begin{enumerate}
\item Instantiate the target model.
\item Walk the target model's own parameter and buffer names.
\item For each target name, find or construct the source tensor that belongs there, applying the growth transform at that point.
\item Build the target model's state dictionary from that walk.
\item Require a strict load to succeed, so any unfilled or misnamed parameter becomes a construction-time error rather than a silent runtime fallback.
\end{enumerate}

The same builder applies the expert-expansion transform of Section~\ref{sec:growth-experts}, with the expansion ratio and router perturbation passed as explicit arguments and a fixed seed for reproducibility.

\paragraph{Principle.} A grown checkpoint should not be trained from or released unless the converter is target-keyspace driven and the target model loads it strictly. Key compatibility is part of the learned-function transfer rather than incidental metadata, and a permissive loader conceals precisely the failure that most needs to be visible.

\paragraph{Evidence.} The builder and its invocation are reproducible, and the strict-load requirement is the verification. The released 120B lineage was built with the corrected target-keyspace initializer and trained through the full path before consolidation, as detailed in Section~\ref{sec:fail-misload}; the broken-load case described there was a predecessor conversion failure rather than the release.

\subsection{Expert expansion: clone the function, perturb the symmetry, protect the differentiation}
\label{sec:growth-experts}

\paragraph{Transformation.} The final expansion takes the twenty routed experts of the 9B to the 460 of the 120B. Backbone depth and width stay fixed, and the added capacity comes entirely from expert count.

\paragraph{Mechanism.} Each of the twenty source experts is cloned into twenty-three variants, reaching 460. The clones are produced by drop-upcycling \citep{DropUpcycling}: each clone retains most of its source expert's weights while a random fraction of intermediate positions, half in this configuration, is reinitialized to break exact symmetry, with the same mask applied across the gate, up, and down projections of a given expert. The router is tiled from twenty to 460 entries and perturbed with a small noise term. The intent matches the dense-to-MoE step, which is to preserve the trained function, introduce enough perturbation that the copies are not identical, and let training specialize them.

\paragraph{Why cloning alone is insufficient.} Twenty-three near-identical clones of each source expert give the router almost no early signal for distinguishing same-family clones. Under hard top-$k$ routing it does not separate them: it selects a representative per family and starves the remainder, and the nominal capacity for 460 experts collapses back toward the original twenty. The expansion brings the model to a workable initialization, but keeping the expanded experts alive long enough to differentiate is a routing problem, addressed with probabilistic selection during the early differentiation window, and it must be addressed alongside the initialization rather than after it. The collapse trajectory and the routing fix are in Section~\ref{sec:fail-clone-collapse}.

\paragraph{Principle.} Expert expansion should clone and perturb trained experts rather than adding random ones, and the expansion should be paired with a routing scheme that protects near-duplicate experts through the differentiation window. The initialization preserves the function and the routing scheme preserves the capacity, and the expansion fails to deliver its nominal capacity if either is omitted.

\paragraph{Evidence.} The expansion transform is reproducible from the builder and its arguments. The collapse-and-fix comparison is partly transcript-grade and partly from saved balance logs, and Section~\ref{sec:fail-clone-collapse} marks which is which. The released path used drop-upcycling rather than the clone-based variant that produced the cleanest collapse demonstration, and that distinction is stated explicitly rather than presenting one curve as the other.

\subsection{The common thread}
\label{sec:growth-common}

The five steps are a single idea applied at five seams. The dense-to-MoE step carries the feed-forward function forward into the shared and routed experts. The depth step carries the terminal-layer-to-head interface forward by ending on the trained terminal layer. The positional step carries plasticity forward to the moment the disruption requires it. The keyspace-first step carries forward the guarantee that every parameter is actually transferred rather than silently left random. The expert-expansion step carries the expert function forward while protecting its differentiation under routing. In every case the failure has the same shape: a transformation that appeared locally reasonable broke a learned relationship, and the larger model paid by relearning what the smaller one already encoded. Growth succeeds when the trained model is treated as a set of interfaces to carry forward, and fails, quietly and at high cost, when it is treated as an undifferentiated set of weights to reshape. The systems conditions that make these transformations affordable on a single node, the reversible backbone and the on-the-fly embedding reconstruction, are in Sections~\ref{sec:reversibility} and~\ref{sec:throughput}; the subject of this section is the transfer, and the transfer is governed by interfaces.

\subsection{Summary}
\label{sec:growth-summary}

The five growth steps and the one systems enabler that supports them are collected in the following table.

\begin{table}[h]
\centering
\caption{Growth steps and the one systems enabler that supports them, each tied to a preserved interface and a principle.}
\label{tab:growth-summary}
\footnotesize
\setlength{\tabcolsep}{3pt}
\begin{tabularx}{\textwidth}{X X X X X}
\toprule
Stage & Growth operation & Preserved interface & Main risk & Principle \\
\midrule
2B dense to 5B MoE & Dense FFN to shared expert, routed experts as random partitions & Dense feed-forward function & Experts start too random; router untrained & Copy the function first, specialize later \\
5B to 9B, 8 to 20 layers & Re-enter trained layers in order & Terminal-layer to head hidden-state contract & Ending on an interior layer & End on the trained terminal layer \\
9B positional change & DroPE recalibration before the final anneal & Plasticity during the representational change & Disrupting after the anneal & Disrupt before the anneal, then anneal the recovered state \\
9B to 120B checkpoint & Target-keyspace construction with drop-upcycling & Every target parameter actually populated & Silent random fallback on key mismatch & Instantiate the target and fill its keys strictly \\
20 to 460 experts & Clone and perturb source experts & Expert feed-forward function & Clone-family collapse under top-$k$ & Preserve expert function, then protect routing diversity \\
Kronecker embeddings (Section~\ref{sec:throughput-kronecker}) & On-the-fly reconstruction instead of a resident table & Embedding semantics & GPU memory pressure & Reconstruct on the fly when the memory saved exceeds the compute cost \\
\bottomrule
\end{tabularx}
\end{table}

The final row is not a growth step. It is the systems optimization that makes the others affordable on a single node, included here because it follows the same reasoning: preserve the property that matters, the embedding semantics, while changing how it is realized in memory.

\section{Single-Node Throughput}
\label{sec:throughput}

The preceding sections describe how the models were made to fit on one node. Fitting is necessary but does not on its own make a model trainable. A model that fits but trains too slowly is not trainable in any practical sense, and the early versions of this stack were in that state. At the 2B scale the system trained in the high teens of thousands of tokens per second, the 5B MoE managed roughly 12k in early sweeps, and the 9B MoE only a few thousand. The production stack reached roughly 59k ordinary tokens per second at 2B, roughly 37k at 5B, and made the 120B sparse MoE trainable at roughly 17k to 18k ordinary tokens per second on a single eight-GPU node. This section describes how that gap was closed and how the numbers should be read.

Three kinds of throughput evidence appear in this section, and they are not interchangeable. Early runtime sweeps, taken before the kernel and automated-search work, are the starting floor. The automated kernel and configuration search is the middle stage. Production throughput, measured from the actual run logs, is the result. A benchmark sweep, a production run, and an attempted configuration that ran out of memory are three different categories of measurement, and mixing them produces the conflicting numbers this section exists to prevent.

A further distinction separates two throughput fields in the logs, because they measure different things.

\begin{table}[h]
\centering
\caption{Ordinary versus wall-clock throughput as recorded in the run logs.}
\label{tab:throughput-fields}
\begin{tabular}{p{2.8cm}p{6.5cm}p{4.5cm}}
\toprule
Field & What it measures & Purpose \\
\midrule
Ordinary throughput & Speed of the training step itself on the GPU, with no surrounding overhead & Judging the compute path: kernels, model, runtime \\
Wall-clock throughput & End-to-end speed over a real run, including checkpoint saves, logging, data loading, and all between-step cost & Budget and elapsed-time accounting \\
\bottomrule
\end{tabular}
\end{table}

These are not one number reported two ways. Ordinary throughput measures the speed of the training step itself. Wall-clock throughput measures the speed of the whole system over a real run, including everything incurred between useful steps. Wall-clock throughput is therefore always at or below ordinary throughput, at every stage, because that overhead is always present even when it is small. At the one stage where the OPUS selector was active, its scoring time is added on top of the ordinary overhead, which widens the gap between the two fields considerably. The ordinary column is the measurement to use when judging the compute path, and the wall-clock column is the measurement to use when budgeting machine time.

\subsection{The pre-kernel baseline}
\label{sec:throughput-baseline}

The first throughput sweep was a systems-tuning sweep on a node of eight 40\,GB A100s, at batch size 8 and sequence length 4096 for representative 2B, 5B, and 9B tests. It predates the kernel and automated-search work, so these figures are the floor rather than the system's capability.

\begin{table}[h]
\centering
\caption{Table P1. Early throughput before kernel and configuration search.}
\label{tab:p1-baseline}
\footnotesize
\setlength{\tabcolsep}{4pt}
\begin{tabular}{lp{2.4cm}rrrlrr}
\toprule
Model & Hardware & Seq & Batch & Baseline & Best tuned & Throughput & Delta \\
\midrule
2B non-reversible & 8x A100-40GB & 4096 & 8 & 16{,}284 tok/s & live-params capped & 17{,}409 tok/s & +6.9\% \\
5B MoE            & 8x A100-40GB & 4096 & 8 & 12{,}114 tok/s & baseline & 12{,}114 tok/s & 0.0\% \\
9B MoE            & 8x A100-40GB & 4096 & 8 & 3{,}535 tok/s & combined defaults & 3{,}644 tok/s & +3.1\% \\
\bottomrule
\end{tabular}
\end{table}

The sweep ruled out several tempting changes. Larger allocator bucket sizes hurt. Standard graph-capture compilation did nothing, because the dominant operations were already custom attention and linear-attention kernels with no PyTorch graph to capture. One collective-communication setting crashed. Disabling the memory-hardening flags caused large regressions rather than gains. What survived was narrow, mostly limiting how many parameters the sharding kept live at once to cut repeated gather overhead without destabilizing memory.

The 5B row is diagnostic. It barely moved under runtime tuning, which is the signature of a configuration that is already compute-bound rather than launch-bound or memory-bound. No amount of runtime configuration changes the compute, so the next gains had to come from the compute path itself: fused loss, the attention and linear-attention kernels, grouped expert computation, and the embedding memory pressure addressed in Section~\ref{sec:throughput-kronecker}.

\paragraph{Principle.} When runtime tuning stops moving a number, the bottleneck has moved to the compute, and further effort belongs in the kernels rather than the runtime. A flat response to ZeRO and bucket settings is the diagnostic signal that this point has been reached.

\subsection{Automated optimization: search for loss-stable speed}
\label{sec:throughput-search}

The gains that mattered came from two coupled efforts. The compute path was rebuilt with fused and specialized kernels, most importantly a fused cross-entropy that avoids materializing full-vocabulary logits during training, along with custom linear-attention paths, grouped MoE expert computation, and fused normalization. On top of that, an automated configuration sweep at 2B scale scored each run on loss, memory, and throughput jointly rather than on speed alone.

\begin{table}[h]
\centering
\caption{Table P2. Selected rows from the 2B automated configuration sweep.}
\label{tab:p2-sweep}
\footnotesize
\setlength{\tabcolsep}{4pt}
\begin{tabular}{lrrlp{4.5cm}}
\toprule
Run & Loss at step 145 & Throughput & Kept & Interpretation \\
\midrule
baseline           & 7.0009  & 38{,}179 tok/s & yes & ZeRO-1 AdamW baseline \\
softcap            & 6.6803  & 48{,}726 tok/s & yes & logit soft-cap improved loss and speed \\
fused softcap      & 6.6818  & 50{,}604 tok/s & yes & recovered speed and memory, loss held \\
softcap + beta2    & 6.6337  & 50{,}858 tok/s & yes & better loss at similar speed \\
kernel fused-size  & 6.6181  & 51{,}023 tok/s & yes & best balance at that point \\
rope-base          & 6.6127  & 51{,}730 tok/s & yes & new best loss in this slice \\
fast-but-worse     & 6.6383  & 51{,}900 tok/s & no  & fastest, but loss regressed \\
bad-init           & 10.9079 & high           & no  & initialization change was catastrophic \\
\bottomrule
\end{tabular}
\end{table}

The rejection pattern carries the lesson. The single fastest run in the sweep was not kept, because its loss regressed against slightly slower configurations. A pure throughput objective would have shipped it and quietly cost model quality. The objective used here treated throughput as worth pursuing only after a run was already loss-stable and memory-stable.

\paragraph{Principle.} Throughput is a constraint rather than the optimization objective. Loss and stability are optimized first, and the fastest configuration is then selected from among those already established as good. An automated search pointed at speed alone will find a fast bad model and rank it first.

The sweep moved the 2B harness from roughly 38k to above 51k tokens per second while improving loss. With the runtime fixes and the checkpoint handling of Section~\ref{sec:throughput-survive} added, the 2B production run reached roughly 59k ordinary tokens per second.

\subsection{Throughput has to survive the real run}
\label{sec:throughput-survive}

Single-node training is an operations problem as much as a kernel problem. The system has to checkpoint and upload under spot-instance risk without stalling useful training, and an early version of that machinery did the opposite.

The first checkpoint implementation uploaded to durable storage from a background thread. In practice the upload thread's work interfered with the training process badly enough that throughput during an upload collapsed from the high fifties of thousands of tokens per second to roughly 4k. The model was effectively pausing to upload. The fix moved the upload work off the training-critical path, after which throughput during upload returned to roughly 56k to 57k, with only occasional single-step dips.

This belongs in a throughput section because usable throughput is not the best isolated step. It is the throughput of the whole system under checkpointing, logging, spot handling, and upload pressure, sustained over a real run. The production numbers below are the numbers under that load rather than idealized step timings.

\paragraph{Principle.} Throughput should be measured under the full operational load, including checkpoint upload, rather than on a clean step. Upload and other I/O belong off the training-critical path, because a background thread sharing the training process can cost an order of magnitude during the window it runs.

\subsection{Production throughput}
\label{sec:throughput-production}

Table~\ref{tab:p3-production} reports the production throughput, with ordinary and wall-clock throughput separated. Ordinary throughput is the speed of the training step itself on the GPU. Wall-clock throughput is the speed of the whole system over a real run, including checkpoint saves, logging, data loading, and every cost incurred between useful steps, and is therefore always at or below ordinary throughput. The per-GPU column divides wall-clock throughput by eight.

\begin{table}[h]
\centering
\caption{Table P3. Production throughput by stage. Ordinary is the per-step GPU speed; wall-clock includes all between-step overhead, and at 5B additionally includes OPUS scoring.}
\label{tab:p3-production}
\footnotesize
\setlength{\tabcolsep}{3pt}
\begin{tabular}{lp{2.4cm}rrrrrr}
\toprule
Run & Hardware & Seq & GBS & Tokens/step & Wall tok/s & Ordinary tok/s & Per-GPU wall \\
\midrule
2B production    & 8x A100-80GB     & 4096 & 32  & 131{,}072 & 58{,}573         & 59{,}154         & \textasciitilde 7{,}322 \\
5B production    & 8x A100-80GB     & 4096 & 32  & 131{,}072 & 30{,}429         & 36{,}737         & \textasciitilde 3{,}804 \\
9B 4K recovery   & 8x H200          & 4096 & 120 & 491{,}520 & \textasciitilde 28k--30k & \textasciitilde 28k--30k & \textasciitilde 3.55k--3.75k \\
9B 8K stable     & 8x H200          & 8192 & 56  & 458{,}752 & \textasciitilde 27{,}700 & \textasciitilde 27{,}700 & \textasciitilde 3{,}460 \\
120B Stage 1     & 8x B200/B300     & 8192 & 64  & 524{,}288 & 17{,}575         & 17{,}982         & \textasciitilde 2{,}197 \\
120B continuation & 8x B200/B300    & 8192 & 64  & 524{,}288 & 16{,}966         & 17{,}136         & \textasciitilde 2{,}121 \\
\bottomrule
\end{tabular}
\end{table}

Every row shows ordinary throughput at or above wall-clock, because between-step overhead is always present. In most rows the gap is small and reflects only that overhead: the 2B row shows 59{,}154 ordinary against 58{,}573 wall-clock, and the 9B rows show a similarly narrow separation, all of it ordinary operational cost with no data selector involved. The 5B row is the exception, at 36{,}737 ordinary against 30{,}429 wall-clock. That roughly 6k difference is much larger than between-step overhead accounts for, because the OPUS data selector was active at 5B and its scoring time falls outside the training step. The ordinary column is the measurement that reflects the compute path, and the wall-clock column is the measurement that reflects elapsed machine time.

The OPUS selector ran at the 5B stage only. The reasons it was disabled at 2B and at 9B differ from each other and are recorded in the data methodology of Section~\ref{sec:data}. In throughput terms, its absence is why the 2B and 9B rows show ordinary and wall-clock nearly coinciding, and its presence is the entire 6k gap at 5B.

The size of that gap, and the engineering behind it, are worth stating because the reported cost of OPUS reflects a heavily optimized implementation. The originating work reports an additional compute overhead of about 4.7 percent, and that figure is an empirical end-to-end measurement of the complete pipeline, covering proxy gradient estimation, candidate scoring, CountSketch projection, and sampling. The figure is small relative to the naive alternative rather than relative to the training step. A naive gradient-based selector performs separate forward and backward passes for every candidate sequence, which drives overhead into the range of tens to hundreds of percent depending on candidate buffer size and model scale. The originating work's own efficiency analysis shows this directly: a direct online-selection implementation took 6{,}875 minutes against 1{,}985 for random sampling, roughly a 3.5-times slowdown, which the Ghost technique and CountSketch projection reduce to 2{,}083 minutes, the reported single-digit overhead. Two optimizations produce that reduction. Candidate utility is estimated from 512-token slices rather than full training contexts, and the Ghost technique extracts per-sample gradient statistics from a shared forward and backward pass without materializing full per-sample gradients, while CountSketch compresses the gradient representations into a low-dimensional sketch for efficient similarity computation. The OPUS contribution is therefore the reduction of dynamic-selection overhead from potentially hundreds of percent to single digits while preserving ranking quality, rather than the utility function alone.

The reason this did not translate into a small overhead in the present pipeline is that the machinery responsible for the reduction was itself costly to build and to run at this scale and in this configuration. The shared-pass Ghost extraction, the CountSketch projection at sketch dimension 8192, the construction of the optimizer-induced preconditioner each step, and the isotropic approximation of the Hessian interaction term all had to be implemented as custom kernels to be tractable here. The selector also forces the model to switch between training mode and scoring mode and back at every selection. The originating work runs in synchronous data-parallel training with reduce-scatter gradient synchronization and does not use ZeRO-style parameter or optimizer-state sharding, so that mode switching never had to coexist with sharded parameters. In this pipeline it did, and that integration was a substantial part of the real cost.

The strategy that kept the cost to the observed 6k gap was to amortize each scoring pass across several training steps. Rather than scoring at every step, the system fits the largest candidate batch it can at the 512-token scoring length and selects from it enough high-utility samples to supply the next ten training steps, so a single OPUS pass is amortized over ten steps of training. The roughly 6k wall-clock gap at 5B is the cost of that amortized scoring, and without the amortization the per-step cost would have been far higher.

These rows are not a clean size-only comparison. They differ in hardware generation, sequence length, batch, optimizer path, expert count, selector status, and whether the model is dense, MoE, or adapter-trained sparse MoE. The 2B row measures the throughput-engineering improvement at small scale. The 120B rows support a different claim, made precise in Section~\ref{sec:throughput-120b}, that the sparse 120B path ran at practical wall-clock speed on one node.

The 9B rows need one explicit correction, because an earlier evidence table mixed in an invalid configuration.

\begin{table}[h]
\centering
\caption{Table P4. Corrected 9B H200 throughput at 4K and 8K context.}
\label{tab:p4-9b-corrected}
\footnotesize
\setlength{\tabcolsep}{4pt}
\begin{tabular}{lrrrrl}
\toprule
Setting & Seq & GBS & Tokens/step & Throughput & Status \\
\midrule
4K recovery & 4096 & 120 & 491{,}520    & \textasciitilde 28k--30k tok/s & valid \\
8K stable   & 8192 & 56  & 458{,}752    & \textasciitilde 27{,}700 tok/s & valid \\
8K sweep    & 8192 & 64  & 524{,}288    & \textasciitilde 29{,}400 tok/s & valid benchmark \\
8K sweep    & 8192 & 128 & 1{,}048{,}576 & out of memory & excluded \\
\bottomrule
\end{tabular}
\end{table}

The 8K phase should not be described as a 54k-tokens-per-second phase. The batch-128 8K run ran out of memory and is excluded. Stating this plainly also forecloses a wrong inference, that moving to 8K context or disabling the main positional encoding somehow doubled throughput. It did neither. The stable 8K configuration before the positional change ran with the main encoding on, and its valid throughput was close to the 4K phase, as the table shows.

\subsection{On-the-fly Kronecker embeddings}
\label{sec:throughput-kronecker}

One memory-and-throughput optimization is specific enough to this architecture to warrant its own treatment. The model uses the Kronecker-structured token embeddings introduced in Section~\ref{sec:system} rather than an ordinary lookup table. Section~\ref{sec:system-other} covers the trainable-parameter consequence of that choice, the reduction of the input embedding from roughly 537M trained parameters to about 33.55M. This section covers the separate resident-memory consequence, which is a property of how the embedding is computed at runtime rather than of how many parameters it carries.

The naive way to use the Kronecker embedding materializes the full reconstruction table on the GPU. For a 131{,}072-token vocabulary and an 8192-wide reconstruction dimension that table costs roughly 2.15\,GB per GPU in bf16, about 17.2\,GB across an eight-GPU node. That table is never stored. Because the embedding is a deterministic function of a token's bytes and positions rather than a stored lookup, each token's vector can be rebuilt from a compact description of the token, so the resident state is only the token bytes and lengths, a few megabytes in total, and the vectors are reconstructed on the fly each microbatch with a scatter-and-normalize before projection into hidden space. Resident embedding memory drops from about 2.15\,GB per GPU to about 4.5\,MB. The reconstruction adds roughly one to four milliseconds per microbatch, about 0.01 to 0.24 percent of step time. This on-the-fly route is available specifically because the embedding is computable rather than a lookup, a property a standard learned table does not have.

\begin{table}[h]
\centering
\caption{Table P5. Precomputed Kronecker table versus on-the-fly reconstruction (batch 8, seq 4096).}
\label{tab:p5-kronecker}
\footnotesize
\setlength{\tabcolsep}{4pt}
\begin{tabular}{lrrrr}
\toprule
Model & Precomputed table & On-the-fly & Throughput change & Memory saved \\
\midrule
2B non-reversible  & 20{,}233 tok/s & 20{,}026 tok/s & $-1.0\%$  & 2.1 GB/GPU \\
5B MoE reversible  & 10{,}185 tok/s & 12{,}752 tok/s & $+25.2\%$ & 2.2 GB/GPU \\
\bottomrule
\end{tabular}
\end{table}

The 2B row is the expected trade, near-identical throughput for a large memory saving. The 5B reversible row is the consequential one, where the on-the-fly path improved throughput by 25.2 percent rather than costing anything. The most plausible explanation is allocator pressure. A reversible model already churns activation tensors through repeated recomputation, so removing a multi-gigabyte persistent table relieved memory pressure that was hurting the reversible model and barely affecting the non-reversible one. The relief was worth more than the reconstruction cost. This resident-memory saving is also the reason the smaller models fit a single node at the batch sizes that give good throughput, as noted in Section~\ref{sec:system-backbone}; a standard table at this vocabulary and width would have occupied that memory permanently.

\paragraph{Principle.} A structured embedding is worth reconstructing on the fly when the resident memory it frees exceeds the small compute cost of rebuilding it, and the trade is most favorable on a memory-pressured stack such as a reversible one. Structured embeddings are therefore a resident-memory tool as much as a parameter-count tool, since the full table never has to exist in memory.

\subsection{Reading the 120B single-node claim precisely}
\label{sec:throughput-120b}

The 120B rows in Table~\ref{tab:p3-production} require careful reading, because the claim is narrower than the headline could suggest. The work did not perform dense all-parameter bf16 training of a 120B dense transformer on one node. The model is a sparse mixture of experts with a small active fraction per token, and the released 120B path trained through quantized adapters, with a consolidation step afterward that merged the trained adapter updates into bf16 base weights for release.

The result is still material. The training system handled a sparse model of the 118B-parameter class at 8K context on a single eight-GPU node, with hundreds of gigabytes of model state, expert routing, checkpointing, and the continued-data phase, at roughly 17k to 18k ordinary tokens per second. The operational claim the section supports is stated without inflation. It is not dense 120B training on a node. It is an end-to-end 120B-class sparse MoE training pipeline made practical on one.

\subsection{Summary}
\label{sec:throughput-summary}

The single-node story has two halves. Reversibility and sparse activation made the larger models fit, covered in Section~\ref{sec:reversibility}. Kernel work, loss-stable configuration search, on-the-fly embeddings, and production-grade checkpoint handling made them fast enough to train, covered here. The 2B run shows the magnitude of the speedup, from early high-teens throughput to roughly 59k ordinary tokens per second. The 120B run shows the optimized single-node stack was more than a small-model convenience. It made an end-to-end 120B-class sparse MoE training pipeline practical on one eight-GPU node, which is the claim the rest of the paper is built on.

\section{Failure Modes}
\label{sec:failure-modes}

The previous sections describe the path that succeeded. This section describes the failures that shaped it, at the same level of detail, because a growth recipe that names only its final choices is half a recipe. At this scale the dangerous failures are usually not the ones that crash. The dangerous ones produce a plausible checkpoint, a plausible loss, or a plausible throughput number while quietly violating an invariant the run depended on, and they survive precisely because nothing about them looks wrong.

The failures are organized by the invariant each one broke. Section~\ref{sec:growth} stated the growth principles as interfaces to preserve, and this section is the same set of interfaces seen as the damage done when one is not preserved. Each failure is given as what was tried, the observed breakage, the root cause, the correction, and an explicit statement of how strong the evidence is. Where a failure mirrors a growth principle, the invariant is named in the same liftable form the principles use.

A note on evidence applies throughout. Some failures are backed by saved logs and surviving checkpoints. Others are reconstructed from training-session records because the relevant artifact was never committed to durable storage. Each failure is marked for which it is, because a transcript-grade claim is weaker than a log-grade one, and stating the difference is preferable to implying a uniformity the records do not have.

\subsection{Spectral upcycling: diversity without function preservation}
\label{sec:fail-spectral}

\paragraph{Invariant violated.} Function preservation before expert diversity.

\paragraph{What was tried.} The first dense-to-MoE conversion work included a spectral, SVD-based upcycling strategy alongside the partition strategy that eventually won. The motivation was reasonable on its face. A mixture of experts benefits from diverse experts, so an initializer that explicitly separates experts by spectral transformation and rotation is superficially attractive. The error was optimizing diversity before function.

\paragraph{Breakage and root cause.} The spectral initializer opened near random-initialization loss, with a step-one next-token-prediction loss near 12, against roughly 5.0 to 5.8 for partition and roughly 2.9 for the dense baseline the experts were grown from. The rotations made the experts different but destroyed the dense function they were built from, so the model gave up almost the entire MoE advantage at initialization. Difference between experts is not the same property as usefulness, and the spectral initializer produced the first while destroying the second. Partition preserved usefulness first, by copying the dense feed-forward function into an always-active shared expert and seeding the routed experts from overlapping subsets of the dense neurons, and let the router learn specialization from a state that already worked.

\begin{table}[h]
\centering
\caption{Dense-to-MoE initialization strategies and their cold-start outcomes.}
\label{tab:spectral-vs-partition}
\footnotesize
\begin{tabularx}{\textwidth}{l X X X}
\toprule
Strategy & Mechanism & Intended advantage & Outcome \\
\midrule
Random partition & Overlapping random neuron subsets copied into routed experts & Preserve dense subfunctions, give the router related choices & Selected; surviving production path \\
Drop-upcycling & Copy source weights, reinitialize a fraction & Preserve function while breaking symmetry & Implemented and evaluated \\
Spectral / SVD & Transform and rotate the dense structure & Stronger expert diversity at init & Rejected; near-random cold loss \\
\bottomrule
\end{tabularx}
\end{table}

\paragraph{Correction and invariant.} Function preservation precedes the optimization of expert diversity. Specialization is acquired later by the router, from a working model, and diversity bought at the cost of the function leaves nothing to specialize.

\paragraph{Evidence.} Transcript-grade. The three strategies survive as code and the partition checkpoints survive in durable storage, but the saved three-way comparison output was never committed and the spectral and drop checkpoints were not retained. The loss figures are reported as recorded in session rather than as a reproducible benchmark.

\subsection{Silent checkpoint misload: the source keyspace is not the target model}
\label{sec:fail-misload}

\paragraph{Invariant violated.} Strict, complete parameter transfer.

This is the most dangerous failure in the project, because for a period it did not present as a failure.

\paragraph{What was tried.} An early version of the 9B-to-120B conversion wrote the grown tensors under source-model key names. The target 120B model used a different parameter namespace, because its stack was wrapped differently. Under non-strict loading, the mismatched keys were silently skipped and those target parameters kept their random initialization, while the load reported success.

\paragraph{Breakage and root cause.} The broken-load runs opened at a step-one loss around 2.95, against around 2.55 for the corrected load. A third of a nat at step one was the only signal, and a gap that size reads as initialization noise unless it is being watched for. The reason this is the worst failure in the project is that every cheap check passes.

\begin{table}[h]
\centering
\caption{Cheap checks that all passed against a silently-misloaded checkpoint.}
\label{tab:misload-checks}
\begin{tabularx}{\textwidth}{X X}
\toprule
Check that passed & Why it was insufficient \\
\midrule
Checkpoint file exists & Existence does not prove the target tensors were populated \\
Checkpoint is the right size & Size can come from tensors stored under unused or mismatched keys \\
The model load returns without error & Non-strict loading skips missing target keys instead of failing \\
Parameter count looks right & Randomly initialized parameters still count \\
Training starts and loss is finite & A partly random initialization still produces a plausible loss \\
\bottomrule
\end{tabularx}
\end{table}

\paragraph{Correction.} The direction of the transfer is inverted. Rather than walking the source and writing source-style names, the target model is instantiated, its own parameters and buffers are walked, and for each target name the source tensor that belongs there is found or constructed, with the growth transform applied at that point. The target model's own state dictionary is saved and a strict load is required to succeed, so a missing or misnamed parameter becomes a construction-time error instead of a silent random fallback. This is the keyspace-first principle of Section~\ref{sec:growth-keyspace}.

\paragraph{The released lineage is the corrected one.} The broken load was a predecessor conversion failure and not the release. The released 120B was built with the corrected target-keyspace initializer and trained through the full path before consolidation: the 9B harvest checkpoint at step 66{,}048 became the consolidated 9B source, which the corrected builder expanded into the 120B initialization, which trained through the TQP and continued-data phases to the released periodic checkpoint, which was then consolidated with the adapter merged into bf16 base weights and sharded for release. The lineage is backed by the builder code and the training logs.

\paragraph{Invariant.} Every growth converter is target-keyspace driven and verified by a strict load. Key compatibility is part of the learned-function transfer rather than incidental metadata, and a permissive loader conceals exactly the failure most in need of being visible.

\paragraph{Evidence.} Log-grade and code-grade. The broken-versus-fixed step-one losses are from saved run logs, and the corrected builder and the released lineage are backed by code and training logs.

\subsection{Clone-family collapse: expanded experts can die before they differentiate}
\label{sec:fail-clone-collapse}

\paragraph{Invariant violated.} Utilization of the expanded expert capacity.

\paragraph{What was tried.} The 120B expansion raised routed capacity from 20 source experts to 460 slots by cloning each source expert into perturbed variants and letting training specialize them. The cloning preserves function, which is correct, but it creates a routing problem that the initialization alone does not solve.

\paragraph{Breakage and root cause.} Sibling clones begin highly related, so a hard top-$k$ router gets very little early signal to tell them apart. It selects an early winner per source family and starves the siblings before they receive enough gradient to diverge. The dead-expert count under top-$k$ routing climbed from 27 to 38 to 122 to 154 to 168 out of 460 over the first few hundred steps, while the top-10 routing share sat around ten percent. The model held nominal capacity for 460 experts and used a fraction of it. The mechanism is not the classic MoE collapse in which a few global experts come to dominate; it is collapse along clone families, driven by the router being unable to distinguish near-duplicates early.

\paragraph{Correction and invariant.} Hard top-$k$ is replaced with probabilistic selection during the early differentiation window, so near-duplicate experts get a stochastic chance at gradient before the router hardens. On the same initialization this held maximum expert load to a fraction of a percent, kept zero experts dead, and brought the top-10 share down to a few percent. The invariant has two coupled halves: expert function is preserved at initialization, and routing diversity is then protected through the differentiation window. The initialization preserves the function and the routing scheme preserves the capacity, and the expansion fails to deliver its nominal capacity if either is omitted.

\paragraph{Evidence.} Mixed, and separated. The sharp 27-to-168 trajectory is transcript-grade and comes from the clone-based initialization, which was not the released path. The released path used drop-upcycling but kept top-$k$ routing in its recovered segments and showed the same early concentration, with per-expert loads spanning a wide range and the top-10 share rising from around five percent, from saved balance logs. The transcript-grade clone trajectory and the log-grade released-path curve are kept distinct rather than presenting one as the other.

\subsection{Role-blind expansion: auxiliary heads inflate if expanded like experts}
\label{sec:fail-role-blind}

\paragraph{Invariant violated.} Role-aware rather than shape-aware expansion.

\paragraph{What was tried.} The expert-expansion logic, applied naively during checkpoint conversion, expanded every tensor with an expert-like shape by the same factor. The multi-token-prediction head carries its own expert-shaped tensors.

\paragraph{Breakage and root cause.} Those auxiliary tensors were expanded as if they were main routed experts, inflating parameters in the auxiliary training path that did not need the capacity. The converter matched on shape where it should have matched on role. An expert-shaped tensor is not necessarily an expert.

\paragraph{Correction and invariant.} Auxiliary heads are audited separately during conversion and copied appropriately rather than expanded as main-stack experts. The invariant is narrow: expansion is determined by the role of a tensor rather than by its shape. A conversion that asks only whether a tensor has an expert-like shape will expand tensors that merely resemble experts.

\paragraph{Evidence.} Code-grade, from the conversion logic and its fix.

\subsection{Mixture-shift instability}
\label{sec:fail-mixture-shift}

\paragraph{Invariant violated.} Bounded gradients across curriculum transitions under frozen parameters.

\paragraph{What was tried.} The curriculum advances the data mixture from easier to harder across the run. Some of those transitions are sharp, and they occur while parts of the model, including the embeddings, are frozen.

\paragraph{Breakage and root cause.} Sharp mixture shifts under frozen parameters concentrated the adjustment onto the unfrozen parameters, which raised the risk of large gradients at transition points. When the distribution moves abruptly and part of the model cannot move with it, the parts that can move absorb the entire shift, and a transition that appears harmless in the schedule can produce a gradient spike in practice.

\paragraph{Correction and invariant.} Transitions are blended rather than switched in one step, the same warmup-at-the-seam approach the growth transitions use, applied to curriculum transitions for the same reason. The invariant is that distribution changes are ramped rather than stepped, especially when parameters are frozen.

\paragraph{Evidence.} This is the thinnest-evidenced failure in the section, and it is marked as such. The qualitative behavior and the adopted mitigation are recorded, but no clean saved trace isolates the instability, so it is described at the level the records support, with no specific numbers attached that cannot be stood behind. It is included because the lesson is real and consistent with the curriculum story, not because the spike can be shown.

\subsection{Curriculum accounting: a guarantee is only as good as its measurement}
\label{sec:fail-curriculum}

\paragraph{Invariant violated.} The per-batch composition the guarantee claims to enforce.

This failure directly undercuts a claim made in Section~\ref{sec:data}, and it is recorded here rather than omitted for that reason.

\paragraph{What was tried.} Section~\ref{sec:data} describes the Always-ON tier as injecting a fixed eight percent of every batch. The injection share was set by a hardcoded per-step count.

\paragraph{Breakage and root cause.} That hardcoded count interacted with the batch configuration so that the effective Always-ON share ran closer to 57 percent than to the intended level, for roughly 23{,}000 steps of the 9B training, before the imbalance was caught and a rampdown fix landed. Across the 9B run this meant something like 10B of the roughly 17.5B tokens were Always-ON, split between benchmark and Indic data, far above the design intent. The guarantee was being enforced by a constant, and the constant did not evaluate to the intended number under the real batch configuration.

\paragraph{Correction and invariant.} Per-batch composition is measured from the batches the loader actually emits, and logged, rather than trusted to a constant that no one is watching. The invariant is that a guaranteed mixture is verified from emitted batches rather than assumed from configuration.

\paragraph{Evidence.} Transcript-grade for the exact share and step count, since the per-batch source tags were console-only by design and not all console logs were preserved. The token totals are reconstructed from that record. The model was not visibly harmed by the over-exposure, and it may even have helped the guaranteed-capability results, but that is luck rather than design, and the honest reading is that the imbalance was caught late.

\subsection{Flush divergence at scale: a small-scale technique that did not transfer}
\label{sec:fail-flush}

\paragraph{Invariant violated.} Verification of a training technique at the scale it is deployed, rather than inheritance from smaller runs.

\paragraph{What was tried.} The periodic flush of TurboQuant Training, in which the low-rank adapter delta is folded into the base and reset on a fixed cadence so that accumulated low-rank increments reach a high-rank target, was the assumed training mechanism going into the 120B. It had worked in the sub-1B experiments that preceded the run, consistent with the prior reports that introduced it, and it was carried into the 120B bring-up on that basis.

\paragraph{Breakage and root cause.} At 120B the flush-based configuration diverged during bring-up, before the pretraining run could proceed, and the divergence persisted across the mitigations attempted against it. Unlike the other failures in this section, this one did not produce a plausible artifact that had to be caught by a subtle signal; it failed loudly. Its cost was not a hidden defect but lost time and compute, since reaching a stable configuration required abandoning the flush and restarting. The cause is not fully isolated, and Section~\ref{sec:turboquant} states it at the level the records support: it is consistent with the per-window update at 120B being too high-rank for a rank-16 increment to track even with frequent merging, so that repeated absorption injects more disturbance than it resolves, but the firm claims are the fact of the divergence and the stability of the flushless alternative rather than the mechanism.

\paragraph{Correction and invariant.} The flush was removed entirely, and both 120B runs, pretraining and the supervised run, were flushless from the start. The largest single loss reduction observed in the 120B training belongs to the flushless regime. The invariant is that a training technique validated at small scale is re-verified at the target scale before being relied upon, because a method whose justification rests on training dynamics can reverse between scales, and the verification is cheaper than the restart its absence forces.

\paragraph{Evidence.} Log-grade for the flushless configuration and its loss trajectory, which are the released lineage and are backed by training logs. The flush-based divergence during bring-up is recorded from the bring-up logs and the abandoned configuration, with the mechanism left as the candidate account of Section~\ref{sec:turboquant} rather than an established cause. The full configuration is in Section~\ref{sec:turboquant}.

\subsection{The common thread}
\label{sec:fail-common}

Every failure here is a violated invariant that Section~\ref{sec:growth} names as a thing to preserve, or, in the last case, an assumption carried across scales without re-verification. Spectral upcycling broke function preservation. The silent misload broke complete parameter transfer. Clone-family collapse broke capacity utilization. Role-blind expansion broke role-aware transfer. Mixture-shift broke bounded gradients at transitions. The Always-ON imbalance broke the composition guarantee it claimed to provide. The flush divergence broke the assumption that a small-scale technique transfers unchanged to a much larger model. The pattern is the growth section seen in negative. A transformation, a setting, or an inherited technique that appeared sound broke a relationship the run depended on, and in most cases it did so without crashing, leaving a plausible artifact that had to be caught by watching the single number that exposed it. The flush divergence is the exception that failed loudly, and its cost was time and compute rather than a hidden defect. The reusable practice, more than any single fix, is to identify which invariant each step depends on and to instrument the measurement that proves it, rather than treating a clean-looking checkpoint or a finite loss as evidence that the invariant held.

\section{Data Methodology}
\label{sec:data}

The data pipeline for LightningLM 0.1V was built on one conviction. What the model is guaranteed to see matters more than what a selection heuristic happens to prefer. This section describes the three-tier corpus architecture that enforces that guarantee, the tokenizer and corpus failures found and fixed, and the difficulty curriculum whose marks appear in the loss curves of Section~\ref{sec:results}. Several of the stronger downstream results, in particular held-out code loss and Indic generation, follow directly from choices made here, and each is traced to the held-out number it produced.

One scope note first. Where this section gives sizes, pool weights, and the shape of the selection system, those describe design intent and come from the corpus specification. Where it gives per-pool held-out loss, those describe outcomes and are computed from evaluation logs against the released checkpoints. The two are kept apart deliberately. The design program was larger than any single run actually consumed, and a planned budget should not be read as a trained one.

\subsection{The selection problem}
\label{sec:data-selection}

Pretraining at constrained compute leans on data selection. Rather than training uniformly over a corpus, a scorer ranks candidate batches and the optimizer spends its budget on the ones that help most. The pipeline uses OPUS for this \citep{OPUS}, which scores a candidate by projecting its optimizer-shaped update onto a target direction taken from a held-out reference proxy. When the proxy represents the target capability, this places compute where it is most useful.

The reason to accept the cost documented in Section~\ref{sec:throughput-production} is the efficiency OPUS offers when it works. The originating work reports an eight-times reduction in computation on GPT-2 XL using FineWeb, meaning a model trained on OPUS-selected data matches one trained on random selection over roughly eight times as many tokens \citep{OPUS}. A multiplier of that shape on the bulk corpus is a large saving at constrained compute, and capturing it on the pools where it applies is the entire reason the selector was used despite its overhead. That same framing bounds where it was worth using, which is taken up below and in Section~\ref{sec:throughput-production}.

The catch is that the scorer inherits whatever bias sits in its reference direction. The proxy used here had a bias that mattered a great deal for the intended model. The golden proxy is English-dominant, with a cosine similarity of 0.876 to the main English web band. Any scorer aligning to that direction marks down data that sits far from it, and nothing sits farther than Indic-script text, whose token statistics do not project well onto an English reference. Run on its own, the scorer would have quietly starved the data most in need of protection. The same trap waits for anything essential but statistically unusual: benchmark reasoning traces, function-calling formats, and the long tail of code. They are scored down for being specialized, which is the property that makes them worth keeping.

That is why the data system is not a single selected stream. It is three tiers, and they stand in different relationships to the scorer.

\begin{table}[h]
\centering
\caption{The three corpus tiers and their relationship to the OPUS selector.}
\label{tab:tiers}
\begin{tabularx}{\textwidth}{l X l}
\toprule
Tier & Role & Relationship to OPUS \\
\midrule
OPUS-eligible (D1 to D4) & The main body: web foundation, diverse web, code, STEM. The scorer ranks and keeps the best-aligned fraction. & Scored and filtered \\
Always-ON (AON) & Data that must survive regardless of what the scorer prefers. Injected straight into every batch. & Bypasses the scorer \\
Golden Proxy (GP) & The scoring reference direction. & Never trained on \\
\bottomrule
\end{tabularx}
\end{table}

The corpus underneath these tiers is laid out in Table~\ref{tab:t1-corpus}. The design logic is visible in the numbers: the easy, clean pools are large and feed the early stages, the hard pools are protected and ramped in later, and the Indic data, small as a fraction of the whole, is carved out and guaranteed rather than left to compete on the scorer's terms.

\begin{table}[h]
\centering
\caption{Table T1. Training corpus by pool (design specification).}
\label{tab:t1-corpus}
\footnotesize
\setlength{\tabcolsep}{4pt}
\begin{tabularx}{\textwidth}{l X r r r X}
\toprule
Pool & Role & Shards & Tokens & Share & Sources \\
\midrule
D1 Web Foundation & cleanest English; early-stage backbone & 4{,}894 & 164.1B & 14.7\% & CommonCrawl head, Reddit \\
D2 Web Diverse & largest pool; carries nearly all Indic & 18{,}710 & 627.4B & 55.1\% & CC tail/middle, RefinedWeb, AI4Bharat, IndicCorpV2 \\
D3 Code & hardest modality by entropy & 5{,}933 & 199.0B & 17.8\% & StarCoder/The Stack, plus tab and CRLF code bands \\
D4 STEM & capability bottleneck; tightest pool & 1{,}464 & 49.1B & 4.4\% & peS2o, arXiv, proof-pile-2, FLAN \\
AON bench\_train & guaranteed reasoning, code, instruction & 356 & 11.2B & 2.0\% & math/code traces, function-calling, FLAN \\
AON indic\_guaranteed & guaranteed Indic, native numerals & 1{,}996 & 66.9B & 6.0\% & Indic-heavy shards with numeral swap \\
\midrule
\textbf{Training total} & & \textbf{33{,}353} & \textbf{\textasciitilde 1.12T} & \textbf{100\%} & \\
Golden Proxy & scoring reference; never trained on & 11 & 0.007B & n/a & benchmark validation splits \\
\bottomrule
\end{tabularx}
\end{table}

The corpus is English-dominant by token count, roughly 74\% Latin script, with the nine guaranteed Indic scripts together under 4\%. That imbalance is the reason the scorer's English-aligned proxy is dangerous and the reason the Indic data is moved out of the scorer's reach. A signal that is 4\% of the corpus and projects poorly onto the reference direction is the kind of data a utility-maximizing scorer discards, and the kind the model needs guaranteed.

The pools also differ sharply in statistical character, which is what the difficulty curriculum in Section~\ref{sec:data-curriculum} is built around. Table~\ref{tab:t3-pool-stats} gives the headline measures.

\begin{table}[h]
\centering
\caption{Table T3. Pool statistics (design specification).}
\label{tab:t3-pool-stats}
\footnotesize
\setlength{\tabcolsep}{4pt}
\begin{tabular}{lrrrr}
\toprule
Metric & D1 & D2 & D3 & D4 \\
\midrule
Token entropy & 10.96 & 11.71 & 12.08 & 10.3 to 11.0 \\
Repetition & 2.0\% & 3.8\% & 40.3\% & 39 to 42\% \\
Chars per token & 4.63 & 4.40 & 3.48 & \textasciitilde 3.17 \\
Difficulty composite & 0.37 & 0.52 & 0.61 & 0.41 \\
\bottomrule
\end{tabular}
\end{table}

Code is the highest-entropy, highest-repetition pool, the repetition driven by boilerplate, imports, and license headers. STEM appears deceptively low-entropy because formal mathematics is formulaic at the token level, yet it is the hardest capability for the model to acquire, which is why it carries the sharpest ramp despite its modest difficulty composite. The numbering of the pools follows capability difficulty rather than raw text entropy, and that distinction is the reasoning behind the curriculum order.

The scorer maximizes efficiency over the bulk corpus. The Always-ON tier enforces coverage where the scorer cannot be trusted. The proxy stays held out so scoring never leaks into training. That separation, more than any single pool, is the part of the data work most worth reproducing.

\paragraph{Principle.} A single selector should not govern data whose value it cannot see. The categories a scorer is structurally biased against, meaning anything statistically far from its reference direction, are identified and moved into a tier that bypasses the scorer entirely, rather than left to a reweighting that is unlikely to rescue them.

\subsubsection*{Where selection was used, and where it was not}

The efficiency case for OPUS holds only at the stage where its reference direction is informative and the data it must rank is not uniformly essential. Across the production stages, the selector was active at one stage and disabled at the two surrounding it, for reasons that follow directly from the bias analysis above.

At the 2B stage the model was non-reversible and ran at the highest throughput of any stage, and the data drawn there was the easy end of the difficulty curriculum, where candidates differed little in utility. The selector had little to discriminate between, and the scoring cost of Section~\ref{sec:throughput-production} would have bought selection pressure the easy-token regime did not reward, so the stage was run without it, at full speed, to consume the maximum volume of easy tokens while throughput was high.

At the 5B stage the candidate pool was mixed in difficulty and the selector's discrimination was both reliable and useful, so OPUS was active and the eight-times efficiency argument applied to a pool where it could be realized. This is the one stage that paid the scoring cost recorded in Section~\ref{sec:throughput-production}.

At the 9B stage the mixed-difficulty pool on which selection had paid off was exhausted, and the remaining data was uniformly hard, including large volumes of LaTeX and assembly. This is the case the bias analysis predicts most sharply. The OPUS proxy is constructed by retrieving samples aligned to benchmark validation sets, so it systematically undervalues content far from that distribution, and LaTeX and assembly are exactly such content. Running the selector at 9B would have scored down the hard data the stage existed to consume, which is the same failure the Always-ON tier exists to prevent, arriving through the bulk-selection path rather than the guaranteed path. The selector was therefore disabled at 9B, and the hard data was trained on directly. The blind spot that motivates the three-tier design is, at this stage, the concrete reason selection was switched off rather than an abstract risk.

\subsection{Guaranteed per-batch composition}
\label{sec:data-aon}

Always-ON is not an adjustment to selection probabilities. It is a hard injection. A fixed slice of every batch, at every stage, comes straight from AON and never touches the scorer. In production that slice is eight percent, split as two percent benchmark reasoning, code-engineering, and instruction data, and six percent Indic text across the major scripts of the subcontinent.

The result is a per-step invariant. No batch, at any scale, lacked Indic and lacked code and reasoning. Whatever the scorer did on top of that floor, the floor held. The phrasing is deliberate, because it converts a soft commitment into a hard one. The claim is not that training used a multilingual, code-rich corpus, a statement available to any team, but the stronger and checkable claim that every gradient update was taken over a batch containing guaranteed Indic and guaranteed code.

\paragraph{Principle.} Capability commitments are expressed as per-batch invariants rather than corpus-level averages. A guarantee that holds on average over the corpus can be violated on any given batch by the selector, whereas a guarantee enforced as a fixed fraction of every batch cannot. The commitment is stated as a per-step floor and enforced underneath selection.

Eight percent was a chosen value rather than a default. A smaller floor makes the guarantee decorative, and a larger one eats into the scorer's efficiency on the bulk corpus while over-exposing a narrow slice, so the value was set to be large enough to guarantee coverage and small enough to leave the selector its budget. Section~\ref{sec:fail-curriculum} records a production incident in which a loader bug pushed the effective AON share far above its target for a long stretch of the 9B training, which is its own lesson: a guarantee is worth only what its accounting is worth, and per-batch composition has to be read from the batches the loader actually emits rather than from the configuration intended to produce them. Once corrected, AON held at its intended eight percent for the remainder of training.

\subsection{The dead-token audit}
\label{sec:data-deadtokens}

Guaranteeing categories of data does not guarantee that every token in the vocabulary is exercised. A check of token frequencies across the corpus surfaced a failure that is easy to walk past and obvious once named. A large group of tokens in the 131{,}072-token vocabulary had zero occurrences in the entire corpus. These tokens were not merely rare; they had never appeared.

The dead tokens fell into two telling groups. The first was native Indic numerals. The tokenizer carries dedicated tokens for the script-native digits of each Indic system, the Devanagari \devtext{०} through \devtext{९}, the Bengali \bentext{০} through \bentext{৯}, and their counterparts across roughly ten scripts. Every one had zero count, because the source text used Arabic digits 0 through 9 even inside otherwise-Indic passages. The model held capacity reserved for native numerals and had no data to train it on. The second group was code structure: tab-prefixed indentation tokens and Windows CRLF line-ending tokens, again at zero, because the bulk code had been normalized in ways that erased exactly those byte patterns.

Both were fixed in the corpus rather than in the model. For the numerals, the shards carrying Indic content were located and Arabic digits adjacent to Indic script were swapped for the matching native digits, roughly 4.7 million substitutions across about 2{,}000 shards, so every native numeral token acquired real frequency. For the code tokens, a corpus heavy in the missing constructs was assembled from kernel, firmware, CUDA, and embedded-systems repositories, where tab indentation and Windows line endings are normal, and tokenized into dedicated bands. The previously dead tab and CRLF tokens reached full coverage.

This warrants a paragraph because it rarely appears in a paper and routinely costs other teams weeks. A vocabulary and a corpus are one design rather than two. Reserving a token is not the same as training it, and the only way to distinguish the two is to inspect the occurrence counts directly. The general rule is that after any tokenizer change or corpus normalization, the occurrence histogram over the whole vocabulary is checked for zeros before training rather than after.

\paragraph{Principle.} The full-vocabulary occurrence histogram is audited for zero-count tokens before training, and dead tokens are fixed at the corpus level rather than the model level. A token the corpus never produces is capacity the model cannot train, and the fix is to make the corpus produce it, through targeted substitution or targeted data, rather than to rely on the model to cope.

\subsection{The difficulty curriculum, and the role of held-out evaluation}
\label{sec:data-curriculum}

The corpus is not served at a fixed mixture. It shifts across stages from easier to harder, with the bulk weights moving off general web text and onto code and STEM as the model grows into the capacity to handle them. Web falls from roughly seventy percent of the bulk mix at the smallest scale to under twenty at the largest. Code climbs from about an eighth to a third. STEM climbs from under a tenth to nearly two-fifths. The eight-percent Always-ON floor stays fixed underneath all of it.

Table~\ref{tab:t2-curriculum} makes the ramp concrete. The bulk weights rotate from web toward code and STEM across the planned stage program, while the Always-ON floor holds flat at eight percent throughout.

\begin{table}[h]
\centering
\caption{Table T2. Pool mixture by stage (design specification; weights are fractions of each batch).}
\label{tab:t2-curriculum}
\footnotesize
\setlength{\tabcolsep}{6pt}
\begin{tabular}{lrrrr}
\toprule
Pool & 2B & 5B & 9B & 120B \\
\midrule
D1 Web Foundation & 0.42 & 0.22 & 0.09 & 0.06 \\
D2 Web Diverse    & 0.30 & 0.28 & 0.18 & 0.12 \\
D3 Code           & 0.13 & 0.25 & 0.33 & 0.35 \\
D4 STEM           & 0.07 & 0.17 & 0.32 & 0.39 \\
AON (always-on)   & 0.08 & 0.08 & 0.08 & 0.08 \\
\bottomrule
\end{tabular}
\end{table}

The table describes the design program across the full stage sequence through 120B. The schedule is shown because it is the one the stages actually followed, and because the direction of the ramp rather than its endpoint is the point: web falls, code and STEM rise, and the guaranteed floor never moves.

This curriculum explains a feature of the Section~\ref{sec:results} loss curves that would otherwise read as trouble. Each stage's training loss has upward excursions, places where the loss climbs before resuming its fall. The quantity that trended down throughout was the held-out evaluation, not the training loss, and the held-out evaluation is therefore the instrument that distinguished a working difficulty increase from a failing one. When the mixture moves toward code and STEM, or a difficulty band inside the benchmark pool is advanced, the model is suddenly predicting harder text and the training loss rises until it adapts. The fall that follows is the model absorbing the harder distribution. Without the held-out signal, a rising training loss would invite a retreat to whatever data showed a lower number, with no view of what that data's held-out performance actually was. The held-out set is what showed that a difficulty increase was working rather than backfiring.

The transitions are not each marked with an exact step. The difficulty schedule was adjusted during training in ways that cannot be fully reconstructed afterward, so the mechanism is described rather than precise boundaries claimed. What the records support is that the curriculum moved from easier to harder material, that the loss curves carry the marks of those moves, and that the measure of progress was continued loss reduction on newly introduced harder data rather than a single smooth training-loss line.

\subsection{From mechanism to outcome}
\label{sec:data-outcome}

Everything above is mechanism. Whether it worked is a question the held-out evaluation answers directly. Each released checkpoint is evaluated against held-out shards from each pool separately, so performance reads per domain rather than as one blended number. Measured at each stage's released checkpoint, which is the consolidated harvest used to grow the next stage rather than a hand-picked low point, the per-domain picture is consistent. The figures are the released-checkpoint values, which differ from the exact harvest-step readings only at the third decimal and are the ones reproducible against the public models.

\begin{figure}[t]
\centering
\includegraphics[width=0.92\textwidth]{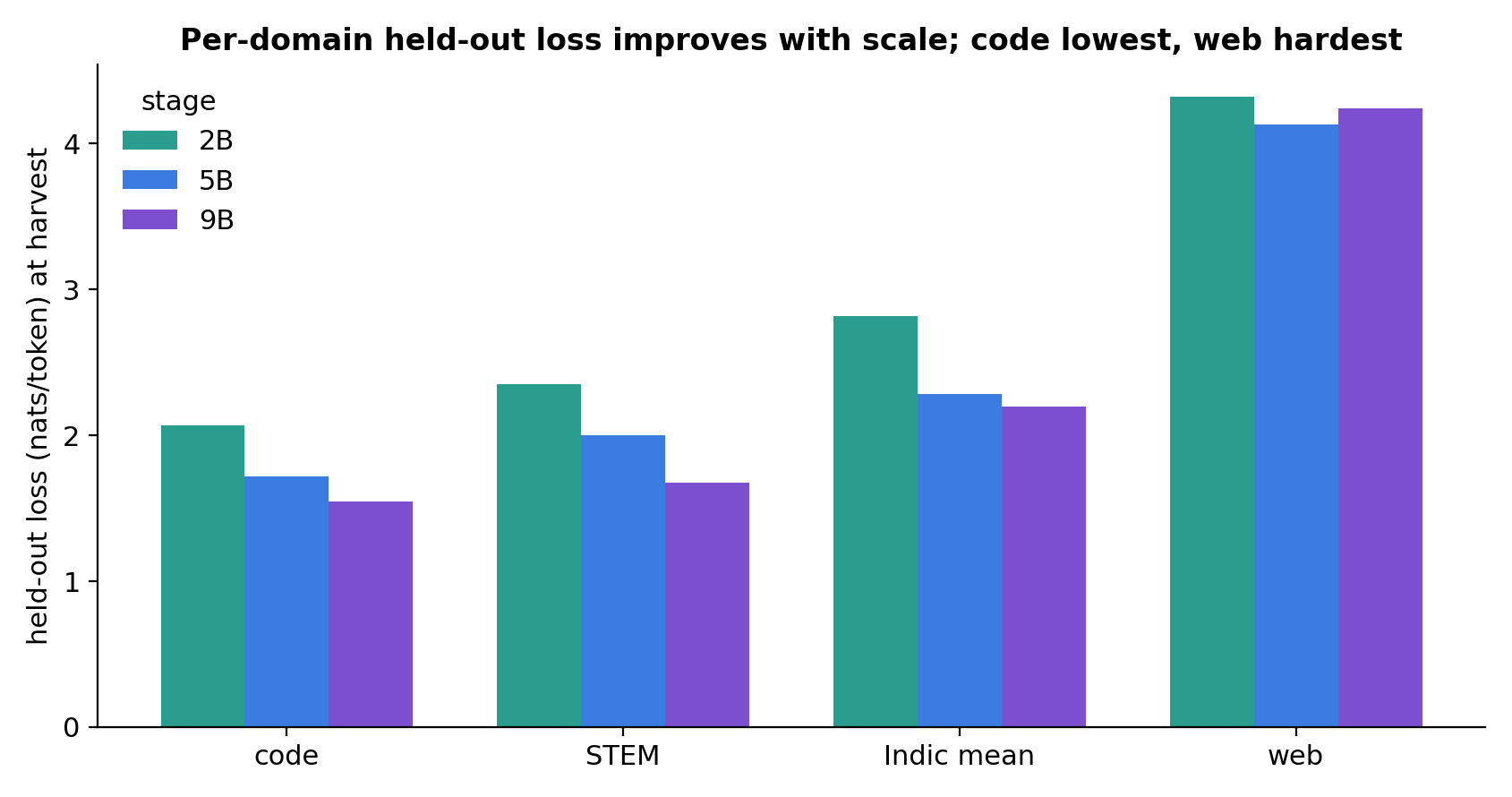}
\caption{Per-domain held-out loss at each stage's harvest checkpoint, for code, STEM, the Indic-script mean, and web text. Loss improves with scale in every domain; code is lowest, web hardest. Discussed as mechanism here and revisited as a result in Section~\ref{sec:results}.}
\label{fig:pool_domains}
\end{figure}

Code is the strongest domain at every scale. Its held-out loss falls from 2.07 at 2B to 1.72 at 5B to 1.54 at 9B (Figure~\ref{fig:pool_domains}). STEM sits just above, 2.35 to 2.00 to 1.68. Both improve with every growth step and both stay well under general web text, which holds above 4.1 throughout. The high web number is not a problem. Web is the hardest and most varied bucket, the one the curriculum deliberately backs away from over time, and a model that found open web as easy as curated code would only indicate that its web evaluation was too soft.

\begin{figure}[t]
\centering
\includegraphics[width=0.92\textwidth]{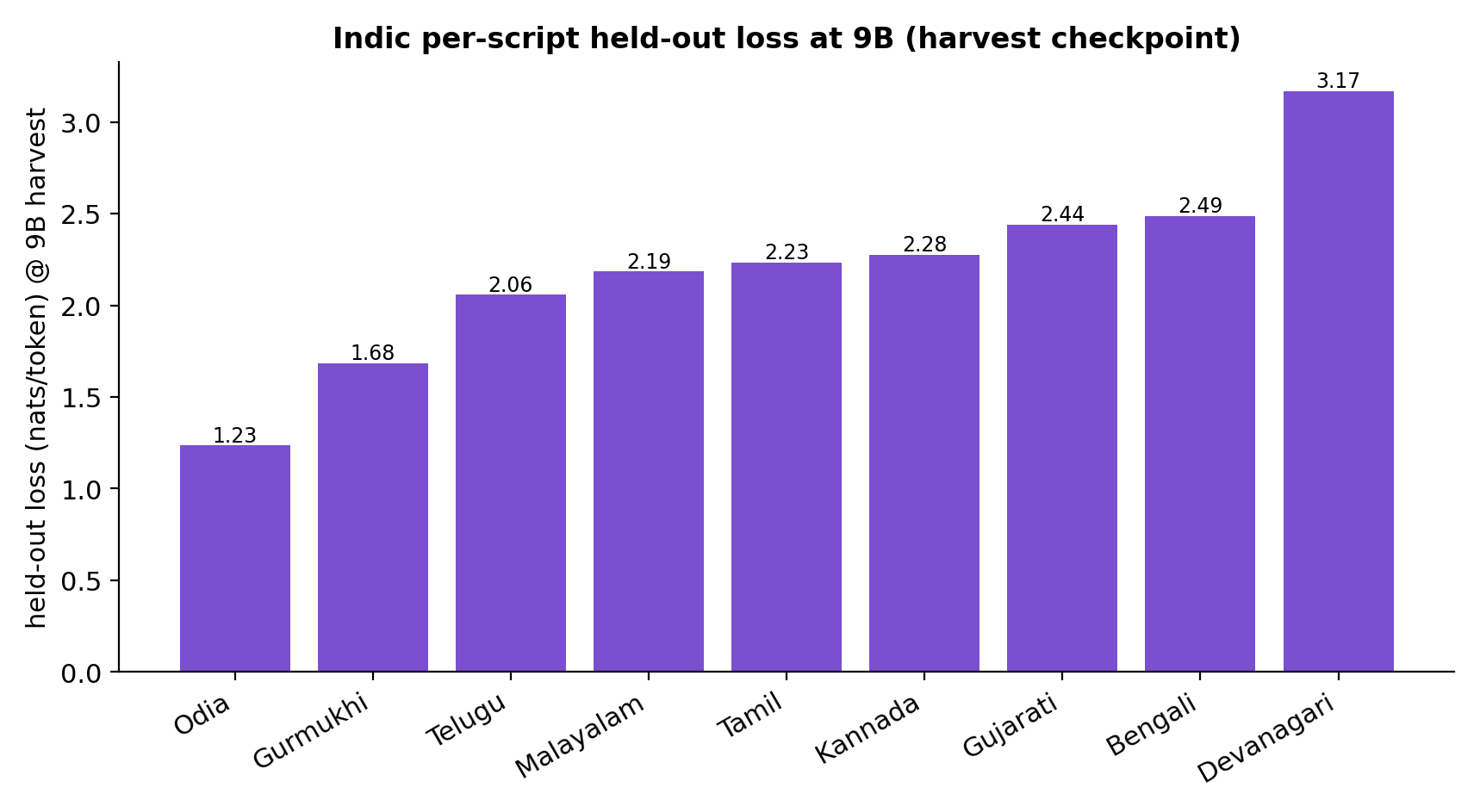}
\caption{Indic per-script held-out loss at the 9B harvest checkpoint. Each protected script reaches a usable loss; the higher Devanagari and Hindi values are most plausibly an evaluation-set difficulty artifact (Section~\ref{sec:limitations}). Revisited in Section~\ref{sec:results}.}
\label{fig:indic_scripts}
\end{figure}

The code result is the return on guaranteeing code in every batch: the domain protected by construction is the domain the model is best at on held-out data. Indic is the same story along the axis that drove the three-tier design. Indic held-out loss improves with scale across every script, and at the 9B harvest it runs from 1.24 for Odia and 1.68 for Gurmukhi up through the low twos for most major scripts (Figure~\ref{fig:indic_scripts}). Those are the numbers of a model that saw Indic on every step rather than treating it as a later addition. One detail is left visible rather than smoothed over: the highest-resource scripts, Devanagari and Hindi, show the highest Indic loss at 9B rather than the lowest. The likely cause is the evaluation rather than the model, since the held-out sets for the highest-resource scripts are larger, more varied, and therefore harder than those for the smaller scripts. It is flagged and left visible.

The cross-stage comparability of some evaluation sets is treated as a real limit rather than a result, and Section~\ref{sec:limitations} says so plainly. The pools relied on as evidence here, code, STEM, and the individual Indic scripts, are the ones whose held-out sets are confidently toward harder rather than merely different material. Figures~\ref{fig:pool_domains} and~\ref{fig:indic_scripts} show these numbers as they stand, and Section~\ref{sec:results} returns to them as results. The reading taken from them is modest in shape and specific in content: the work did not target a leaderboard but protected particular capabilities by construction, and the held-out loss indicates those capabilities were learned.

\section{TurboQuant Training}
\label{sec:turboquant}

The 120B model could not be trained the way the smaller stages were. Full optimizer state over its parameters does not fit on the node, and no amount of careful scheduling changes that arithmetic. TQP, for TQ-PreTraining, is the strategy used to get around it, where TQ is TurboQuant, the vector quantizer of Zandieh and colleagues \citep{TurboQuant}. This section explains why TQP targets what it targets, why that target choice is driven by parameter mass rather than by any low-rank property of the weights, why the quantizer that holds the base weights was repurposed from a setting it was not designed for, and why a periodic-merge technique that worked at small scale had to be abandoned at 120B at considerable cost. Two of these findings run against the explanation the work began with, and the corrected version is the more useful one to carry forward, because it states when the strategy applies and when it does not.

\subsection{The memory wall at 120B}
\label{sec:tq-memory-wall}

A 120B mixture-of-experts model spends almost all of its parameters in the routed experts. Twenty layers, 460 routed experts per layer, three weight packs per expert at $4096 \times 1024$, comes to something over 100B parameters in the experts alone. Attention, by contrast, is under 2B parameters across the whole model, and the shared expert is around half a billion. The memory goes almost entirely into routed-expert weights, and more consequentially into the optimizer state that shadows them.

That optimizer state is the binding constraint. Adam keeps two moment estimates per trainable parameter. In fp32 over 100B parameters that is on the order of 800\,GB, before the master weights or activations are counted. On a single eight-GPU node this does not fit, and does not approach fitting. The training problem at 120B reduces to one question: how the experts can continue to train without paying full Adam state on 100B parameters.

TQP answers it by holding the base expert weights in a compressed representation and training low-rank adapters on top, so the only expert parameters carrying optimizer state are the adapters. The adapter is rank 16 on each of the three weight packs per expert, across all 460 experts in each of the 20 layers, which comes to about 2.26B trainable parameters in the expert path, down from over 100B. The optimizer state for that path drops from roughly 800\,GB to about 18\,GB, a factor of about 45. That reduction is the reason TQP exists in this project. It is a memory strategy before it is anything else.

\subsection{The quantized base: repurposing TurboQuant as a trainable representation}
\label{sec:tq-quantbase}

The compressed base in TQP is a TurboQuant representation of the expert weights. TurboQuant was introduced as a data-oblivious online vector quantizer with near-optimal distortion, and its demonstrated targets were inference-time vectors, in particular KV-cache compression and nearest-neighbor search \citep{TurboQuant}. Its use here is different. The TurboQuant quantizer holds the base weights of a 460-expert mixture-of-experts stack, and the model is trained through that representation with rank-16 adapters. Using an inference-side vector quantizer as the trainable base of an expert stack is, as far as can be determined, not previously reported. The quantizer is not a contribution of this work; the contribution is the training construction around it and the demonstration that it is trainable at this scale on one node.

The mechanism TurboQuant provides, and that TQP uses, is the MSE path of the method. For a weight vector, the quantizer separates scale from direction, normalizes the vector and stores its norm, applies a seeded random orthogonal rotation, and then quantizes each rotated coordinate independently against a scalar codebook. The rotation is the load-bearing step. After a random orthogonal rotation the coordinates of a unit vector are distributed like the coordinates of a random point on the sphere, which is a known law, a shifted Beta distribution on the interval, so the Lloyd-Max scalar codebook for a given dimension and bit width can be computed analytically from that law rather than fitted to calibration data. The rotation is a signed Hadamard transform when the dimension is a power of two and a QR rotation from a Gaussian matrix with sign fixing otherwise. Reconstruction looks up the codebook values and rescales by the stored norm. The inner-product-oriented TurboQuant variant, which adds a one-bit residual correction to debias inner-product estimates, was not used; TQP uses the MSE path only.

The property that makes this a viable training base rather than only a compression scheme is that the codebook is analytic and therefore calibration-free. The base weights move throughout training as adapter deltas are absorbed into them, and a quantizer that required a fitted codebook would need recalibration each time the base changed. A codebook derived from the fixed coordinate law of the rotation does not, because the law does not depend on the particular weights. The base is held at 8-bit weight precision, with the per-expert weight packs quantized row by row. The combination, an analytically-quantized moving base under a low-rank trained adapter, is what the next sections examine, first for why the experts are the right place to spend the adapter budget and then for how the adapter is made to do high-rank work, including the technique that failed.

\paragraph{Principle.} A data-oblivious quantizer whose codebook is analytic rather than fitted can serve as a trainable base, because a moving base does not invalidate a calibration-free codebook. A quantizer that depends on calibration to the specific weights is unsuitable for a base that changes during training.

\subsection{Where the optimizer state lives}
\label{sec:tq-opt-state}

When this section was first written, TQP was justified the way these methods usually are. A well-trained network has low-rank structure, low-rank adapters match that structure, and adapting the experts through low-rank updates should therefore lose little. It is the standard story, and for this model it is not correct, which a measurement showed.

The released 9B model, the exact checkpoint drop-upcycled into the 120B, was taken and the singular value spectrum of every weight matrix in it computed. For each matrix the question asked was how many singular directions are needed to capture 90\% of the spectral energy, as a fraction of full rank. A genuinely low-rank matrix needs a small fraction. The embedding projection sits at 0.21, with a single direction carrying a quarter of its energy, which is what low-rank looks like.

\begin{figure}[t]
\centering
\includegraphics[width=0.92\textwidth]{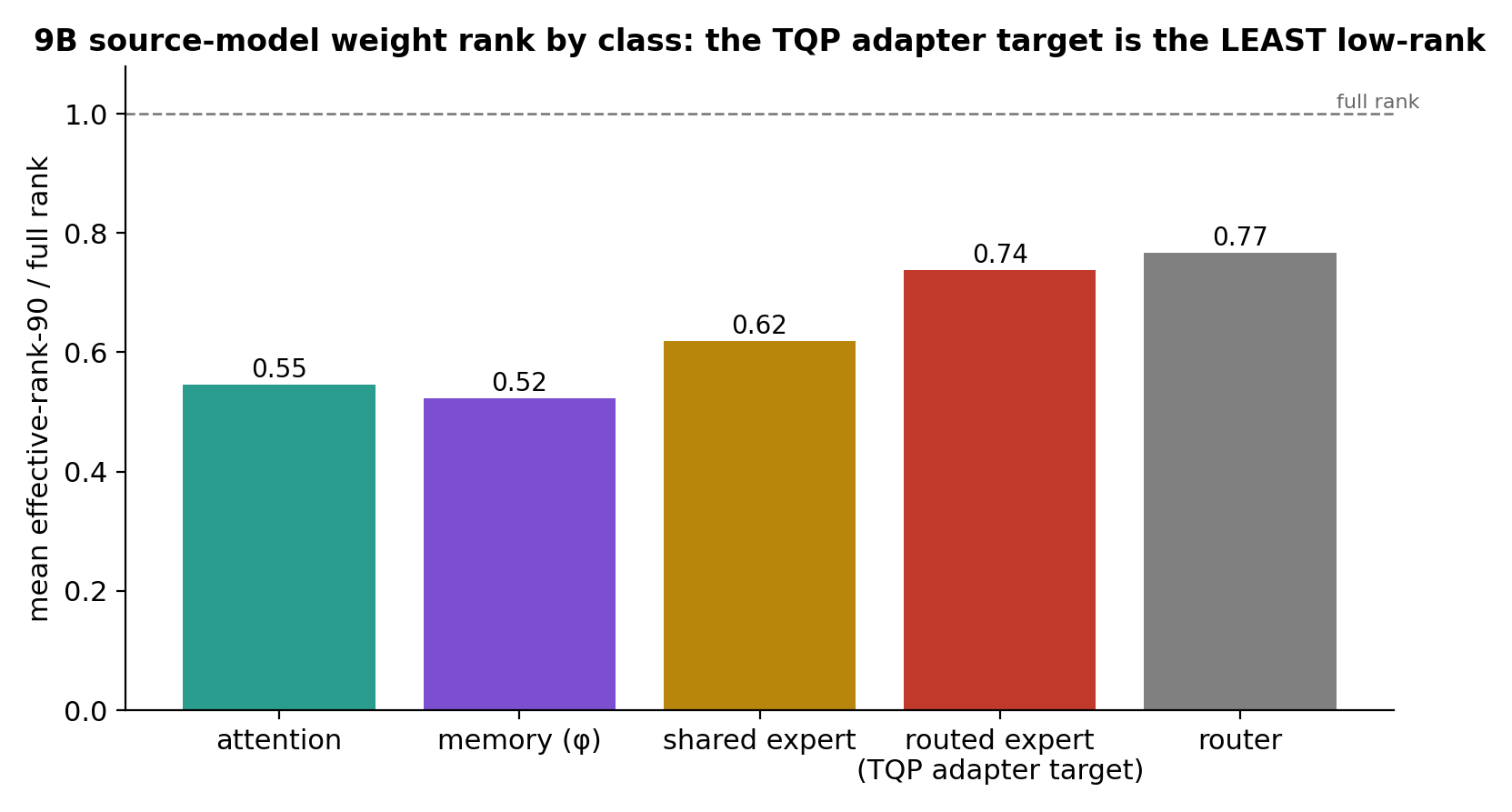}
\caption{Effective rank by weight class in the released 9B, as the fraction of the spectrum needed to capture 90\% of the energy. The routed experts, the TQP adapter target, are the least low-rank class, which inverts the usual low-rank justification for adapting them.}
\label{fig:tql_rank}
\end{figure}

The routed experts do not look like that. They sit at 0.74, so capturing 90\% of their energy needs nearly three-quarters of the full spectrum. Attention is markedly more compressible at 0.55, the memory-feedback matrices at 0.52, and the shared expert at 0.62. The routed experts, the one class adapters were placed on, are the least low-rank class in the model (Figure~\ref{fig:tql_rank}). On a criterion of where the weights are actually low-rank, the priority order is inverted, and the choice would be to adapt attention and leave the experts alone.

The reason TQP targets the experts therefore has nothing to do with their rank. It targets them because that is where the parameters are. Adapting attention would save almost nothing, since attention is under 2B parameters to begin with, and adapting the shared expert saves little for the same reason. The experts hold the 100B parameters and the 800\,GB of Adam state, so the experts are where adapters have to go when memory is the constraint. The target was chosen by parameter mass, the rank of those parameters was not part of the decision, and the rank measurement, once taken, argued against the standard justification rather than for it.

\paragraph{Principle.} Adapter targets are chosen by where the optimizer-state mass sits rather than by where the weights appear low-rank. Both can be measured, and parameter mass decides. The class that dominates memory is the class to adapt, even when it is the least compressible class in the model.

\subsection{The flush that worked at small scale and failed at 120B}
\label{sec:tq-flush}

A rank-16 adapter cannot, in one shot, capture a matrix whose update needs three-quarters of its spectrum. A single rank-16 update against a high-rank target discards most of the update direction. The established way to recover the missing rank is to merge and restart: periodically fold the adapter delta into the base, reset the adapter factors, and let the next window accumulate a fresh low-rank increment, so that the cumulative update across many windows reaches whatever rank the training needs. This periodic flush is the core idea of ReLoRA \citep{ReLoRA} and PeriodicLoRA \citep{PeriodicLoRA}, and it is not a contribution of this work.

The flush worked in the sub-1B experiments that preceded the 120B, where the accumulated low-rank increments tracked the target updates and the base weights moved as intended, consistent with the prior reports. On that basis it was carried into the 120B bring-up as the assumed training mechanism.

At 120B it diverged. During bring-up, before the pretraining run could proceed, the flush-based configuration drove the model to divergence, and the divergence persisted across the mitigations attempted against it. Reaching a stable configuration meant abandoning the flush entirely and moving to flushless training, and the path to that point was expensive, costing substantial compute and time and a full restart of the run before flushless training was settled as a precondition rather than discovered mid-run. Both 120B runs, the pretraining run and the subsequent supervised run, were flushless from the start. The largest single loss reduction observed in the 120B training belongs to the flushless regime, and the flush-based attempts never reached a comparable trajectory.

This inverts the assumption the method began with. The justification for the flush is not that the trained weights are low-rank, which the measurement of Section~\ref{sec:tq-opt-state} contradicts, but that each short window's update can be approximated as low-rank if the flush refreshes the basis often enough. That is a claim about training dynamics, and it held at small scale and failed at 120B. The cause is not fully isolated. It is consistent with a scale-dependent interaction in which the per-window update at 120B is too high-rank for a rank-16 increment to track even with frequent merging, and the repeated absorption injects more disturbance than it resolves, but the records support the fact of the divergence and the success of the flushless alternative more firmly than they support any single mechanism for it.

\paragraph{Principle.} A periodic-merge scheme that succeeds at small scale is not assumed to transfer to a much larger model. Where the per-window update is too high-rank for the adapter to track, repeated merging can destabilize rather than help, and a flushless configuration that accumulates a single adapter delta can be the more stable choice. The transfer is verified at the target scale rather than inherited from smaller runs.

\subsection{Relationship to prior work}
\label{sec:tq-prior-work}

The components TQP composes are largely established, and the boundary between what is reused and what is new is worth stating precisely. The periodic merge-and-reset is from ReLoRA \citep{ReLoRA} and PeriodicLoRA \citep{PeriodicLoRA}, the former for pretraining from scratch and the latter for fine-tuning, and both predate this work. Quantized low-rank adaptation is established, as is dynamic-rank adaptation. The TurboQuant quantizer is from Zandieh and colleagues \citep{TurboQuant}. None of these primitives is claimed here.

What is not found in prior work is the combination in the setting it was applied: a TurboQuant-quantized base, adapted by low-rank factors, used as the training strategy for a drop-upcycled 120B mixture-of-experts, with a reversible backbone underneath that makes the configuration fit at 8K sequence length and batch 64 on a single node. ReLoRA and its relatives are demonstrated on dense models, mostly small and mostly from scratch. The integration with quantization, with expert upcycling, and with reversibility at this scale is, as far as can be determined, not previously reported, and the repurposing of an inference-side quantizer as a trainable base is itself part of that novelty. The claim is about the combination and the scale rather than any primitive.

There is corroboration from outside this work for the flush finding. A study of ReLoRA on small language models reports that the merge-and-restart approach helps larger models but not capacity-limited small ones, because a small model is already rank-deficient and the rank-expanding update has nothing to expand into \citep{ReLoRASLM}. That result and the divergence reported here both indicate that whether flush-based adaptation pays off depends on the rank headroom available, which varies with scale and with how far training has progressed, though they point at opposite ends of the scale range and neither isolates the mechanism at 120B.

\subsection{The rank measurement as a forward-looking instrument}
\label{sec:tq-rank-instrument}

The purpose of building TQP was not to show the weights were already low-rank. They are not, and the 9B sits some distance from that regime. The purpose was a training pipeline that works before the weights are low-rank, so that a 120B model can be trained on a single node at all, and in the configuration that worked, the flushless one, a fixed low-rank adapter accumulates a single delta over a long span rather than relying on repeated rank refreshes.

\begin{figure}[t]
\centering
\includegraphics[width=0.92\textwidth]{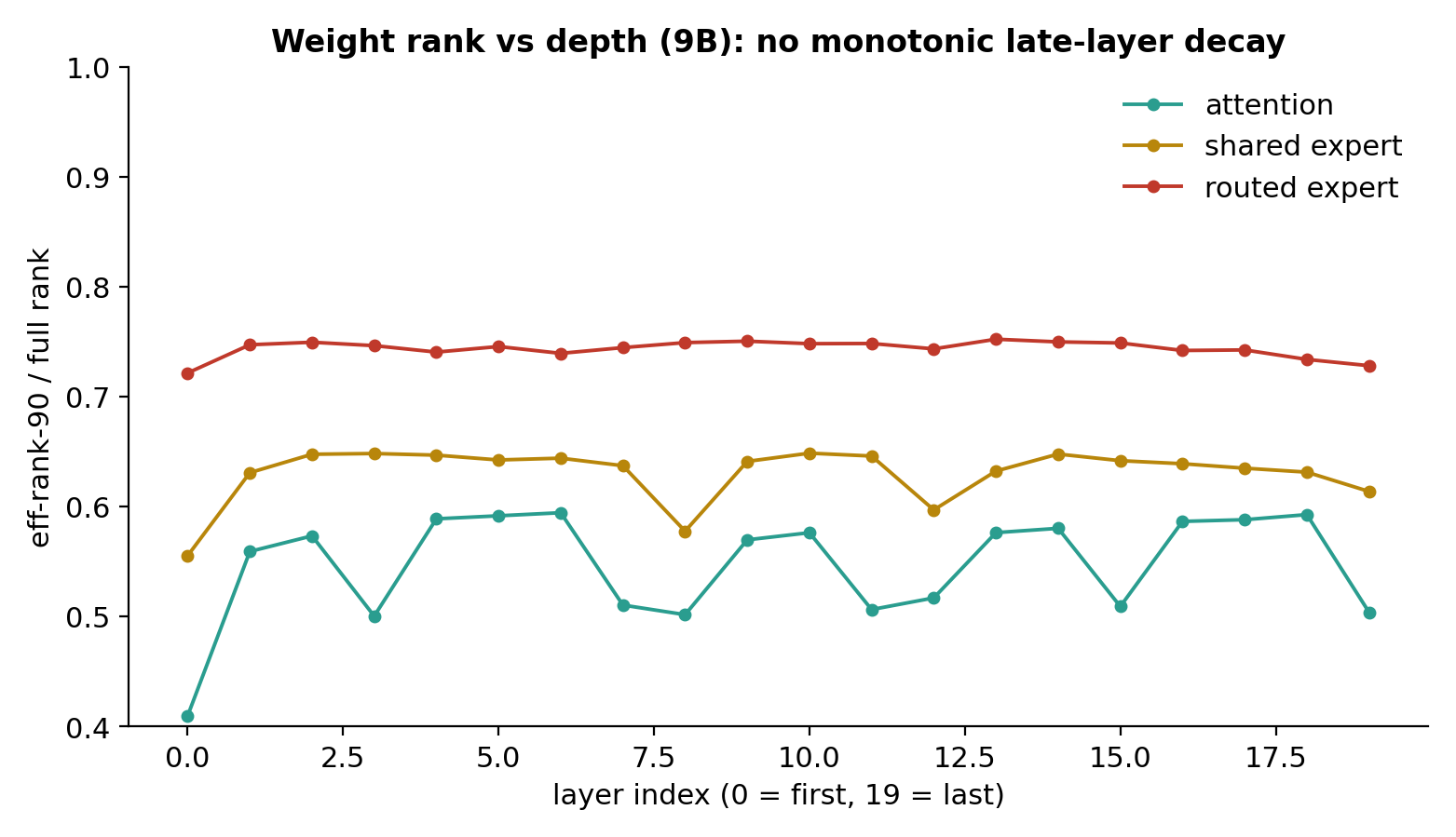}
\caption{Effective rank by layer depth in the released 9B, by weight class. The routed-expert rank is roughly flat across depth rather than tapering toward the later layers, which is why a single fixed adapter rank was used at every layer.}
\label{fig:rank_depth}
\end{figure}

The rank measurement then serves as an instrument rather than a justification. The routed experts at 0.74 are below full rank, and a well-trained model tends to shed effective rank as it absorbs more data, a tendency with support in the literature on implicit rank minimization under stochastic optimization with weight decay \citep{Galanti,Yunis,Kobayashi} and consistent with the observation that larger and better-trained models have lower intrinsic dimension \citep{Aghajanyan}. The depth profile of that fraction is roughly flat rather than tapering toward the later layers (Figure~\ref{fig:rank_depth}), which is the evidence behind using a single fixed adapter rank at every layer. Tracking that fraction over training indicates how much rank headroom remains, which is the quantity the flush decision turned on. The measurement is read here as a way to know how close the adapted class is getting to a genuinely low-rank regime, rather than as a claim about where it already is. For this run the experts were not there, and the rank-headroom account of the flush divergence in Section~\ref{sec:tq-flush} is offered as a candidate explanation that the measurement makes checkable in future runs rather than as an established cause.

\paragraph{Principle.} The effective rank of the adapted class is tracked over training as a diagnostic for which adaptation regime is appropriate. A high residual update rank is a reason to expect periodic merging to struggle and to prefer a flushless accumulation, and the measurement is the evidence on which that choice is made rather than a fixed schedule.

\subsection{Configuration}
\label{sec:tq-config}

Rank was fixed at 16 across every layer and both 120B phases. There was no per-layer or per-phase rank variation, and the depth profile of the rank measurement, roughly flat rather than tapering toward the later layers, would not have justified one. Both phases were flushless. The two 120B runs are a pretraining run, with next-token loss over the full sequence on the remaining pretraining pool, and a supervised run on question-and-answer data with loss computed on the answer tokens only and the prompt masked. The released checkpoint comes from the flushless supervised run and carries a final unmerged adapter delta, which is folded into the base during release consolidation. The full run lineage is described in Section~\ref{sec:training-120b}.

\begin{table}[h]
\centering
\caption{Table Q1. TQP configuration for the 120B runs.}
\label{tab:q1-tqp-config}
\begin{tabular}{lr}
\toprule
Setting & Value \\
\midrule
Quantizer & MSE TurboQuant, no QJL residual \\
Weight bit width & 8 \\
Adapter rank & 16 \\
Expert parallel size & 8 \\
Flush cadence & 0 (flushless) \\
Adapter learning-rate multiplier & 20.0 (supervised run; 3.0 in the pretraining run) \\
Router / rest learning-rate ratio & 0.1 (supervised run; 0.01 in the pretraining run) \\
\bottomrule
\end{tabular}
\end{table}

The adapter learning-rate multiplier and the router-to-rest ratio differ between the two 120B runs, and the reason is expert stabilization rather than the change of objective. When the model expanded to 460 experts the router carried early preferences, so the pretraining run held the adapter learning rate low, at a multiplier of 3.0, and the routing controls tight, at a router-to-rest ratio of 0.01, to let the experts settle before the router hardened onto them. Once the loss stabilized, indicating the experts had settled, the supervised run raised the adapter multiplier to 20.0 and loosened the ratio to 0.1. This loosening is coincident with the pretraining-to-supervised transition but is not caused by it; the values track expert stabilization, and Section~\ref{sec:train-controls} describes the stabilization controls in full.

\section{Training at 120B Scale}
\label{sec:training-120b}

Section~\ref{sec:turboquant} described TurboQuant Training as a method. This section describes what happened when it was run at 120B, as a continuous training lineage on one node across two logged runs separated by a restart. The method is transferable and the run is specific, and the specifics are where the operational lessons of training a model this large on constrained hardware live. The section covers the lineage as one chain, the two runs that compose it, the flush telemetry from bring-up, the routing health, the controls that kept the experts alive, and the consolidation that produced the released artifact.

\subsection{One lineage, two runs}
\label{sec:train-lineage}

The 120B is a single model lineage rather than two separately built artifacts. The same model continued without reinitialization across two logged runs: a pretraining run and a supervised run. The expanded 460-expert initialization, built by the corrected target-keyspace converter of Section~\ref{sec:fail-misload}, trained through the pretraining run on general pretraining data, and the supervised run then resumed from a checkpoint of that run and continued the same model on question-and-answer data. The two runs differ in objective. The pretraining run takes next-token loss over the full sequence, and the supervised run takes loss on the answer tokens only with the prompt masked, which is a supervised fine-tuning objective. They are reported as two runs for that reason, while remaining one model lineage with no reinitialization between them.

The lineage, end to end, is a single chain. The 9B harvest checkpoint became the consolidated 9B source. The corrected builder expanded it into the 460-expert 120B initialization. That initialization trained through the pretraining run and then the supervised run to a released periodic checkpoint, which was consolidated by merging the trained adapter into bf16 base weights and sharded for public release. There is one model in this chain and one released artifact. When the loss is read from the run, it is read from the logs of this continuous lineage rather than from any earlier standalone expansion experiment that shares a name.

Both runs were flushless. The flush experiment and its divergence belong to pre-production bring-up and are covered in Section~\ref{sec:tq-flush}; the production lineage described here never used it.

\begin{table}[h]
\centering
\caption{Table S1. 120B run specification.}
\label{tab:s1-120b-spec}
\begin{tabular}{ll}
\toprule
Setting & Value \\
\midrule
Hardware & one 8-GPU B200/B300 node \\
Sequence length & 8192 \\
Global batch size & 64 \\
Routing & top-12 over 460 routed experts \\
TQP rank & 16 (fixed) \\
Expert parallel size & 8 \\
Flush & off, both runs \\
Pretraining-run objective & next-token loss, full sequence \\
Supervised-run objective & loss on answer tokens, prompt masked \\
OPUS selector & off for 120B \\
Released checkpoint & step 5161 (supervised run) \\
\bottomrule
\end{tabular}
\end{table}

\subsection{The two runs and the loss trajectory}
\label{sec:train-loss}

The pretraining run trained for the first 3009 steps. The supervised run resumed from the step-2909 checkpoint of that run rather than its final logged step, because the step-3009 checkpoint was corrupted by a shard-save error and the clean step-2909 checkpoint was the latest sound one to continue from. The supervised run continued from step 2909 onward to a last logged step of 5184. This is the reason the two runs' step ranges overlap in the logs: the supervised run resumes a hundred steps before the pretraining run's logging endpoint, by design, because the step-3009 checkpoint was unreadable and the latest sound checkpoint was step 2909, rather than at a clean cut.

The loss trajectory across the two runs is the 120B's contribution to the descent shown in Section~\ref{sec:results}, and both runs are flushless. The pretraining run brought the next-token loss down to about 2.40 by its end. The supervised run carried it down further, with a run minimum of 1.520 at step 5059 and a last logged step of 1.862 at step 5184. The figure reported as the headline 120B loss is tied to the released checkpoint specifically, step 5161 in the supervised run, whose instantaneous training loss is 1.790 and whose trailing-100-step mean is 1.780. The trailing-100 mean at the released step, 1.78, is reported rather than the run minimum or the log endpoint, because it is the stable level of the checkpoint that was shipped. All of these are training-loss figures from the run logs; the held-out per-domain evaluation is in Section~\ref{sec:data-outcome}.

\begin{table}[h]
\centering
\caption{The two 120B runs of the continuous lineage and their loss trajectories.}
\label{tab:120b-two-runs}
\footnotesize
\begin{tabular}{lp{1.5cm}lp{3.0cm}p{2.5cm}p{4.2cm}}
\toprule
Run & Steps & Flush & Objective & Data & NTP/loss \\
\midrule
Pretraining & 1 to 3009 & off & next-token, full sequence & general pretraining mix & \textasciitilde 2.40 at run end \\
Supervised & 2909 to 5184 & off & answer-only, prompt masked & question-and-answer data & released step 5161: trailing-100 1.78 (min 1.520 at 5059; log endpoint 5184: 1.862) \\
\bottomrule
\end{tabular}
\end{table}

\paragraph{Principle.} The loss of a continued run is read from the continuous log of the actual job, and a trailing-window mean at the harvest point is reported rather than a run minimum. The minimum is a single step and overstates the shipped model, whereas the trailing mean is the stable level the released checkpoint came from.

\subsection{What the flush did during bring-up, before it diverged}
\label{sec:train-flush-telemetry}

The flush experiment that preceded the production runs left the most detailed mechanism telemetry, in the debug-class bring-up runs where the per-step flush metadata was recorded. It is presented here as evidence of what the flush did mechanically, not as production-run telemetry, because it comes from those early bring-up logs rather than the flushless production lineage, and because the flush configuration ultimately diverged and was abandoned, as Section~\ref{sec:tq-flush} describes. With flushing every ten steps, two quantities were logged: the mean adapter norm across the injected layers, and the share of each effective weight still coming from the frozen base rather than the adapter delta.

\begin{table}[h]
\centering
\caption{Bring-up flush telemetry: adapter norm and base-weight share over the first few hundred steps of the abandoned flush-based configuration.}
\label{tab:flush-telemetry}
\begin{tabular}{rrr}
\toprule
Step & Mean adapter norm & Base-weight share \\
\midrule
10  & 0.211 & 98.8\% \\
30  & 0.529 & 96.1\% \\
50  & 1.732 & 94.2\% \\
100 & 3.626 & 87.1\% \\
130 & 4.120 & 86.9\% \\
220 & 4.271 & 83.4\% \\
260 & 5.005 & 83.7\% \\
\bottomrule
\end{tabular}
\end{table}

Over a few hundred steps the adapter norm rose from about 0.21 to about 5.0 and the base-weight share fell from nearly all of the weight to around 83 percent, with small reversals rather than a strict monotone path. The flush was, in this window, transferring reshaping from the adapters into the base as intended, which is the cumulative-update mechanism of Section~\ref{sec:tq-flush} operating before the configuration later diverged. The telemetry shows the mechanism doing what it was designed to do over a short horizon, and Section~\ref{sec:tq-flush} records that the same mechanism failed to remain stable at the scale and over the duration of a full run, which is why the production lineage is flushless. The value of the table is as a direct picture of base-to-adapter transfer rather than as evidence that the flush was the production strategy.

\paragraph{Principle.} A flush is instrumented directly, logging adapter norm and base-weight share, rather than assumed to be working. A flat adapter norm or an unmoving base share is an early sign that the adapters are idling, and short-horizon evidence that the mechanism transfers reshaping is not on its own evidence that it remains stable over a full run, which is verified separately at scale.

\subsection{Routing health at scale}
\label{sec:train-routing-health}

\begin{figure}[t]
\centering
\includegraphics[width=0.92\textwidth]{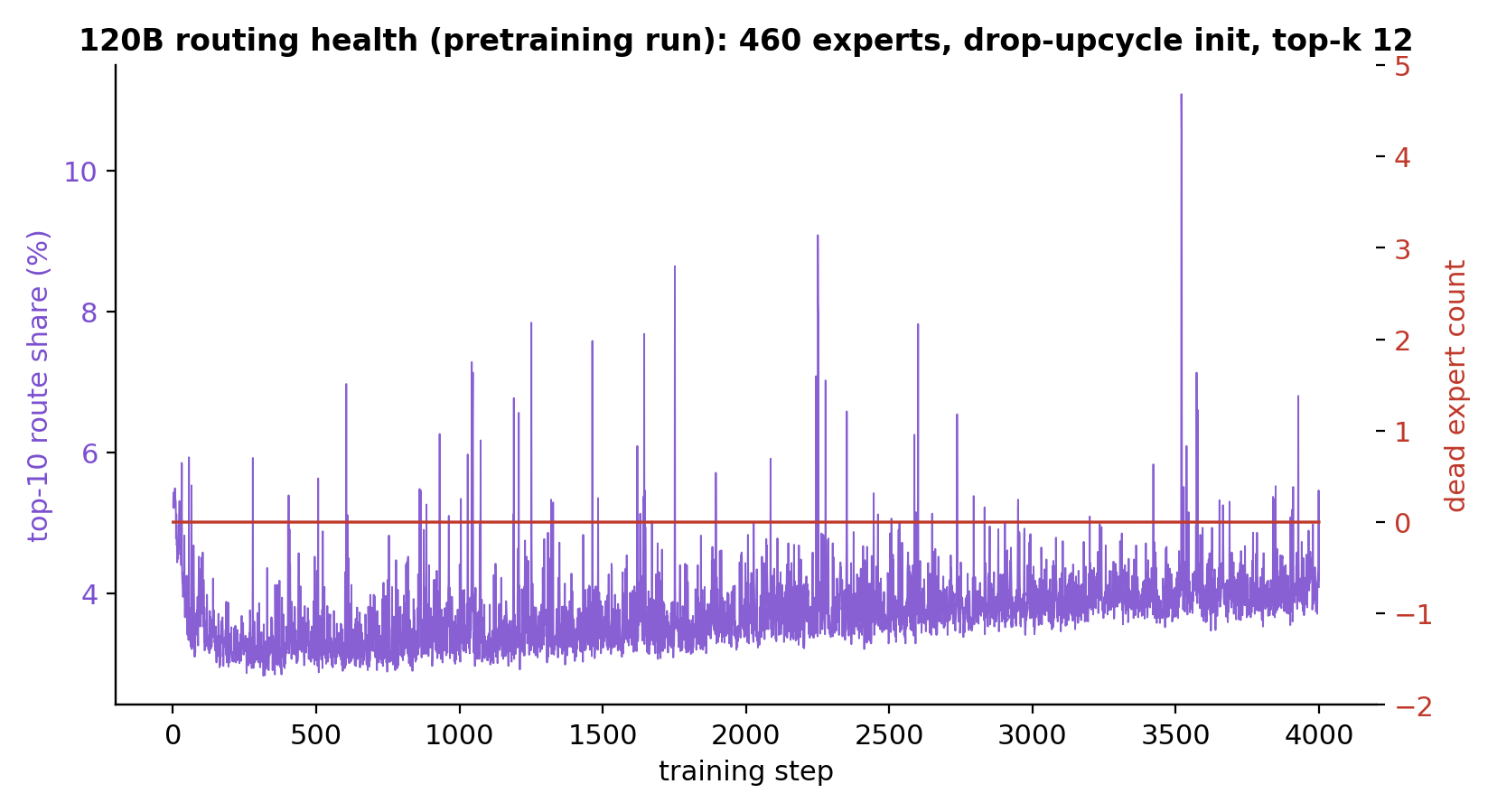}
\caption{120B routing health through the logged balance window (to step 4000): zero dead experts throughout, and a top-10 route share holding near 4 percent against a 2.17 percent uniform baseline.}
\label{fig:expert_balance}
\end{figure}

A 460-expert model can fail quietly by collapsing its routing, the failure mode of Section~\ref{sec:fail-clone-collapse}, so routing health is watched rather than assumed. The expert-balance telemetry shows the routing in a healthy regime through the logged balance window, which runs to step 4000. The bound is stated carefully: the balance record ends at step 4000, before the released checkpoint at step 5161, so the reported health covers the window that exists rather than the full run. At the last logged balance step no experts were dead and the maximum single-expert load was about half a percent of traffic. The logged concentration metric is the share of routing taken by the top ten experts, which sat at about 4.09 percent. Because the model routes top-12 over 460 experts but the logged metric is a top-ten share, the baseline for comparison is the top-ten uniform share, ten over 460, about 2.17 percent. At roughly 1.9 times that uniform baseline the routing is mildly concentrated rather than collapsed, and with zero dead experts and half-a-percent maximum load it is healthy without being perfectly flat (Figure~\ref{fig:expert_balance}). The routing controls described next held this window healthy. No single one of those controls is claimed as solely responsible across the whole run, because the balance artifacts cover the window to step 4000 rather than the segment through release.

\subsection{Controlling expert divergence}
\label{sec:train-controls}

Keeping 460 routed experts alive was an active part of running the 120B rather than a property that held on its own. The clone-family collapse of Section~\ref{sec:fail-clone-collapse} is the failure this control system exists to prevent: under hard top-$k$ routing the expanded experts concentrate onto a few winners and the rest die before they differentiate. The controls below are what held the routing in the healthy regime that Section~\ref{sec:train-routing-health} reports, and they were not all set once and left. Two of them were retuned mid-run in response to the balance telemetry.

The balancing is loss-free across the whole MoE family, in the sense of Section~\ref{sec:system}: load is equalized by moving a per-expert routing-logit bias rather than by adding a differentiable load-balancing term to the objective. What changed at 120B was the geometry of the controller that moves that bias. The 5B and 9B used a sign-based update that nudges each expert's bias against the sign of its load error. The 120B replaced this with a three-tier controller that applies a quadratic boost once an expert's load falls outside a target band, so an expert drifting far from its share is corrected more strongly than one near the band edge, which the larger expert count made necessary. The expansion itself seeds the 460 experts with a small router perturbation, a noise term of 0.05 on the tiled router, so the clones do not begin identical, and probabilistic selection is applied during the early differentiation window so near-duplicate clones receive gradient before the router hardens onto a representative. The z-loss weight is zero in the 120B objective, so the z-loss columns that appear in the logs did not contribute to the trained loss.

The learning-rate regime is itself a divergence control, and it changed between the two 120B runs for that reason. When the model expanded to 460 experts the router arrived with early preferences, so the pretraining run held the adapter learning rate low and the routing controls tight, with a router-to-rest learning-rate ratio of 0.01, so that routing moved slowly while the experts settled and the router did not harden onto its initial preferences. Once the loss stabilized, indicating the experts had settled, the supervised run raised the adapter learning rate and loosened the ratio to 0.1, giving the experts room to move. This loosening coincides with the pretraining-to-supervised transition but is driven by the observed stabilization rather than by the change of objective, and the per-run learning-rate values are in Section~\ref{sec:tq-config}.

Three controller settings were adjusted during the 120B run from the balance telemetry rather than fixed in advance, the learning-rate loosening above being one of them. The band ceiling of the three-tier controller was tightened early, from 0.005 to 0.003 at step 193. The bias-update gain was reduced partway into the run, from $1\!\times\!10^{-4}$ to $5\!\times\!10^{-5}$ at step 732, slowing the controller after it was seen to over-correct. Both retunes are recorded in the reproducibility timeline of Section~\ref{sec:reproducibility}, and reproducing the routing health of the run depends on replaying these controller settings rather than only matching the architecture and optimizer.

\begin{table}[h]
\centering
\caption{Table S2. Expert-divergence controls at 120B.}
\label{tab:s2-divergence-controls}
\footnotesize
\setlength{\tabcolsep}{4pt}
\renewcommand{\arraystretch}{1.15}
\begin{tabularx}{\textwidth}{l X X X}
\toprule
Control & 5B / 9B & 120B & Purpose \\
\midrule
Load balancing & loss-free, per-expert bias & loss-free, per-expert bias & equalize load without a gradient term \\
Bias-controller geometry & sign-based update & three-tier band with quadratic boost outside the target band & correct far-out-of-band experts more strongly at high expert count \\
Bias-update gain & controller-specific & \texttt{MOE\_BIAS\_GAMMA} \texttt{1e-4} to \texttt{5e-5}, retuned at step 732 & slow the controller after observed over-correction \\
Band ceiling & not applicable & \texttt{MOE\_EXPERT\_CAP\_HI} \texttt{0.005} to \texttt{0.003}, retuned at step 193 & set the load band the boost acts outside of \\
Expansion perturbation & not applicable & router noise 0.05 on the tiled router & start clones non-identical \\
Probabilistic selection & not used & on, early differentiation window & give near-duplicate clones gradient before the router hardens \\
Router / rest learning-rate ratio & 0.1 of base & 0.01 (pretraining run) then 0.1 (supervised run) & move routing slowly while experts settle, then give room \\
z-loss weight & small or zero & \texttt{MOE\_W\_Z = 0} & not used in the 120B objective \\
\bottomrule
\end{tabularx}
\end{table}

\paragraph{Principle.} At large expert counts, expert survival is treated as an actively controlled quantity rather than an emergent one. The controller geometry is matched to the expert count, near-duplicate experts are protected by perturbation and probabilistic selection through the differentiation window, the routing parameters are moved slowly relative to the experts, and the controller settings are retuned from balance telemetry during the run rather than fixed in advance. The settings that produced a given routing-health result are part of what has to be replayed to reproduce it.

\paragraph{Evidence.} Config-grade for the controller geometry, the router perturbation, and the zero z-loss weight, which are from the run configuration. The two mid-run retunes are graded separately: the step-732 gain reduction is log-grade, recorded in the run logs, while the step-193 band-ceiling change is transcript-grade, recorded in the session record rather than a durable metrics field; both appear in the Section~\ref{sec:reproducibility} timeline. The learning-rate regime per run is config-grade, from Section~\ref{sec:tq-config}. The resulting routing health is the step-4000-bounded window of Section~\ref{sec:train-routing-health}.

\subsection{Consolidation and release}
\label{sec:train-consolidation}

The released artifact is not the training checkpoint as-is, because the training checkpoint carries quantized base weights and a separate unmerged adapter delta from the flushless supervised run. Consolidation closes the loop. The trained adapter is merged into the base weights, the result is materialized as bf16, and the model is sharded into the released safetensors files. The released periodic checkpoint, after this merge, is the public 120B model.

This step is what makes the released model an ordinary bf16 mixture-of-experts rather than a quantized-plus-adapter object that only the training stack can load. It is worth stating because it is the point at which the TurboQuant Training machinery, which exists to make training fit on one node, leaves the artifact. A user of the released model sees a normal bf16 MoE. The quantization and the adapters were a training-time strategy, and consolidation is where they are paid back into ordinary weights.

\paragraph{Principle.} When the training strategy leaves the released checkpoint in a non-standard form, such as a quantized base plus an unmerged adapter, consolidation is an explicit and verified release step rather than an afterthought. The artifact released to users is the ordinary-weights model with the training-time machinery merged away, rather than the training object itself.

The consolidated bf16 120B-MoE is publicly released at \url{https://huggingface.co/theschoolofai/LightningLM-0.1V-120B-MoE}.

\subsection{What the 120B run demonstrates}
\label{sec:train-demonstrates}

The claim of this section is narrow and concrete. A 120B-class sparse mixture of experts, with 460 routed experts and hundreds of gigabytes of state, was trained as one continuous lineage on a single eight-GPU node, at 8K context, across a flushless pretraining run and a flushless supervised run, to a released-checkpoint trailing-100 training loss of 1.78 at step 5161, with routing held healthy through the logged balance window by an actively tuned control system and a clean consolidated release at the end. The strategy that made it fit is Section~\ref{sec:turboquant}, the systems that made it fast are Sections~\ref{sec:reversibility} and~\ref{sec:throughput}, and the growth that built the initialization is Section~\ref{sec:growth}. This section is the evidence that the combination ran end to end at the largest scale, as an actual training job that finished and produced a released model.

\section{Results}
\label{sec:results}

\subsection{What this section reports}
\label{sec:results-what}

The evidence in this paper is a loss trajectory, not a benchmark table. This section reports two things: the held-out loss behavior of the family across the full lineage, which is the quantitative result, and a set of generation samples, which are qualitative illustration. A reader should hold the two to different standards. The loss curves are measured on held-out data and are the basis for every capability claim. The samples are selected completions from base pretrained models, shown to illustrate what the loss numbers correspond to in practice, and they carry the caveats stated next.

The models are base models. None of the released checkpoints was instruction-tuned in the conventional sense; the 120B supervised run of Section~\ref{sec:training-120b} masked loss to the answer span of question-and-answer data but did not produce a chat-aligned assistant. The prompts below are therefore completion prompts, written as the opening of a passage for the model to continue, rather than instructions or questions in a chat format. The samples were selected from a larger generation pool rather than sampled at random, so they show what the models can do at their best rather than their average behavior. Generation stop and drift points are shown with their control tokens intact, written here as \texttt{[END\_TURN]} and \texttt{[END\_OF\_TEXT]}, so the reader can see where a sample ended cleanly and where it ran on. Section~\ref{sec:results-baseboundary} shows a sample that drifts, because the drift is part of the honest picture of a base model.

\subsection{The continuous descent}
\label{sec:results-descent}

\begin{figure}[t]
\centering
\includegraphics[width=0.92\textwidth]{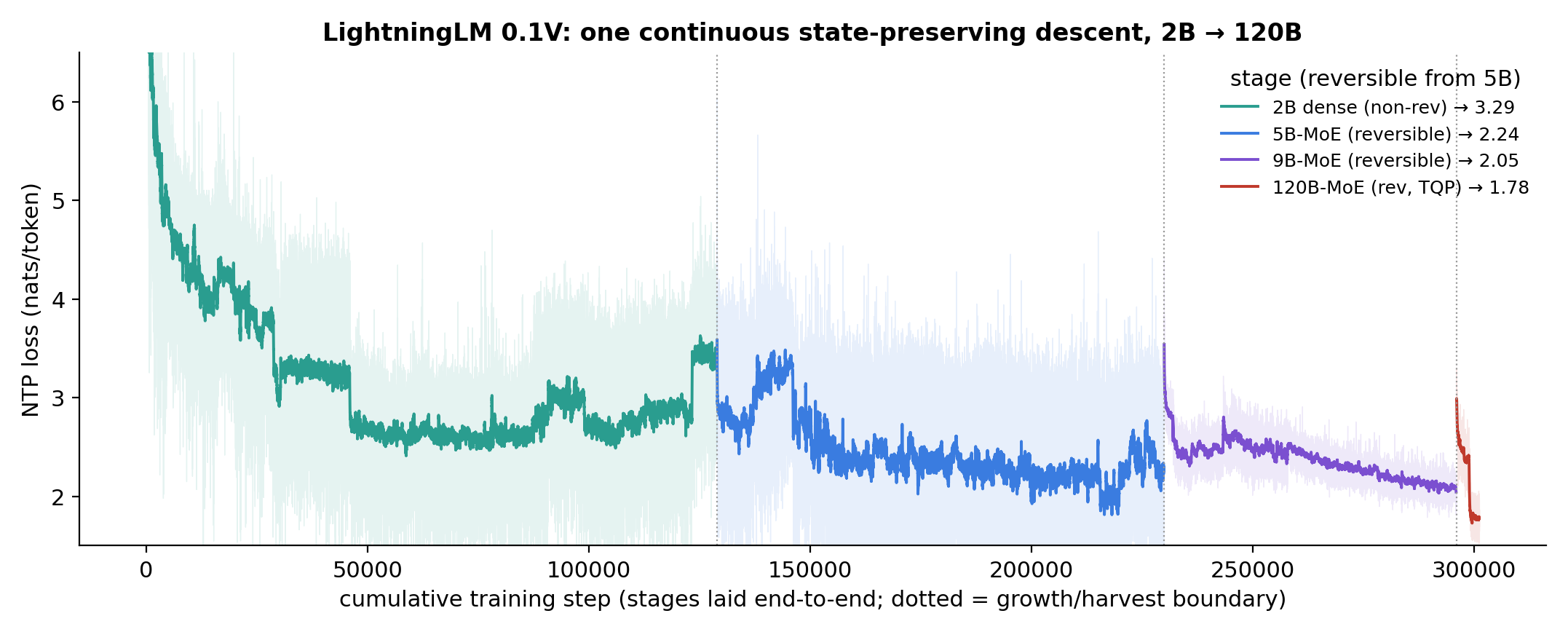}
\caption{LightningLM 0.1V: one continuous state-preserving descent from the 2B dense seed to the 120B mixture of experts, with the four stages laid end to end on a cumulative step axis. Dotted lines mark growth and harvest boundaries.}
\label{fig:combined_descent}
\end{figure}

The central quantitative result is a single state-preserving descent. Figure~\ref{fig:combined_descent} lays the four stages end to end on a cumulative step axis and plots next-token loss across the whole lineage. The loss falls from above six nats per token at the start of the 2B dense seed to 1.78 at the released 120B checkpoint, and at each growth boundary the larger model continues downward from where the smaller one was harvested rather than restarting from a high loss. The vertical excursions inside each stage are curriculum transitions to harder data buckets, after which the loss re-descends as the model adapts, so the absolute level is not directly comparable across stages with different bucket mixes. The shape that matters is that growth did not reset the descent. Each expansion preserved the learned state well enough that training resumed near the prior loss and kept falling, which is the visual statement of the growth thesis of Section~\ref{sec:growth}.

\subsection{Per-stage trajectories and the 120B run}
\label{sec:results-perstage}

\begin{figure}[t]
\centering
\includegraphics[width=0.92\textwidth]{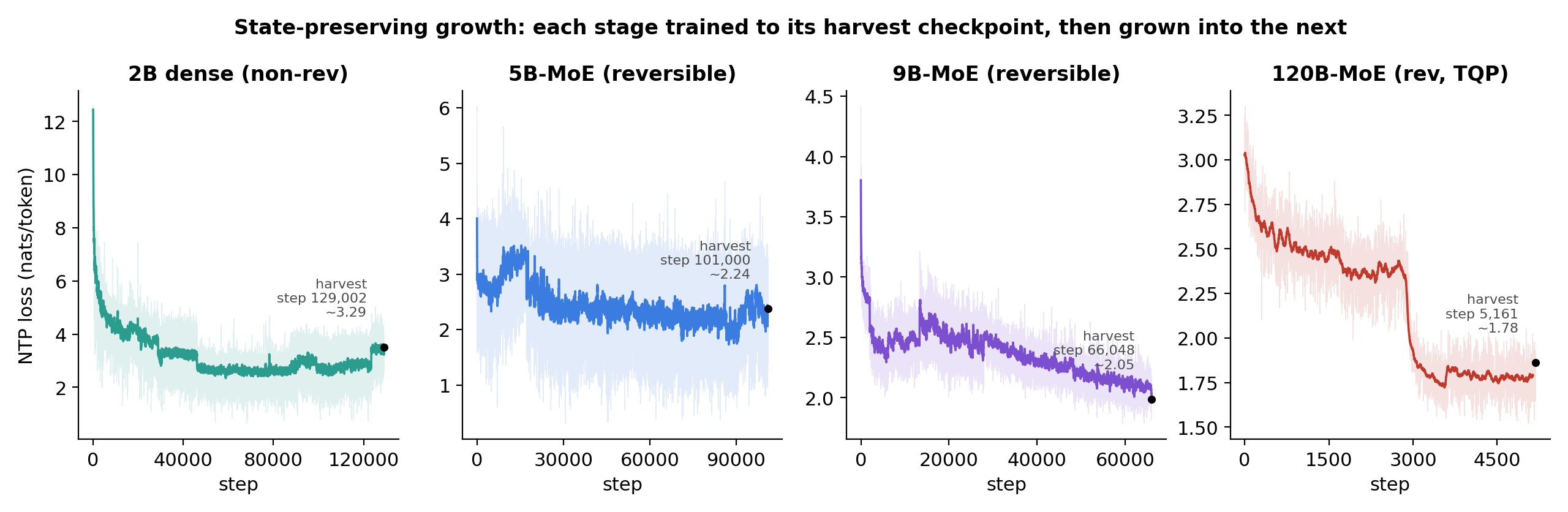}
\caption{Per-stage loss trajectories, each truncated at its harvest checkpoint: 2B dense (step 129{,}002, \textasciitilde 3.29), 5B-MoE (step 101{,}000, \textasciitilde 2.24), 9B-MoE (step 66{,}048, \textasciitilde 2.05), and 120B-MoE (step 5{,}161, \textasciitilde 1.78).}
\label{fig:per_stage}
\end{figure}

Figure~\ref{fig:per_stage} separates the four stages and marks the harvest checkpoint of each, the point at which it was consolidated and grown into the next. The trailing-window harvest losses fall monotonically across the family, from about 3.29 at the 2B dense seed, to 2.24 at the 5B mixture of experts, to 2.05 at the 9B, to 1.78 at the 120B. The upward excursions within each panel are again the harder-bucket transitions, most visible in the 5B and 9B panels, with the loss recovering below its pre-transition level each time.

\begin{figure}[t]
\centering
\includegraphics[width=0.92\textwidth]{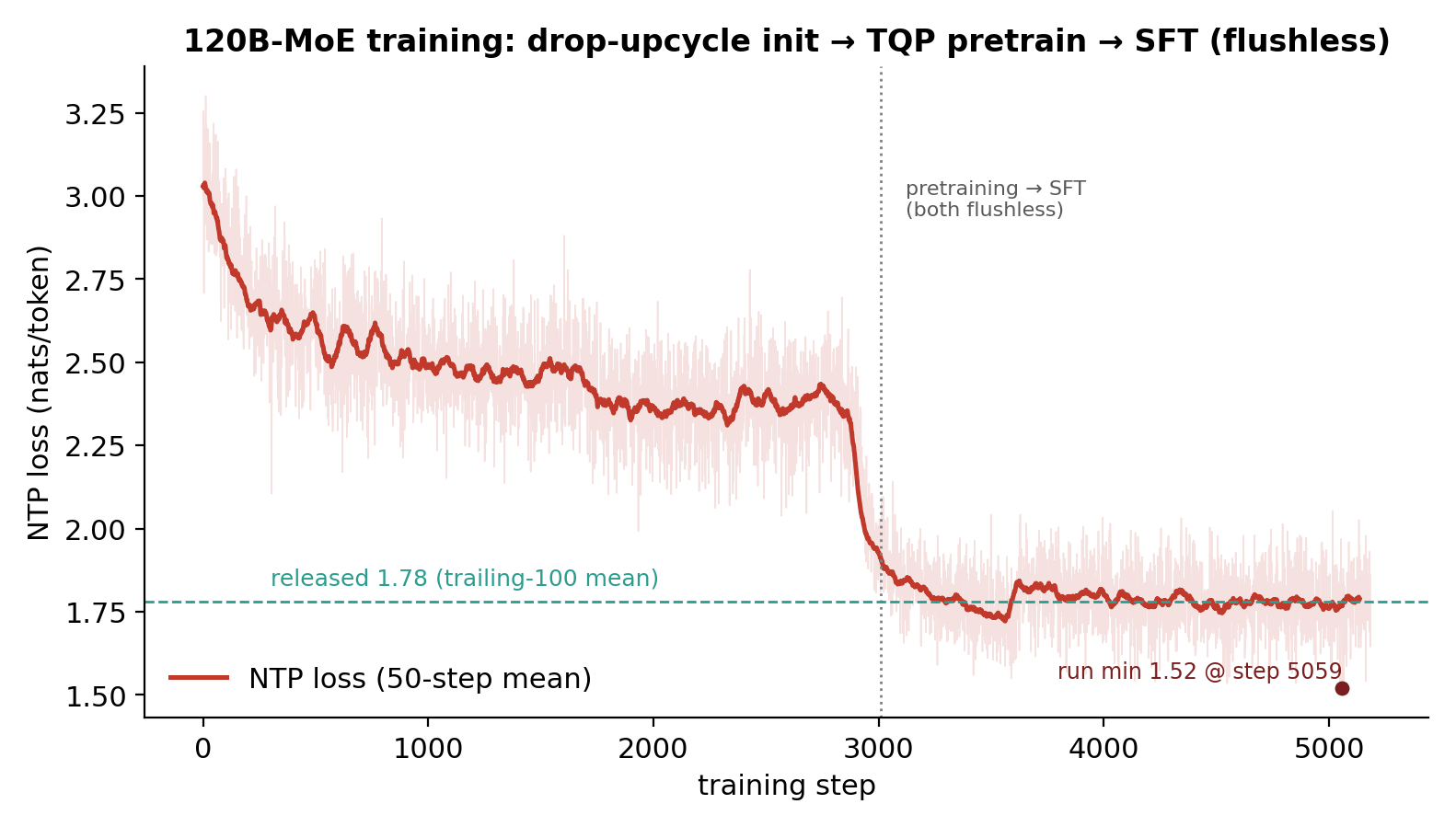}
\caption{The 120B-MoE training trajectory: drop-upcycle initialization, then a flushless TQP pretraining run and a flushless supervised run, one continuous lineage. The released checkpoint is step 5{,}161 at a trailing-100 loss of 1.78. The consolidated bf16 release is at \url{https://huggingface.co/theschoolofai/LightningLM-0.1V-120B-MoE}.}
\label{fig:loss_120b}
\end{figure}

Figure~\ref{fig:loss_120b} is the 120B run on its own. The trajectory shows the two flushless runs of Section~\ref{sec:training-120b} as one continuous descent: the pretraining run to about step 3009, then the supervised run resuming from the step-2909 checkpoint and continuing to step 5184. The largest single drop in the curve is at the transition into the supervised run, after which the loss settles to the released level. The released checkpoint at step 5161 has an instantaneous training loss of 1.790 and a trailing-100-step mean of 1.78, which is the figure reported as the headline 120B loss; the run minimum of 1.52 at step 5059 is marked but is not the released level, for the reasons given in Section~\ref{sec:train-loss}. All quantities in these three figures are training and held-out loss values from the run logs, not benchmark scores.

\subsection{Per-domain held-out loss}
\label{sec:results-domain}

The loss-based evaluation is most informative per domain, because the family was built to acquire specific capabilities and the per-domain held-out loss is the direct measurement of whether it did. Figure~\ref{fig:pool_domains} shows held-out loss at each stage's harvest checkpoint for code, STEM, a mean over the Indic scripts, and web text. Three patterns hold across the lineage. Code and STEM are the lowest-loss domains and improve fastest with scale, reaching 1.54 and 1.68 nats per token respectively at the 9B harvest. Web text is the hardest held-out domain at every scale, above four nats per token throughout, which is expected for the most heterogeneous pool. The Indic mean improves steadily with scale rather than stalling, which is the measured payoff of the guaranteed-composition data system of Section~\ref{sec:data}: the Indic share was protected in every batch against a selector that would otherwise have starved it, and the falling per-script loss is the evidence that the protection produced learning rather than mere exposure.

Figure~\ref{fig:indic_scripts} breaks the Indic mean into individual scripts at the 9B harvest. The per-script losses span a range that mostly reflects script and corpus difficulty rather than uneven learning, and the lower-resource scripts are not abandoned: each script the data system protected reaches a usable held-out loss at the 9B scale. The Devanagari and Hindi figures sitting above some lower-resource scripts is most plausibly an evaluation-set difficulty artifact rather than a model-quality result, a point made precise in the limitations of Section~\ref{sec:limitations}.

\subsection{Generation samples}
\label{sec:results-samples}

The samples below illustrate what the loss numbers correspond to. Each is a verbatim completion of the shown prompt, selected from the generation pool, with control tokens preserved. They are grouped by the capability each one is included to show.

\subsubsection*{Knowledge and technical reasoning at 120B}

The 120B produces correct multi-step technical content from a bare completion prompt. Given the opening ``Black holes are formed when'', it states the Schwarzschild-radius condition and derives the mass threshold:

\begin{quote}
\small
\textbf{Prompt:} \texttt{Black holes are formed when}\\[2pt]
\textbf{Output:} the mass of a massive star is compressed within its Schwarzschild radius, and the event horizon is reached. The Schwarzschild radius of a non-rotating black hole is calculated by the formula:
\[ R_s = \frac{2GM}{c^2} \]
\ldots A black hole forms when $R \leq R_s = 2GM/c^2$, equivalently the mass required for an object of radius $R$ to become a black hole is $M \geq R c^2 / (2G)$. \ldots once matter or light crosses $R_s$, it cannot escape to a distant observer.
\end{quote}

Given ``To prove that the square root of 2 is irrational,'', it produces the standard proof by contradiction, carrying the parity argument through to the shared factor of two and the contradiction with the lowest-terms assumption. Given ``The speed of light in a vacuum is'', it returns the exact constant 299{,}792{,}458 m/s with the rounded form. These are completions, not answers to questions, and the model continues them as well-formed expository prose with correct LaTeX-style mathematics.

\subsubsection*{Indic fluency}

The Indic samples are the qualitative counterpart to the per-script loss curves, and they are the visible payoff of the data guarantee of Section~\ref{sec:data}. The prompt is in the script and the continuation is in the same script: Indic in, Indic out. A Hindi prompt about Mahatma Gandhi continues into a correct account of his independence struggle:

\begin{quote}
\small
\textbf{Prompt:} \devtext{महात्मा गांधी ने भारत की स्वतंत्रता के लिए}\\[2pt]
\textbf{Output:} \devtext{संघर्ष किया था। उन्होंने भारतीय राष्ट्रवादी आंदोलन और सामाजिक सुधारों में महत्वपूर्ण भूमिका निभाई।}

\devtext{उन्होंने अहिंसा और सत्याग्रह के मार्ग से ब्रिटिश शासन का विरोध किया। असहयोग आंदोलन, नमक सत्याग्रह, सविनय अवज्ञा आंदोलन और भारत छोड़ो आंदोलन उनके प्रमुख आंदोलनों में शामिल हैं।}
\end{quote}

The continuation correctly describes the non-violence and satyagraha path of resistance and names the major movements (non-cooperation, the salt satyagraha, civil disobedience, and Quit India). A Kannada prompt about Bengaluru continues with an accurate description of the city as a technology center and the capital of Karnataka:

\begin{quote}
\small
\textbf{Prompt:} \kantext{ಬೆಂಗಳೂರು ಭಾರತದ ಸಿಲಿಕಾನ್ ವ್ಯಾಲಿ ಎಂದು}\\[2pt]
\textbf{Output:} \kantext{ಕರೆಯಲಾಗುತ್ತದೆ. ಇದು ಭಾರತದ ಪ್ರಮುಖ ತಂತ್ರಜ್ಞಾನ ಕೇಂದ್ರಗಳಲ್ಲಿ ಒಂದಾಗಿದೆ. ಇಲ್ಲಿ ಅನೇಕ ಐಟಿ ಕಂಪನಿಗಳು, ಸ್ಟಾರ್ಟ್‌ಅಪ್‌ಗಳು, ಸಂಶೋಧನಾ ಸಂಸ್ಥೆಗಳು ಮತ್ತು ಎಂಜಿನಿಯರಿಂಗ್ ಕೇಂದ್ರಗಳು ಕಾರ್ಯನಿರ್ವಹಿಸುತ್ತವೆ.}

\kantext{ಬೆಂಗಳೂರು ಕರ್ನಾಟಕದ ರಾಜಧಾನಿಯಾಗಿದೆ. ನಗರದ ತಂತ್ರಜ್ಞಾನ ಉದ್ಯಮ, ಶಿಕ್ಷಣ ಸಂಸ್ಥೆಗಳು ಮತ್ತು ಉದ್ಯಮಶೀಲತೆಯ ವಾತಾವರಣದಿಂದಾಗಿ ಇದನ್ನು ಭಾರತದ ಸಿಲಿಕಾನ್ ವ್ಯಾಲಿ ಎಂದು ಕರೆಯಲಾಗುತ್ತದೆ.}
\end{quote}

A second Kannada prompt about the history of Karnataka continues into a substantive answer that names the dynasties and the heritage sites:

\begin{quote}
\small
\textbf{Prompt:} \kantext{ಕರ್ನಾಟಕದ ಇತಿಹಾಸ ಬಹಳ}\\[2pt]
\textbf{Output:} \kantext{ಷ್ಟು ಸಮೃದ್ಧ ಮತ್ತು ವೈವಿಧ್ಯಮಯವಾಗಿದೆ. ಈ ಪ್ರದೇಶದಲ್ಲಿ ಕದಂಬರು, ಗಂಗರು, ಚಾಲುಕ್ಯರು, ರಾಷ್ಟ್ರಕೂಟರು, ಹೊಯ್ಸಳರು, ವಿಜಯನಗರ ಸಾಮ್ರಾಜ್ಯ ಮತ್ತು ಮೈಸೂರು ಒಡೆಯರು ಸೇರಿದಂತೆ ಅನೇಕ ರಾಜವಂಶಗಳು ಆಳಿವೆ.}

\kantext{ಕರ್ನಾಟಕದ ಇತಿಹಾಸವು ಕನ್ನಡ ಭಾಷೆ, ಸಾಹಿತ್ಯ, ದೇವಾಲಯ ವಾಸ್ತುಶಿಲ್ಪ, ಸಂಗೀತ, ಶಿಲ್ಪಕಲೆ ಮತ್ತು ವ್ಯಾಪಾರ ಪರಂಪರೆಯೊಂದಿಗೆ ನಿಕಟವಾಗಿ ಸಂಬಂಧಿಸಿದೆ. ಹಂಪಿ, ಪಟ್ಟದಕಲ್ಲು, ಬೇಲೂರು, ಹಳೆಬೀಡು ಮತ್ತು ಮೈಸೂರು ಈ ಪರಂಪರೆಯ ಪ್ರಮುಖ ಉದಾಹರಣೆಗಳಾಗಿವೆ.}
\end{quote}

The continuation names the Kadamba, Ganga, Chalukya, Rashtrakuta, Hoysala, Vijayanagara, and Mysore dynasties, and points to Hampi, Pattadakal, Belur, Halebidu, and Mysore as the heritage sites of that lineage. The fluency is in the script and the content at once, which is what the per-batch Indic guarantee was built to produce, and it is the direct qualitative complement to the per-script loss curve of Figure~\ref{fig:indic_scripts}.

\subsubsection*{Code}

The 120B and 9B complete code prompts with correct, idiomatic implementations. Given the signature and docstring of a Fibonacci function, the 120B completes the recursive base cases and recurrence. Given a merge-sort function header, the 9B completes a correct divide, recurse, and merge implementation. These are short completions, but they are syntactically and algorithmically correct continuations of the given context.

\subsubsection*{Tokenizer-level Indic capability}

A distinct and less obvious capability shows up in the spelling and character samples, and it is a direct consequence of the tokenizer rather than the model alone. The 9B can manipulate Indic text at the level of individual aksharas. Asked, in Kannada, how many letters the word for Bengaluru contains, it answers four and segments the word into its four aksharas:

\begin{quote}
\small
\textbf{Prompt:} ``\kantext{ಬೆಂಗಳೂರು}'' \kantext{ಪದದಲ್ಲಿ ಎಷ್ಟು ಅಕ್ಷರಗಳಿವೆ?}\\[2pt]
\textbf{Output:} 4. ``\kantext{ಬೆಂಗಳೂರು}'' \kantext{ಪದವನ್ನು ಬೆಂ, ಗ, ಳೂ, ರು ಎಂದು ನಾಲ್ಕು ಅಕ್ಷರಗಳಾಗಿ ವಿಭಜಿಸಬಹುದು.}
\end{quote}

The prompt asks how many aksharas are in the word, and the continuation answers four and splits it as \kantext{ಬೆಂ, ಗ, ಳೂ, ರು}. Asked to continue the spelling of Kannada and Hindi words akshara by akshara, it produces the correct sequence in the same way. This sub-word manipulation is possible because the BrahmicTokenizer-131K of the companion work represents Indic scripts at a granularity the model can address, and it is the kind of capability that a vocabulary which merely tolerated Indic text, rather than representing it economically, would not support.

\subsubsection*{What the samples do not show}

The samples are selected and the models are base models, so they show capability at its best rather than reliability. They do not show instruction-following, multi-turn coherence, or refusal behavior, none of which a base model has, and they should not be read as evidence of any of those. The next subsection shows the characteristic failure that the selection process otherwise filters out.

\subsection{The base-model boundary, shown}
\label{sec:results-baseboundary}

A base pretrained model continues the statistical structure of its prompt, and when that structure shifts mid-context the continuation can drift. The clearest example in the pool is a spelling completion that begins correctly in English and then drifts into another script:

\begin{quote}
\small
\textbf{Prompt:} \texttt{The spelling of umbrella is u, m, b, r, e, l, l, a.}\\
\texttt{The spelling of necessary is n, e, c,}\\[2pt]
\textbf{Output:} \texttt{e, s, s, a, r, y.} ``\teltext{పెద్దముద్దునూరు}'' \teltext{పదంలో 7 అక్షరాలు ఉన్నాయి: పె,}
\end{quote}

The English spelling of ``necessary'' is completed correctly, and then the model switches into Telugu and begins a different, unprompted akshara-counting task (stating that the Telugu place-name contains seven aksharas and starting to list them). Similar cross-script drift appears in other multilingual spelling samples. This is not a defect to be hidden; it is the expected behavior of a base model that has learned many scripts and tasks and has no instruction-following layer to hold it to one. It is shown here so that the strong samples above are read for what they are, the best of a base model's behavior, and not mistaken for the reliability of an aligned system. Post-training to convert this base capability into reliable instruction-following is outside the scope of this paper.

\section{Evidence Boundary and Limitations}
\label{sec:limitations}

This section is an audit of what the paper can and cannot prove. Evidence grade and scope are marked in place throughout, and the purpose here is to make the boundary precise: which claims rest on which kind of evidence, which inferences that evidence licenses, and which it does not. The aim is exactness rather than hedging, so that the strength of each result is stated rather than left to be reverse-engineered. None of this weakens the central result, that the pipeline ran end to end and produced the released models. It bounds the generality of the supporting evidence, and that bound is worth stating precisely.

\subsection{The kind of study this is}
\label{sec:lim-kind}

This is a systems and experience report. Its evidence is the feasibility, stability, throughput, and loss trajectory of an actual training lineage rather than a matrix of controlled ablations. The distinction governs how every systems claim should be read, and reversibility is the clearest case. No matched reversible-versus-non-reversible comparison of final loss was run at the larger scales, because the non-reversible configuration is the one that does not fit the single-node regime. The reversibility claim is therefore about reachability, that it made the regime trainable, rather than about a measured perplexity gain over an alternative that was run. The same holds for the growth principles and the throughput work: the report states what was chosen, why, and what happened, and where alternatives were run, as in the depth-mapping sweep and the upcycling bake-off, they are reported as such.

\subsection{Evidence grades}
\label{sec:lim-grades}

Evidence is graded on five levels, and each major claim is assigned its grade. Artifact-grade means reproducible from released code and public models. Checkpoint-grade means readable from a released model checkpoint. Log-grade means from a saved training log or metrics file. Transcript-grade means recorded only in a training-session transcript, with no surviving durable artifact. Recollection-grade means qualitative memory with no isolated trace.

\begin{table}[h]
\centering
\caption{Major claims and their evidence grade.}
\label{tab:evidence-grades}
\footnotesize
\setlength{\tabcolsep}{4pt}
\renewcommand{\arraystretch}{1.15}
\begin{tabularx}{\textwidth}{X l X}
\toprule
Claim & Grade & Basis \\
\midrule
Architecture, configuration, integrator settings & artifact-grade & released code \\
Memory Stream trained values & checkpoint-grade & released 2B checkpoint \\
Per-stage and 120B loss trajectories & log-grade & saved run metrics \\
Released 120B loss at step 5161 & log-grade & run metrics \\
Production throughput, ordinary and wall-clock & log-grade & run logs \\
Per-pool held-out eval & log-grade & eval metrics \\
120B bring-up flush telemetry (adapter norm, base share) & log-grade, bring-up runs & debug-class bring-up logs, not the production lineage \\
120B flush divergence at scale & log-grade, bring-up & bring-up logs; abandoned configuration \\
120B routing balance to step 4000 & log-grade, bounded window & balance metrics ending step 4000 \\
Expert-divergence controller retunes (steps 193, 732) & log-grade & run logs and Section~\ref{sec:reproducibility} timeline \\
Memory-feasibility (reversible vs not at 2B) & log-grade & runtime memory observations \\
Spectral upcycling near-random cold start & transcript-grade & session record; bake-off output not committed \\
Clone-family collapse trajectory (27 to 168) & transcript-grade & session record, clone-init variant \\
Always-ON over-exposure share and step count & transcript-grade & console-only composition tags, not all preserved \\
Mixture-shift instability & recollection-grade & qualitative, no isolated trace \\
\bottomrule
\end{tabularx}
\end{table}

Most of the paper is artifact-grade or log-grade. The transcript-grade items are a small set of internal diagnostic numbers whose telemetry was never written to durable storage, rather than the method or the models. The single recollection-grade item is flagged again below.

\subsection{What each load-bearing claim does and does not license}
\label{sec:lim-claims}

The sharpest form a limitation can take is a statement of which inference the evidence supports and which it does not. The load-bearing claims, in that form:

\begin{table}[h]
\centering
\caption{Load-bearing claims, their evidence, and the inferences they do and do not license.}
\label{tab:claim-license}
\footnotesize
\setlength{\tabcolsep}{3pt}
\renewcommand{\arraystretch}{1.15}
\begin{tabularx}{\textwidth}{XXXXX}
\toprule
Claim & Supported by & Not supported by & Valid inference & Invalid inference \\
\midrule
Reversibility enabled single-node scaling & the 2B reversible-vs-not feasibility comparison; the larger stages fitting at target regime & a matched loss ablation & reversibility made the regime reachable & reversibility improves final loss vs non-reversible, or the released 2B used it \\
120B experts were trained through TQP, flushless & production lineage and consolidation; the flushless loss trajectory & an isolated mechanism for why flush diverged at scale & the run used TQP without the flush, and flushless single-delta accumulation sufficed at this scale & the experts are low-rank, or the flush would have worked at this scale with more tuning \\
The flush diverged at 120B though it worked sub-1B & bring-up logs; the abandoned flush configuration; the prior small-scale runs & a controlled isolation of the cause & flush-based accumulation, sufficient below 1B, was not stable at 120B in this setting & a specific named mechanism is the established cause of the divergence \\
Guaranteed-per-batch composition protected code and Indic exposure & per-pool held-out loss; the AON mechanism & a no-AON control run & those domains were learned, and AON guaranteed their exposure & AON alone caused the code and Indic results \\
120B routing stayed healthy & balance metrics through step 4000; the controller and its retunes & balance artifacts past step 4000 & routing was healthy in the logged window under the active controls & routing was healthy through release at step 5161 \\
Growth preserved learned interfaces & the depth-mapping sweep; the lineage training down & controlled interface-break ablations beyond the sweep & the chosen mappings worked & the alternatives provably fail in general \\
\bottomrule
\end{tabularx}
\end{table}

\subsection{Evaluation is loss-based}
\label{sec:lim-eval}

The evaluation protocol is sustained loss reduction on held-out data of increasing difficulty. The paper makes no benchmark-leaderboard claim, and a reader should infer none, because no standardized downstream task suites were run. Within the loss-based evaluation there are three specific bounds. The held-out eval sets were not held fixed across stages, so some pool losses are not comparable from one stage to the next; the pools used as evidence are those whose eval sets moved toward harder rather than merely different material, and the Always-ON benchmark pool is excluded from cross-stage comparison for that reason. The pool losses are evidence about training dynamics, meaning how each domain improves as the model grows, rather than standardized model-to-model comparison numbers. The generation samples, where shown, are qualitative evidence of capability rather than scored results. The Devanagari and Hindi held-out losses sitting above lower-resource scripts is most plausibly an evaluation-set difficulty artifact rather than a model-quality result, and the two cannot be fully separated without size-matched eval sets that were not built.

\subsection{What the TQP rank evidence proves and does not}
\label{sec:lim-tqprank}

The rank measurement in Section~\ref{sec:turboquant} measured the effective rank of each weight class in the released 9B, as the fraction of the spectrum needed to capture ninety percent of the energy. That measurement showed the routed experts at 0.74, the least compressible class. The production lineage and the consolidation step prove the released 120B was trained through TQP, and both production runs were flushless, so the demonstrated result at this scale is that a fixed rank-16 adapter accumulating a single delta over a long span was sufficient to train the experts. The bring-up flush telemetry of Section~\ref{sec:train-flush-telemetry} illustrates, at log grade, that a rank-16 adapter under periodic flushing drives a growing base-to-delta transfer over a short horizon, which is the mechanism the flush was intended to provide; Section~\ref{sec:tq-flush} records that the same flush configuration diverged over a full run at 120B and was abandoned.

What this evidence establishes is bounded on three sides. It does not establish that the experts are low-rank, because they were measured not to be, at 0.74. It does not establish a specific mechanism for why the flush diverged at 120B while succeeding below 1B; the rank-headroom account of Section~\ref{sec:tq-flush}, that the per-window update is too high-rank for a rank-16 increment to track even with frequent merging, is offered as a candidate that the rank measurement makes checkable in future runs rather than as an isolated cause. And it does not establish that flushless accumulation is optimal, only that it was sufficient and stable where the flush was not. TQP therefore remains a training mechanism for fitting the optimization onto one node, rather than a claim that the experts admit a low-rank sufficient parameterization. The reversal of the original expectation, that the flush would be the load-bearing mechanism at scale, is itself one of the paper's findings, and the unisolated cause of that reversal is one of its open limitations.

\subsection{Artifacts that were not preserved}
\label{sec:lim-missing}

The transcript-grade and recollection-grade rows above correspond to specific telemetry that was never written to durable storage. The missing artifacts are: the saved numerical output of the drop, partition, and spectral upcycling bake-off; the drop and spectral bake-off checkpoints, of which only the partition checkpoints survive; the per-batch data-source tags, which were console-only by design, so the exact Always-ON composition cannot be re-derived from the metrics; an explicit positional-encoding-state field in the training metrics, so the positional transition is dated from screenshots rather than a logged flag; and per-expert routing histograms past the step-4000 balance window through release. None of these blocks reproduction of the method or the models from the released code and checkpoints. They mean that a handful of internal diagnostic numbers are evidenced from session records rather than from committed files, and the claims that rest on them are graded accordingly above. The positive reproducibility story, meaning what a third party can rebuild from the released code and models, is Section~\ref{sec:reproducibility}.

\subsection{The one recollection-grade claim}
\label{sec:lim-recollection}

The mixture-shift instability of Section~\ref{sec:fail-mixture-shift} is singled out because it is the only recollection-grade claim in the paper. Sharp curriculum transitions under frozen parameters were observed to threaten gradient spikes, and the mitigation was to blend across transitions, but no isolated saved trace exists, and no numbers were attached to it for that reason. A skeptical reader is right to treat it as the qualitative report it is. It is included because the lesson is consistent with the rest of the curriculum story, not because the spike can be shown.

\subsection{What the paper does not claim}
\label{sec:lim-non-claims}

The headline numbers invite over-reading, so the non-claims are stated directly. The paper does not claim dense all-parameter training of a 120B model on one node; the model is a sparse mixture of experts trained through TQP, a quantized base with a trainable low-rank update state, as Sections~\ref{sec:turboquant} and~\ref{sec:training-120b} set out, and released as consolidated bf16 weights. It does not claim benchmark performance, since the evaluation is loss-based. It does not claim that reversibility or the growth principles produce a measured loss improvement over the alternatives, only that they made the single-node regime feasible. It does not claim the flush primitive as novel, nor that the flush succeeded at this scale; the flush is prior work that was evaluated at 120B, found to diverge, and replaced by flushless training, as Sections~\ref{sec:tq-flush} and~\ref{sec:training-120b} detail. And it does not claim these methods are the only or best way to train at this scale.

The scale claim attracts the most scrutiny and is therefore worth stating with full precision. The 120B is a sparse mixture of experts with 460 routed experts and top-12 routing, activating a small fraction of its total parameters per token, trained as a TQP object of quantized base weights plus a low-rank update state, and released as consolidated bf16 weights. The single-node result concerns that object rather than dense all-parameter training. The claim is the narrower one: substantially more usable training can be extracted from a single node than is generally assumed, and an end-to-end 120B-class sparse MoE pipeline built on that premise ran to completion and produced the released models. Reproducibility of that pipeline is the subject of Section~\ref{sec:reproducibility}.

\section{Reproducibility}
\label{sec:reproducibility}

This section states exactly what the release provides and what each reproduction step requires from the user. The LightningLM 0.1V model repositories are public, gated, Apache-2.0 weights-only releases providing the consolidated bf16 artifacts and tokenizer files. The training-code release at \texttt{github.com/The-School-of-AI/LLM} provides the executable pipeline: model code, training entrypoints, growth transforms, TQP modules, tokenizer artifacts, configs, manifests, runtime hot-configuration documentation, and a stage-by-stage cookbook. A user supplying their own data, compute, storage, and checkpoint paths can follow the released code to grow the lineage end to end through the 120B TQP stage.

Reproducibility is organized around three layers.

\begin{enumerate}
\item Public artifact reproducibility: the released weights, tokenizer, tokenizer repository, Kronecker repository, and companion papers.
\item Released-code reproducibility: the training-code repository, which exposes the pipeline from data preparation through 120B consolidation.
\item Internal evidence reproducibility: the saved logs, checkpoints, and training-session records that establish how the released checkpoints were produced at a level finer than the code can express.
\end{enumerate}

The public artifacts fix the model-family identities. The released code makes the lineage executable. The internal evidence fixes the production history. The three layers are described in order.

\subsection{Released artifacts}
\label{sec:repro-artifacts}

The released model artifacts are consolidated bf16 weights. A user loading the eventual runnable release sees ordinary LightningLM mixture-of-experts weights rather than the training-time TQP object used to fit the 120B run on one node. The 120B training checkpoint carried quantized expert bases plus rank-16 update state; consolidation merged the trained update into bf16 expert weights and wrote safetensors shards. TQP is therefore a training strategy rather than the exposed inference format.

The public artifact state below was captured on 2026-06-03. Hugging Face repository head SHAs can change when model cards are edited, so the SHA column is a captured revision rather than a permanent architectural fact. The submission version pins the SHAs live at paper freeze.

\begin{table}[h]
\centering
\caption{Public artifacts and captured revisions for the LightningLM 0.1V release.}
\label{tab:released-artifacts}
\footnotesize
\setlength{\tabcolsep}{4pt}
\renewcommand{\arraystretch}{1.15}
\begin{tabularx}{\textwidth}{l X l X}
\toprule
Artifact & Public location & Captured revision / identifier & Release state \\
\midrule
2B model & \url{theschoolofai/LightningLM-0.1V-2B} & \texttt{313eb7d1\dots} & gated weights-only, Apache 2.0 \\
5B-MoE model & \url{theschoolofai/LightningLM-0.1V-5B-MoE} & \texttt{88c7a209\dots} & gated weights-only, Apache 2.0 \\
9B-MoE model & \url{theschoolofai/LightningLM-0.1V-9B-MoE} & \texttt{9357fe3e\dots} & gated weights-only, Apache 2.0 \\
120B-MoE model & \url{theschoolofai/LightningLM-0.1V-120B-MoE} & \texttt{35a245ae\dots} & gated, Apache 2.0, 31 safetensors shards \\
Tokenizer & \url{theschoolofai/BrahmicTokenizer-131K} & \texttt{93df154c\dots} & public tokenizer artifact, Apache 2.0 \\
Tokenizer code & \url{github.com/theschoolofai/BrahmicTokenizer-131K} & public repository & construction and verification scripts \\
Tokenizer paper & arXiv \texttt{2605.29379} & BrahmicTokenizer-131K & companion paper \\
Kronecker code & \url{github.com/theschoolofai/kronecker-embeddings} & public repository & embedding reference implementation \\
Kronecker paper & arXiv \texttt{2605.29459} & Kronecker Embeddings & companion paper \\
Training code & \url{github.com/The-School-of-AI/LLM} & \texttt{2c18402e} (pinned) & training pipeline, configs, growth transforms, TQP modules, manifests, cookbook \\
\bottomrule
\end{tabularx}
\end{table}

The public 120B card matches the run-log architecture: $118{,}670{,}450{,}556$ total parameters, $5{,}927{,}559{,}036$ active parameters per token (about 5.93B, or 5.0\% of the total), 460 routed experts, one always-active shared expert, and top-$k$ $12$ routed experts active per token. Earlier internal notes that treated the Hugging Face card as stale are obsolete; the live card has been corrected.

\subsection{Stage configuration}
\label{sec:repro-stages}

The stage table lists the load-bearing configuration facts. The full configuration files in \texttt{configs/} of the training-code release are authoritative; the table is the paper-side audit of the settings that determine the lineage.

\begin{table}[h]
\centering
\caption{Per-stage configuration for the four LightningLM 0.1V stages.}
\label{tab:stage-config}
\footnotesize
\setlength{\tabcolsep}{3pt}
\begin{tabular}{p{4.0cm}rrrr}
\toprule
Setting & 2B & 5B-MoE & 9B-MoE & 120B-MoE \\
\midrule
Layers & 8 & 8 & 20 & 20 \\
Residual width & 4096 & 4096 & 4096 & 4096 \\
Vocabulary & 131{,}072 & 131{,}072 & 131{,}072 & 131{,}072 \\
Feed-forward form & dense & MoE & MoE & MoE \\
Routed experts & n/a & 20 & 20 & 460 \\
Null experts & n/a & 0 & 0 & 0 \\
Always-active shared expert & dense FFN path & yes & yes & yes \\
Routing top-$k$ & n/a & 2 & 2 & 12 \\
Routed expert width & n/a & 1024 & 1024 & 1024 \\
Shared expert width & 2048 (dense) & 2048 & 2048 & 2048 \\
Reversible stack & no (released) & yes & yes & yes \\
Sequence length & 4096 & 4096 & 4096 then 8192 & 8192 \\
Global batch size & 32 & 32 & 120 at 4K, 56 at 8K & 64 \\
OPUS selector & off & on from step 0 & off & off in both runs \\
Always-ON data tier & active & active & active, known rampdown event & active, by run config \\
DRoPE & not used & not used & on at step 53{,}708 & on throughout 120B \\
TQP & no & no & no & rank 16 \\
TQP flush & n/a & n/a & n/a & off in both runs \\
Expert-parallel size & n/a & n/a & n/a & 8 \\
\bottomrule
\end{tabular}
\end{table}

Three values in this table are particularly easy to misreport.

First, the 9B 8K stable phase used global batch size 56 in the actual production run. Some planning files mention 128. That was a benchmark sweep target that ran out of memory and was excluded from production, as Section~\ref{sec:throughput} records. The production value is supported by the run logs showing 7 samples per rank at 8K on 8 GPUs and by the tokens-per-step figure of 458{,}752, equal to $56 \times 8192$, which matches the throughput tables of Section~\ref{sec:throughput}.

Second, DRoPE and learning-rate annealing are separate events. DRoPE turned the main RoPE path off at step 53{,}708. The anneal began around step 56{,}511, roughly 2{,}800 steps later. The intervening steps are the recalibration window. A reproduction that jumps directly from RoPE-on training into the anneal collapses two distinct operations and does not reproduce the run.

Third, the 120B used no periodic flush in either run. The every-ten-steps flush appears only in pre-production bring-up, where it diverged and was abandoned, as Sections~\ref{sec:tq-flush} and~\ref{sec:train-flush-telemetry} record. A reproduction that enables flushing for the 120B does not reproduce the released lineage.

\subsection{Lineage checkpoints}
\label{sec:repro-lineage}

The family is a growth lineage rather than four unrelated trainings. Each stage either trains the seed or grows from the previous stage's trained checkpoint. The distinction that matters is between a stage's public release endpoint and the checkpoint used as the source for the next growth transform.

\begin{table}[h]
\centering
\caption{Per-stage public-release endpoint and growth source.}
\label{tab:lineage-checkpoints}
\footnotesize
\setlength{\tabcolsep}{4pt}
\begin{tabular}{p{2.0cm}p{2.8cm}p{2.8cm}p{6.0cm}}
\toprule
Stage & Public release endpoint & Checkpoint for next growth & Notes \\
\midrule
2B dense & step 129{,}002 & step 129{,}002 & dense seed trained from scratch; 5B was partition-upcycled from this checkpoint \\
5B-MoE & step 101{,}000 & step 101{,}000 & the parallel-machine RYS depth-mapping sweep of Section~\ref{sec:growth-depth} ran on step 88{,}981 while the main 5B run continued; once the mapping was selected, the production 9B grow was applied to the final 5B at step 101{,}000 \\
9B-MoE & step 66{,}048 consolidated & step 66{,}048 consolidated & 120B was grown from the consolidated 9B source, locally \texttt{<local>/9b\_consolidated.pt} \\
120B-MoE & \path{step_5161_periodic} & n/a & training continued to step 5{,}184; release consolidation consumed the periodic checkpoint at step 5{,}161 \\
\bottomrule
\end{tabular}
\end{table}

The release identities for the first three public models were cross-checked by local checkpoint byte size against the Hugging Face total parameter counts. The implied bf16 parameter counts match the cards within $0.002$ to $0.004$ percent, consistent with checkpoint headers and buffers. Before final submission, a tensor-hash comparison against the public safetensors index would turn this from strong identity evidence into exact identity evidence.

\subsection{Reproducing the lineage}
\label{sec:repro-reproduce}

This subsection walks the executable chain exposed by the training-code release. Where the original operation was operator-driven, the release provides a deterministic reconstruction script and names the evidence boundary.

\subsubsection*{Train the dense seed}

The lineage begins with the dense seed, released publicly as the 2B-scale artifact because its shipped total parameter count is $1{,}781{,}570{,}624$. It trains from initialization to step 129{,}002 at sequence length 4096 and global batch size 32. OPUS was off for this stage throughout.

The released seed training entrypoint is \texttt{scripts/run\_2b\_stage.sh}. The dense model config is \texttt{configs/train\_2b.yaml}, the DeepSpeed config is \texttt{deepspeed/zero-3-1b.json}, the curriculum is \texttt{configs/curriculum\_v2.yaml}, and the AON and bulk-pool manifests are under \texttt{manifests/}. A reproducer supplying their own raw data and compute can run the stage from this entrypoint; the cookbook at \texttt{docs/cookbook.md} walks the stage end to end.

\subsubsection*{Grow the dense seed to the 5B MoE}

The 5B-MoE was not trained from scratch. It was grown from the dense step-129{,}002 checkpoint by partition upcycling. The dense FFN is copied into the always-active shared expert. Each routed expert receives an overlapping random partition of the dense FFN's intermediate neurons. This preserves the dense function first and gives the router related, non-identical expert fragments to specialize.

The production lineage uses the partition strategy. The training-code release packages this as \texttt{lightninglm/growth/dense\_to\_moe.py}, with the public command:

\begin{verbatim}
python -m lightninglm.growth.dense_to_moe \
  --src MODELS/2B/step_129002_consolidated.pt \
  --dst MODELS/5B/init_from_2b.pt \
  --strategy partition \
  --seed 1337
\end{verbatim}

The 2B and 5B stages share the same eight-layer backbone, so this step is a dense-to-MoE feed-forward upcycle rather than a depth expansion: the dense FFN populates the always-active shared expert and the random partition initializes the 20 routed experts. The same release exposes a \texttt{--strategy drop} path; the production lineage used partition. The exact seed reproduces the archived partition; with a different seed the method still reproduces but not the byte-identical expert partitions.

\subsubsection*{Train the 5B MoE}

The 5B stage trains from the partition-upcycled initialization to the public endpoint at step 101{,}000. OPUS was active from step 0. The clean rank-0 production JSONL records the run through step 101{,}000, including OPUS scoring passes. This is the stage that produces the released \texttt{LightningLM-0.1V-5B-MoE} checkpoint.

The same run also contains the step-88{,}981 checkpoint that was used for the parallel-machine depth-mapping sweep of Section~\ref{sec:growth-depth}. Running the sweep on the earlier step-88{,}981 checkpoint while the main 5B run continued toward step 101{,}000 on the primary machine was a wall-clock decision: it kept the main 5B run uninterrupted while the depth-mapping strategy was being chosen on a second machine. Once the mapping was selected from the sweep, the production 9B initialization was built by applying the selected mapping to the \textbf{final 5B at step 101{,}000}, not to step 88{,}981. A reproducer should grow 9B from the public step-101{,}000 5B endpoint; step 88{,}981 is the sweep checkpoint, not the production grow source.

\subsubsection*{Grow the 5B to the 9B by depth re-entry}

The 9B-MoE increases depth from 8 layers to 20 while preserving the MoE feed-forward design. The selected mapping is:

\begin{verbatim}
1-8, 1-4, 1-8
\end{verbatim}

Zero-indexed, this is:

\begin{verbatim}
[0,1,2,3,4,5,6,7, 0,1,2,3, 0,1,2,3,4,5,6,7]
\end{verbatim}

The reason for this mapping is the terminal-layer interface. The original layer 7 learned to feed the language-model head. The selected mapping ends the 20-layer model on that same source terminal layer. The alternative that looked slightly better in a short loss sweep ended on an interior layer, whose hidden-state distribution the language-model head had never been trained to read.

The historical production config \texttt{train\_9b\_8k\_stable.yaml} is not itself the fresh growth config; it resumes from an already-created 9B run state. To remove ambiguity, the training-code release includes a reconstruction script at:

\begin{verbatim}
lightninglm/growth/depth_map.py
\end{verbatim}

The public command is:

\begin{verbatim}
python -m lightninglm.growth.depth_map \
  --src MODELS/5B/step_101000_consolidated.pt \
  --dst MODELS/9B/init_from_5b_step_101000.pt \
  --mapping lightninglm_5b_to_9b \
  --layer-prefix layers
\end{verbatim}

The CLI flag value \texttt{lightninglm\_5b\_to\_9b} corresponds to the re-enter-early mapping described in Section~\ref{sec:growth-depth}. The public name is stage-specific so users can tell immediately that this operation is the 5B to 9B depth-growth initializer.

This script makes the mapping explicit for reproducibility. It does not prove that the original operator used the same filename; it captures the operation the release needs to expose. The production grow uses the final 5B checkpoint at step 101{,}000; a reproducer wishing to replicate the parallel sweep of Section~\ref{sec:growth-depth} would re-run the same command with \texttt{--src MODELS/5B/step\_88981\_consolidated.pt} against the three candidate mappings, and the production initialization would still be applied to step 101{,}000.

\subsubsection*{Train the 9B MoE}

The 9B production stage trains the 20-layer MoE through 4K and 8K context phases. OPUS was off in production. The run used Always-ON mixing, and an AON accounting issue later required an operator-side rampdown at step 37{,}179. Runtime hot-configuration controls for reproducing these operator-side events are documented in \texttt{docs/runtime\_hotconfig.md} in the training-code release.

The positional transition is also part of the reproduction. DRoPE activated at step 53{,}708, with the main RoPE path disabled. The anneal began around step 56{,}511. These are distinct events and both must be represented in the reproduction timeline.

The final 9B source for 120B growth is the consolidated step-66{,}048 checkpoint, locally represented as:

\begin{verbatim}
MODELS/9B/step_66048_consolidated.pt
<local>/9b_consolidated.pt
\end{verbatim}

The release identity is supported by the consolidated checkpoint and the public parameter-count match. The detailed per-step loss CSV recovered from the log backups covers steps 1 through 66{,}051, past the step-66{,}048 harvest used as the 120B growth source.

\subsubsection*{Grow the 9B to the 120B by target-keyspace drop-upcycling}

The 120B initialization is built from the consolidated 9B source by expanding each of the 20 source routed experts into 23 drop-upcycled variants, producing 460 routed experts. The correct converter is target-driven: instantiate the 120B model, walk the target model's expected parameter names, and populate each target tensor from the 9B source or a drop-upcycled transform. This avoids the silent-load failure mode where source-format keys fill an unused namespace while real target parameters remain random.

The release invocation is:

\begin{verbatim}
python scripts/build_120b_init.py \
  --src <local>/9b_consolidated.pt \
  --dst <local>/120b_init_proper_v2.pt \
  --config configs/train_120b_tqp.yaml \
  --ratio 0.5 \
  --router_sigma 0.05 \
  --seed 1337
\end{verbatim}

The important reproduction facts are \texttt{ratio=0.5}, \texttt{router\_sigma=0.05}, \texttt{seed=1337}, 23 clones per source expert, 20 source experts, and target-keyspace state-dict construction.

\subsubsection*{Train the 120B in two runs, both flushless}

The 120B is one continuous model lineage with no reinitialization between runs, and it is operationally two resumed runs that differ in objective and in stabilization regime. Both runs are flushless.

\begin{table}[h]
\centering
\caption{The two 120B runs of the continuous lineage.}
\label{tab:120b-runs-detail}
\footnotesize
\setlength{\tabcolsep}{4pt}
\renewcommand{\arraystretch}{1.15}
\begin{tabularx}{\textwidth}{l r r l X X l}
\toprule
Run & Start & End & Flush & Objective & Key knobs & Data state \\
\midrule
Pretraining & 0 & 3{,}009 & off & next-token, full sequence & rank-16 TQP, \texttt{adapter\_lr\_mult} 3.0, \texttt{router\_rest\_lr\_ratio} 0.01, ZeRO-1 & OPUS off \\
Supervised & 2{,}909 & 5{,}184 & off & answer-only, prompt masked & rank-16 TQP, \texttt{adapter\_lr\_mult} 20.0, \texttt{router\_rest\_lr\_ratio} 0.1, ZeRO-1 & OPUS off \\
\bottomrule
\end{tabularx}
\end{table}

The overlap in step ranges is a resume artifact, not a second model: the supervised run resumed from the step-2{,}909 checkpoint of the pretraining run, because the step-3{,}009 checkpoint was corrupted by a shard-save error and step 2{,}909 was the latest sound checkpoint to continue from. The released checkpoint is \texttt{step\_5161\_periodic} from the supervised run. Training continued past it to step 5{,}184, but consolidation consumed the periodic checkpoint at step 5{,}161.

The two runs differ in two independent ways, and the distinction matters for reproduction. The objective changes: the pretraining run takes next-token loss over the full sequence, and the supervised run takes loss on the answer tokens only with the prompt masked, which is a supervised fine-tuning objective. Separately, the learning-rate regime changes for expert-stabilization reasons that are not caused by the objective change. When the model expanded to 460 experts the router carried early preferences, so the pretraining run held the adapter learning rate low and the routing controls tight, at a router-to-rest ratio of 0.01, while the experts settled. Once the loss stabilized, the supervised run raised the adapter multiplier to 20.0 and loosened the ratio to 0.1. Reproducing the run means replaying both changes, and not assuming the learning-rate loosening was a consequence of moving to supervised data.

\subsubsection*{Consolidate the 120B release}

The 120B public artifact is produced by merging the trained TQP update into bf16 expert weights and writing safetensors shards. The local consolidation script reads:

\begin{verbatim}
<local>/ckpt_in/step_5161_periodic
\end{verbatim}

and writes:

\begin{verbatim}
<local>/120B-MoE
\end{verbatim}

The consolidation log records $237.3$\,GB written into $31$ safetensors shards and an index with $994$ keys. This closes the loop from the training-time TQP object to the released bf16 MoE artifact.

\subsection{Environment reproducibility}
\label{sec:repro-env}

The runtime is not a single environment. It changed across the project, and the 120B stack is materially different from the earlier A100/H200 stack. A reproduction document that listed one global \texttt{requirements.txt} would be wrong.

\begin{table}[h]
\centering
\caption{Two distinct runtime environments across the LightningLM project.}
\label{tab:env-repro}
\footnotesize
\setlength{\tabcolsep}{4pt}
\renewcommand{\arraystretch}{1.15}
\begin{tabularx}{\textwidth}{l l X}
\toprule
Scope & Evidence & Runtime \\
\midrule
Early stable stack for the A100/H200 era & \path{requirements-pinned.txt} and leak-fix notes & torch 2.7.1+cu128, triton 3.3.1, deepspeed 0.18.6, flash-linear-attention at commit \texttt{2f18f7d4\dots}, grouped-gemm 0.3.0 \\
120B B300 stack & \path{setup-b300.sh}, launch playbook, 120B logs & nightly PyTorch 2.13.0.dev (cu130), DeepSpeed 0.18.9, FLA 0.4.2, Triton 3.6.x-era runtime with a documented mixed-install hazard \\
\bottomrule
\end{tabularx}
\end{table}

The 120B environment has a load-bearing patch. The launch playbook documents a Triton installation state in which the \texttt{tl.dot} semantic check asserts equal operand dtypes, while FLA kernels can pass mixed bf16/fp32 operands. Without the soft-cast patch, the run can crash on the first backward path through the affected kernel. The training-code release includes:

\begin{verbatim}
scripts/triton_softcast_patch.py
\end{verbatim}

This patch, or an equivalent upstreamed fix, is part of the 120B environment setup. Treating it as an incidental local hack would make the reproduction instructions false.

\subsection{Operator-side reproducibility}
\label{sec:repro-operator}

Some important changes were made through hot configuration during live training rather than through static YAML files. Exact reproduction therefore requires replaying those events or starting from checkpoints taken after them.

The training-code release documents the hot-configuration controls at:

\begin{verbatim}
docs/runtime_hotconfig.md
\end{verbatim}

The load-bearing entries are:

\begin{table}[h]
\centering
\caption{Mid-run hot-configuration changes that must be replayed for exact reproduction.}
\label{tab:hotconfig}
\footnotesize
\setlength{\tabcolsep}{4pt}
\renewcommand{\arraystretch}{1.15}
\begin{tabularx}{\textwidth}{r l X l}
\toprule
Step & Stage & Change & Evidence grade \\
\midrule
193 & 120B pretraining run & \texttt{MOE\_EXPERT\_CAP\_HI} tightened from \texttt{0.005} to \texttt{0.003} & transcript-grade \\
732 & 120B pretraining run & \texttt{MOE\_BIAS\_GAMMA} changed from \texttt{1e-4} to \texttt{5e-5} & log-grade \\
37{,}179 & 9B production & AON rampdown after the effective 57\% AON over-exposure was identified & transcript-grade \\
\bottomrule
\end{tabularx}
\end{table}

The 120B balance mechanism is loss-free in the sense of \citet{LossFreeMoE}: balance is controlled by a per-expert routing-bias controller outside the main differentiable objective. \texttt{MOE\_W\_Z=0} in the recovered 120B hot-config snapshots, so z-loss may appear in logs without contributing to the 120B objective. Exact reproduction of routing health depends on replaying the bias-controller settings, not merely setting the architecture and optimizer. Section~\ref{sec:train-controls} describes these controls and their purpose; this subsection records them for replay.

The 9B AON event is a data-composition correction rather than a routing correction. The intended protected-data fraction and the effective hardcoded fraction diverged in the 8K phase, producing roughly $4/7$, or 57\%, Always-ON exposure for a stretch of training. Because per-batch source tags were not durably written as structured JSONL fields, the correction is transcript-grade rather than artifact-grade. It belongs in reproducibility because it changes the training distribution, and it is the failure of Section~\ref{sec:fail-curriculum}.

\subsection{What can be reproduced from the training-code release}
\label{sec:repro-can}

A third party can reproduce or verify the following from the released artifacts:

\begin{itemize}
\item The tokenizer artifact, its 131{,}072-vocabulary interface, and its construction and verification utilities.
\item The Kronecker embedding companion implementation and paper.
\item The public model identities, parameter counts, MoE top-$k$ values, and staged-release state.
\item The dense-seed to 5B partition upcycle.
\item The 5B to 9B depth growth using the \texttt{1-8, 1-4, 1-8} mapping.
\item The 9B to 120B target-keyspace drop-upcycle with \texttt{ratio=0.5}, \texttt{router\_sigma=0.05}, and \texttt{seed=1337}.
\item The 120B two-run, flushless TQP training sequence.
\item Tensor-hash generation for checkpoint verification.
\item Runtime hot-configuration controls needed for router balance and AON continuation behavior.
\end{itemize}

Full third-party retraining requires the user to supply raw data access, compute, and storage. The released training code provides the pipeline: tokenizer use or rebuilding, data preparation, manifest generation, staged training configs, growth transforms, 120B target-keyspace initialization, TQP training, tensor hashing, and runtime hot-configuration controls.

\subsection{Reproducibility claim}
\label{sec:repro-claim}

The claim supported here is precise. The released LightningLM 0.1V models are public bf16 weight artifacts with public tokenizer, companion embedding and tokenizer work, and a released training-code cookbook for growing the lineage. The reproduction recipe above is the checklist: train or load each stage, apply the growth transform in the correct checkpoint keyspace, replay the runtime hot-configuration controls when reproducing a run segment, run the two flushless 120B TQP runs, consolidate the final checkpoint, and compare tensor hashes against the public release.

This is more than a model dump. It states the public artifact state, the training pipeline, the growth transforms, and the operational controls needed for a user to grow their own LightningLM-style lineage through the 120B TQP stage.

\section{Conclusion}
\label{sec:conclusion}

This work trained a model in the hundred-billion-parameter class on a single eight-GPU node and documented what that required. The result is LightningLM 0.1V, a recurrence-backbone family grown in four stages from a small dense seed to a 120B sparse mixture of experts with 460 routed experts, trained end to end on single nodes, with the larger stages at full 8K context. The released model is the artifact; the recipe and the failures are the part meant to outlast it.

The conviction underneath the work is that more usable training can be drawn from one node than the field assumes, and that the price of doing so is a particular discipline rather than a particular cluster. The discipline has three parts, and they map onto the three things this paper tried to make reproducible. Activation memory is kept from growing with the model, through reversibility, so the larger stages fit. The model is grown by preserving its learned interfaces rather than reshaping its weights, so each stage starts from something that already works. The training cost of the largest stage is bounded by quantizing the expert base and training low-rank adapters over it, so the optimizer state fits where full state would not. None of these is a new primitive. Each is a known idea, and the contribution is the integration: the first two make the growth affordable, the third makes the final stage trainable, and together they let the whole lineage run on hardware a model of this size is not usually trained on.

The boundary of what this shows is stated exactly. It is a systems and experience report, evidenced by feasibility, throughput, and the loss trajectory of an actual run, rather than by controlled ablations or benchmark scores. Where the evidence is strong, from released checkpoints and saved logs, it is marked as such; where it is weaker, reconstructed from training-session records, it is marked as that; and where a claim is a candidate explanation rather than an established result, as with the mechanism by which the periodic flush diverged at this scale, it is labeled as a candidate rather than a finding. The limitations are not an appendix to that posture. They are the posture.

Two threads in the work are worth carrying beyond this particular model. The first is the failure catalogue. The most expensive failures were the ones that did not announce themselves: a checkpoint that loaded successfully with parts of it left random, an expert bank that collapsed back toward its source count, a data guarantee that a hardcoded constant quietly violated. Each was caught only by watching the one number that exposed it. The reusable practice, more than any single fix, is to identify which invariant each step depends on and to instrument the measurement that proves it, rather than trusting that a clean-looking artifact means the invariant held. The second thread is the data discipline. The capabilities that mattered most, Indic competence and code, were guaranteed into every batch against a selector that would otherwise have starved them, and the per-domain held-out loss is the measurement that the guarantee worked. Building capability by construction and then measuring exactly that construction is a pattern worth reusing.

What remains is a working composition of known parts, run at a scale and on hardware where that composition had not been shown to run, and documented at the level a practitioner would need to repeat it. The claim is not that this is the only or the best way to train at this scale, but that it is a way that works, that most of its hard parts are absent from the public record, and that writing them down, including the parts that went wrong, is the useful contribution. The state-preserving path from a small dense seed to a 120B sparse mixture of experts is open. The recipe is in Sections~\ref{sec:reversibility} through~\ref{sec:training-120b}, the boundary of the evidence is in Section~\ref{sec:limitations}, and the means to reproduce it are in Section~\ref{sec:reproducibility}.

\section*{Acknowledgments}
\label{sec:acknowledgments}

This work was carried out at The School of AI, Bangalore India. Training compute was provided by AWS.

A special thanks to Suman Debnath, who facilitated access to AWS credits and services that made the training runs possible, and to Shwetha D, who managed the complete training run end to end across all four growth stages. Without their contributions this would not have been possible.

We owe particular thanks to the following ERA V4 students, whose work on specific parts of the training pipeline during its build-out and bring-up made the production phase possible:

Abhijeet Bharade, Abishek Ajai Satnur, Ankita Mungalpara, Asang Kumar Singh, Balaji A J, Chethan Rao, Gaurav Chawla, Gaurav Sethi, Hemanth K, Jayant Guru Shrivastava, Mohan Raj T, Nandam Sriranga Chaitanya, Nikhil Mahesh, Nirmal Pratheep Natarajan, Nishant Bhansali, Nishanth Reddy, Nitin Vig, Pankaj Kumar, Prateek Mohan Garg, Raghu Rammohan, Sandip Dinkar Jadhav, Sarthak Duggal, Satish Waman Gune, Shanmuga Suntharam Sankaralingam, Shyamant Achar, Sidharth Ghag, Sijuade Oguntayo, Smita Sasindran, Snehashis Panigrahi, Sualeh Qureshi, Varsha Jain, Vikas Gupta, Vikas Kumar Jha, Vinayak Ganapuram, Vishnu Nandam, Yashwant Ram M, and Yasir Reshi.

\bibliographystyle{abbrvnat}
\bibliography{references}

\end{document}